\PassOptionsToPackage{hyphens}{url}
\documentclass[letterpaper]{article}
\usepackage[preprint]{aaai2027}
\usepackage{url}
\usepackage{graphicx}
\DeclareGraphicsExtensions{.jpg,.pdf,.png,.jpeg}
\usepackage{natbib}
\usepackage{caption}
\usepackage{multicol}
\usepackage{paver-layout}
\usepackage{booktabs}
\usepackage{amsmath,amssymb}
\usepackage{colortbl}
\definecolor{oursgray}{gray}{0.92}
\newcommand{\na}{--}
\newsavebox{\perhorizontablebox}
\usepackage[hidelinks]{hyperref}
\hypersetup{
  pdftitle={Planning-Aligned Pretraining of BEV Representations with Sparse Action-Conditioned Targets for End-to-End Autonomous Driving},
  pdfauthor={Jaeha Song, Soonmin Hwang}
}

\renewcommand{\topfraction}{0.95}
\renewcommand{\textfraction}{0.05}
\title{Planning-Aligned Pretraining of BEV Representations\\with Sparse Action-Conditioned Targets for End-to-End Autonomous Driving}
\author{
    Jaeha Song,
    Soonmin Hwang\corresponding
}
\affiliations{
    Hanyang University\\
    archiiive99@hanyang.ac.kr, soonminh@hanyang.ac.kr
}

\def\arxivpreprint{}
\newcommand{\analysisvariant}{_epoch60}
\definecolor{paverlink}{HTML}{209EFF}

\begin{document}
\maketitle

\begin{abstract}
End-to-end driving requires planning-relevant bird's-eye-view (BEV) representations, but existing pretraining approaches often rely on task annotations or dense scene reconstruction. We introduce PAVER, Planning-Aligned BEV Encoder Pretraining. From a single LiDAR sweep, PAVER constructs sparse risk and unknown targets describing occupied and unobserved evidence along rule-based ego motions. A 10K-parameter head predicts these targets from masked BEV features conditioned on the action state, directing supervision toward geometric constraints on candidate motions. Pretraining requires no driving-task annotations or dense reconstruction. Only the BEV encoder is transferred, preserving the downstream architecture and camera-only inference. On nuScenes, PAVER reduces VAD-Tiny's average collision rate from 0.51\% to 0.19\%, while improving planning L2, motion prediction, detection, and mapping. The selected VAD-Tiny and VAD-Base schedules use about 36\% less estimated total training time than scratch training, including pretraining. On Bench2Drive Town05 Long, PAVER improves UniAD-Tiny's closed-loop Driving Score from 48.45 to 58.79.
\ifdefined\arxivpreprint\ The project page is available at \href{https://archiiive99.github.io/PAVER/}{\textcolor{paverlink}{\nolinkurl{https://archiiive99.github.io/PAVER}}}.\fi
\end{abstract}

\section{Introduction}

Camera-based end-to-end driving combines detection, map prediction, motion prediction, and planning through a shared bird's-eye-view (BEV) representation \citep{hu2022stp3,hu2023planning,jiang2023vad,sun2024sparsedrive,zheng2024genad,liao2025diffusiondrive}. Learning cross-view geometry and planning-relevant features requires long downstream training schedules \citep{hu2023planning,jiang2023vad}. Image pretraining initializes the image backbone, but BEV construction must still be learned downstream \citep{deng2009imagenet,he2016resnet,li2022bevformer}.

\begin{figure}[!t]
\centering
\includegraphics[width=\columnwidth]{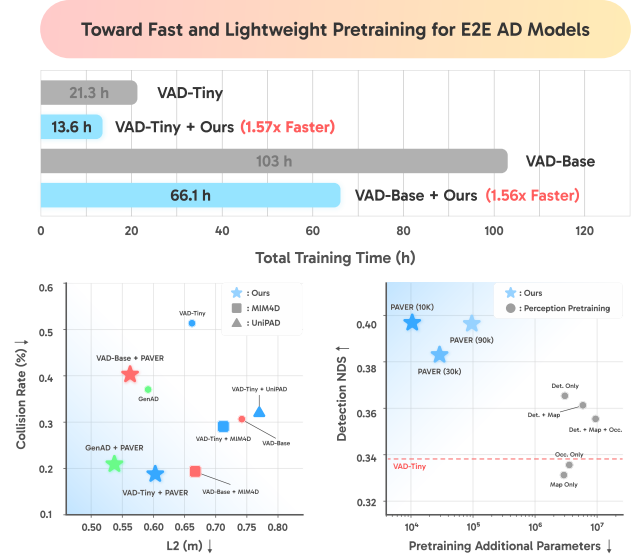}
\caption{PAVER improves planning L2 on VAD-Tiny and VAD-Base with about 36\%
less estimated total training time, including pretraining
(Table~\ref{tab:training_schedule_ablation}). The lower panels compare planning
performance and detection NDS across pretraining methods and auxiliary-head sizes.}
\label{fig:teaser}
\end{figure}

Many systems first train perception, tracking, or mapping before fine-tuning the full end-to-end model. This approach depends on task-specific annotations, heads, and decoder design \citep{hu2022stp3,hu2023planning,jiang2023vad,weng2024paradrive,doll2024dualad,tang2025hipad,li2025ssr}. PAVER instead pretrains BEV features at locations queried by candidate ego motions.

\begin{figure}[!t]
\centering
\includegraphics[width=\columnwidth]{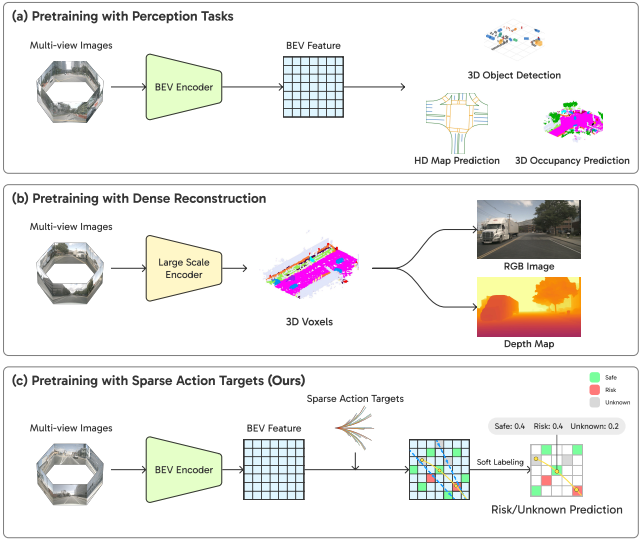}
\caption{Driving pretraining paradigms. Existing work supervises pretraining
with (a) perception labels or (b) dense reconstruction; (c) PAVER introduces
sparse action targets.}
\label{fig:task_comparison}
\end{figure}

Driving-specific pretraining methods provide relevant geometry, yet representative objectives remain centered on reconstructing scene content \citep{yang2024unipad,zou2025mim4d,zhang2025visionpad}. As summarized in Fig.~\ref{fig:task_comparison}, these methods supervise general spatial support rather than the regions most relevant to planning. Planning instead depends on which observed regions constrain candidate actions \citep{hu2023planning}. This mismatch motivates supervising the BEV regions used for planning.

We introduce \textbf{PAVER}, \emph{Planning-Aligned BEV Encoder Pretraining}, which pretrains an otherwise unchanged BEV encoder. Following occupancy-grid mapping \citep{elfes1989occupancy}, a LiDAR sweep is reduced to sparse free, occupied, and unknown evidence. Rule-based actions index supervised BEV regions, where samples across the vehicle footprint form sparse action targets for \emph{risk} and \emph{unknown} evidence. PAVER replaces these corridor features with a shared mask token and predicts the corresponding sparse action targets conditioned on the action state. This directs BEV pretraining toward evidence that constrains candidate actions.

PAVER requires neither task-specific annotations, a learned teacher, nor dense 3D scene reconstruction. The pretraining-only PAVER head adds 10K parameters and is removed before downstream training; only the BEV encoder is transferred, preserving the downstream architecture and inference pipeline. With this lightweight supervision and shorter total training time (Figure~\ref{fig:teaser}), PAVER matches or improves the corresponding baseline on most evaluated multi-task metrics (Tables~\ref{tab:training_schedule_ablation} and~\ref{tab:nuscenes_multitask}). Collision behavior nevertheless varies across architectures (Table~\ref{tab:horizon_planning}).

Our contributions are:
\begin{itemize}
    \item We formulate \emph{planning-aligned BEV encoder pretraining} with supervision at locations along candidate ego motions.
    \item We construct sparse action targets from LiDAR risk and unknown evidence and predict them with a masked, action-conditioned BEV branch, without scene reconstruction, future prediction, or a learned teacher.
    \item We transfer only the BEV encoder and discard the 10K-parameter auxiliary head. We evaluate multi-task transfer and planning performance across several architectures.
\end{itemize}

\begin{table*}[!t]
\ifdefined\arxivpreprint\PaverPageFourTopTable\fi
\centering
\scriptsize
\renewcommand{\arraystretch}{0.90}
\tabcolsep=3.0pt
\resizebox{0.90\textwidth}{!}{%
\begin{tabular}{l>{\centering\arraybackslash}p{2.4em}>{\centering\arraybackslash}p{2.4em}llcccccccc}
\toprule
& \multicolumn{2}{c}{Epochs} & \multicolumn{2}{c}{Auxiliary Tasks} & \multicolumn{2}{c}{Planning} & \multicolumn{3}{c}{Motion} & \multicolumn{2}{c}{Detection} & Map \\
\cmidrule(lr){2-3}\cmidrule(lr){4-5}\cmidrule(lr){6-7}\cmidrule(lr){8-10}\cmidrule(lr){11-12}\cmidrule(lr){13-13}
Method & PT & FT & \multicolumn{1}{c}{Pretraining} & \multicolumn{1}{c}{Fine-tuning} & L2 (m) $\downarrow$ & Col. (\%) $\downarrow$ & ADE $\downarrow$ & FDE $\downarrow$ & MR $\downarrow$ & mAP $\uparrow$ & NDS $\uparrow$ & mAP $\uparrow$ \\
\midrule
VAD-Tiny$^{\dagger}$ & 0 & 60 & None & Det., Map, Mot., Plan. & 0.66 & 0.51 & 0.91 & 1.25 & 0.13 & 0.23 & 0.34 & 0.42 \\
\rowcolor{oursgray}
\textbf{+ Ours} & \textbf{20} & \textbf{30} & Actions & Det., Map, Mot., Plan. & \textbf{0.60} & \textbf{0.19} & \textbf{0.80} & \textbf{1.09} & \textbf{0.12} & \textbf{0.28} & \textbf{0.40} & \textbf{0.44} \\
\midrule
VAD-Base$^{\dagger}$ & 0 & 60 & None & Det., Map, Mot., Plan. & 0.74 & \textbf{0.31} & 0.76 & 1.03 & 0.11 & 0.29 & 0.42 & \textbf{0.50} \\
\rowcolor{oursgray}
\textbf{+ Ours} & \textbf{20} & \textbf{30} & Actions & Det., Map, Mot., Plan. & \textbf{0.56} & 0.40 & \textbf{0.69} & \textbf{0.90} & \textbf{0.09} & \textbf{0.33} & \textbf{0.45} & 0.50 \\
\midrule
GenAD$^{\dagger}$ & 0 & 60 & None & Det., Map, Mot., Plan. & 0.59 & 0.37 & 0.87 & 1.18 & 0.14 & 0.19 & 0.26 & \textbf{0.46} \\
\rowcolor{oursgray}
\textbf{+ Ours} & \textbf{20} & \textbf{30} & Actions & Det., Map, Mot., Plan. & \textbf{0.54} & \textbf{0.21} & \textbf{0.80} & \textbf{1.05} & \textbf{0.11} & \textbf{0.21} & \textbf{0.28} & 0.44 \\
\bottomrule
\end{tabular}
}
\begin{minipage}{0.90\textwidth}
\raggedright
\scriptsize $^{\dagger}$ Reproduced results.
\end{minipage}
\caption{Multi-task performance on the nuScenes validation set. Actions denotes PAVER's sparse action targets. PT and FT denote pretraining and fine-tuning epochs; Det., Mot., and Plan.\ denote detection, motion prediction, and planning. L2 is the trajectory displacement error and Col.\ the box collision rate in percent, both averaged over 1, 2, and 3 seconds; ADE, FDE, and MR are the motion average and final displacement errors and miss rate; mAP and NDS are mean average precision and the nuScenes detection score.}
\label{tab:nuscenes_multitask}
\end{table*}

\section{Related Work}

\ifdefined\arxivpreprint
\begin{figure}[!b]
\centering
\includegraphics[width=0.937\columnwidth]{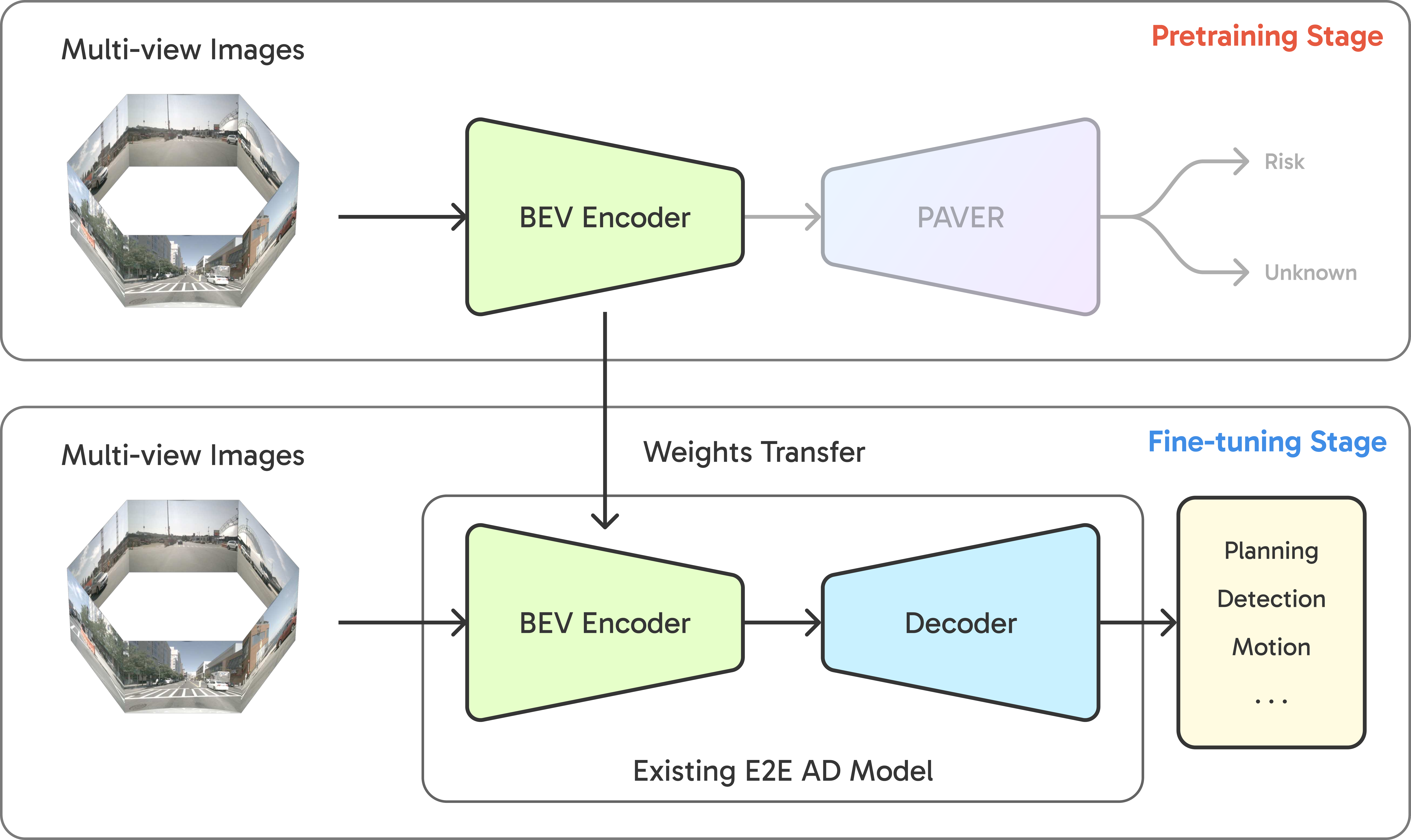}
\caption{Two-stage training protocol. Only the pretrained BEV encoder is
transferred, and the existing end-to-end model is fine-tuned unchanged.}
\label{fig:training_protocol}
\end{figure}
\fi

\begin{figure*}[!t]
\ifdefined\arxivpreprint\PaverPageThreeTopFigure\fi
\centering
\includegraphics[width=\textwidth]{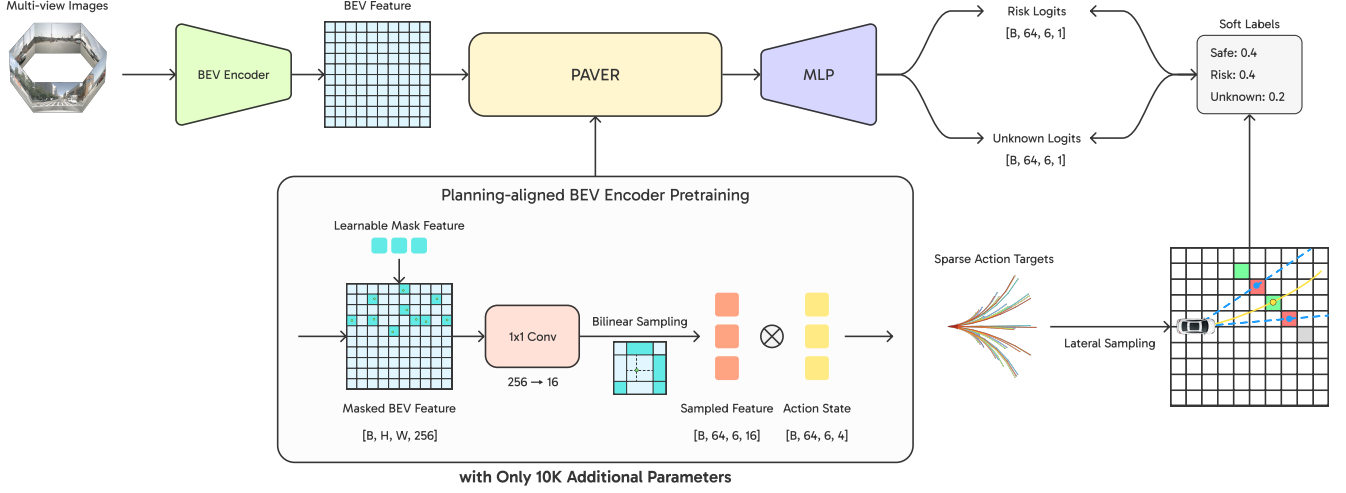}
\caption{PAVER overview.
PAVER constructs sparse action targets for risk and unknown evidence, masks their associated BEV regions, and predicts the targets from the remaining context.
The PAVER head is removed before downstream training.}
\label{fig:paver_overview}
\end{figure*}

\paragraph{Shared representations and initialization in end-to-end driving.}
End-to-end driving systems couple perception, prediction, and planning through a shared spatial representation, ranging from dense BEV grids \citep{li2022bevformer,li2022bevdepth,huang2023tpvformer} to vectorized or sparse scene elements \citep{hu2023planning,jiang2023vad,sun2024sparsedrive,zheng2024genad,li2025ssr}. Initialization is commonly obtained from downstream supervision: ST-P3, UniAD, and VAD stage perception-oriented training before the broader task stack, PARA-Drive redesigns task connectivity for joint optimization, and later systems vary the decoder itself \citep{hu2022stp3,hu2023planning,jiang2023vad,weng2024paradrive,doll2024dualad,tang2025hipad,li2025ssr}. The representation is therefore shaped by model-specific annotations and decoders. PAVER separates initialization from that optimization, targeting architectures that retain a dense, spatially indexed BEV representation and leaving their task heads unchanged.

\paragraph{Driving-specific 3D pretraining.}
Reconstruction methods learn transferable geometry from masked or projected scene content. UniPAD reconstructs continuous 3D structure and appearance, MIM4D extends reconstruction to multi-view video, and VisionPAD uses rendering and temporal consistency to learn 3D-aware features \citep{yang2024unipad,zou2025mim4d,zhang2025visionpad}. These methods lift image features into a 3D volume and supervise it densely, so the pretrained weights that transfer are the image backbone rather than the shared BEV encoder. BEV-MAE and Occupancy-MAE similarly reconstruct masked spatial structure for LiDAR representations \citep{lin2024bevmae,min2022occupancymae}. A separate line distills LiDAR knowledge into camera models through feature or label alignment \citep{chen2022bevdistill,wang2023distillbev,zhou2023unidistill,chen2024labeldistill,hinton2015distilling,vapnik2015lupi}. PAVER pretrains the BEV encoder itself, uses no learned LiDAR encoder, and performs no feature matching: deterministic free, occupied, and unknown evidence is aggregated into sparse action targets.

\paragraph{Predictive world models.}
World-model pretraining learns representations by predicting how scenes evolve. DriveWorld models spatiotemporal latent dynamics, while ViDAR predicts future geometric observations from camera history \citep{min2024driveworld,yang2024vidar}. LAW, DriveX, and DLWM connect future scene or latent prediction to planning \citep{li2025law,shi2025drivex,zhu2026dlwm}. These objectives use future observations, geometry, or learned transition targets. PAVER instead supervises current geometric evidence from a single synchronized LiDAR sweep.

\paragraph{Action and planning signals in representation learning.}
Actions have appeared in driving pretraining as labels, conditioning variables, policy targets, or guides for latent prediction \citep{zhang2022actionconditioned,xiao2021actionbased,wang2026drivejepa,chen2020learningbycheating,li2025law,zhu2026dlwm}. PAVER instead uses rule-based actions to index current LiDAR evidence, not as behavioral prediction targets.

\section{Method}
\label{sec:method}

\subsection{Overview}

PAVER pretrains the BEV encoder with sparse action targets constructed from LiDAR evidence along rule-based ego motions.
Multi-view images $I$ are processed by the same BEV encoder used for downstream end-to-end training \citep{lin2017fpn,li2022bevformer,jiang2023vad}, producing $F=f_{\mathrm{bev}}(I)\in\mathbb{R}^{B\times C\times H\times W}$.
LiDAR measurements are used only to construct pretraining targets.
Detection, mapping, motion-prediction, and planning annotations are not used, and the corresponding task decoders are disabled during pretraining.
All parameters of $f_{\mathrm{bev}}$ are optimized by the pretraining objective and transferred to downstream training (Fig.~\ref{fig:training_protocol}).
The action generator, LiDAR target builder, and the \textbf{PAVER head} (Fig.~\ref{fig:training_protocol}), which comprises the learnable mask token, pointwise projector, and prediction MLP, are used only during pretraining.

\ifdefined\arxivpreprint\else
\begin{figure}[t]
\centering
\includegraphics[width=0.937\columnwidth]{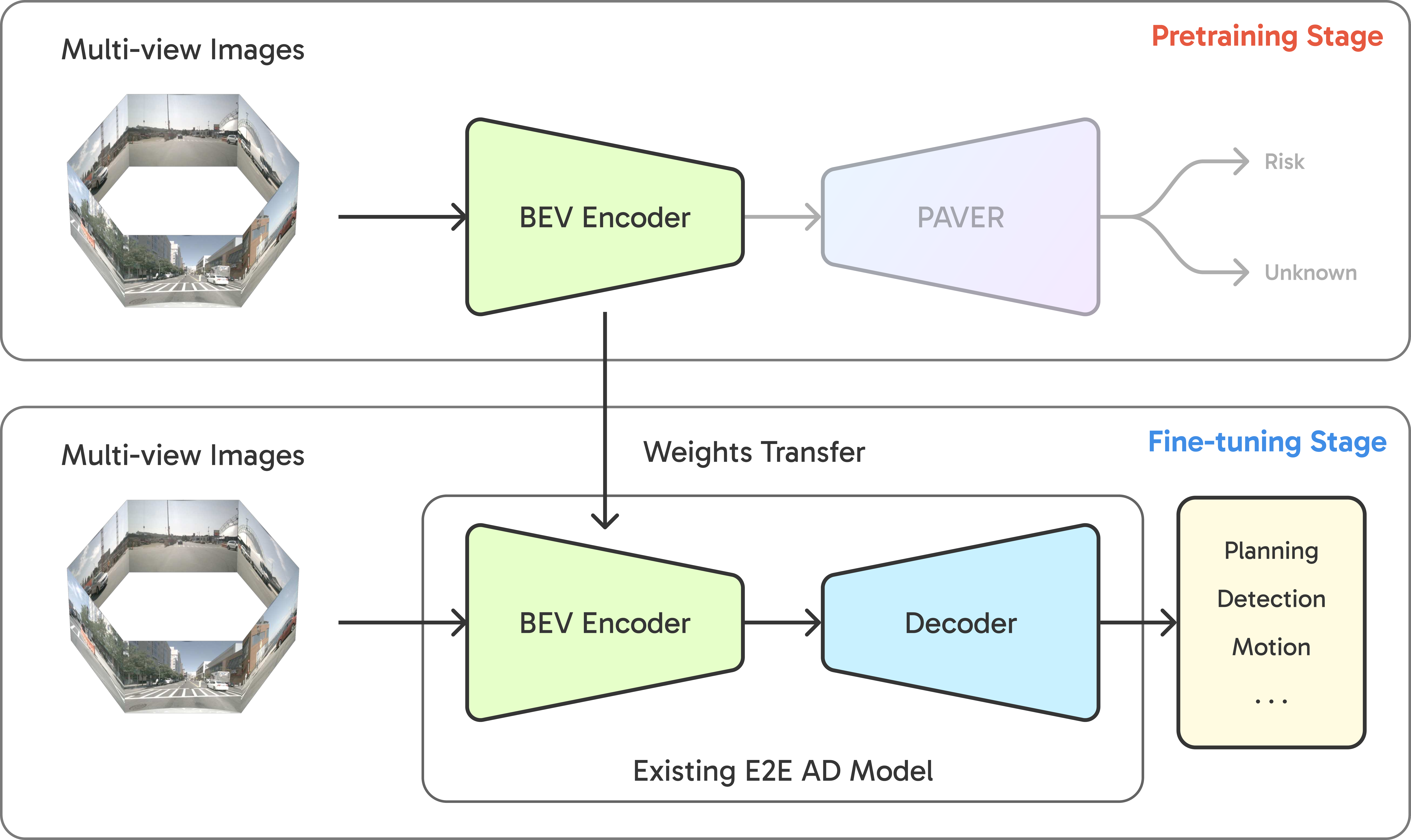}
\caption{Two-stage training protocol. Only the pretrained BEV encoder is
transferred, and the existing end-to-end model is fine-tuned unchanged.}
\label{fig:training_protocol}
\end{figure}
\fi

\subsection{Candidate Ego Motions}

Rule-based candidate motions define the supervision locations; they are not expert trajectories or learned action proposals.

Finite sets of longitudinal accelerations and yaw rates are combined into rule-based control families, tiled across action slots, and independently perturbed during training.
Each control is rolled out from the current ego speed over $T$ horizons, giving the ego-frame state $q_{s,t}^{\mathrm{ego}}=(x_{s,t}^{\mathrm{ego}},y_{s,t}^{\mathrm{ego}},\psi_{s,t}^{\mathrm{ego}},v_{s,t})$ for action slot $s$ and horizon $t$, where $\psi$ is the yaw heading.
Rollout positions and headings are transformed into the LiDAR/BEV indexing frame, yielding $\bar q_{s,t}^{\mathrm{bev}}=(\bar x_{s,t},\bar y_{s,t},\bar\psi_{s,t},v_{s,t})$.
The transformed state is used for LiDAR target construction and BEV feature sampling, while the PAVER head is conditioned on the original ego-frame state.

\ifdefined\arxivpreprint
\ifdefined\arxivpreprint
\par\addvspace{\intextsep}
\noindent\begin{minipage}{\columnwidth}
\captionsetup{type=figure}
\else
\begin{figure}[t]
\fi
\centering
\includegraphics[width=\columnwidth]{figures/targets/soft_labels5_aaai.pdf}
\caption{Sparse action-target construction. LiDAR rays provide Risk and Unknown targets for action-indexed queries; ``Safe'' denotes Free evidence.}
\label{fig:soft_labels}
\ifdefined\arxivpreprint
\end{minipage}
\par\nopagebreak[4]
\else
\end{figure}
\fi

\fi

\subsection{Sparse LiDAR Evidence}

We rasterize LiDAR measurements into a BEV evidence map $E\in\{\textsc{Unknown},\textsc{Free},\textsc{Occupied}\}^{B\times H\times W}$ following occupancy-grid mapping \citep{elfes1989occupancy} (Figure~\ref{fig:soft_labels}).
Cells start as \textsc{Unknown}; samples along each open ray segment are marked \textsc{Free} \citep{amanatides1987voxel}, and measured endpoints override their cells as \textsc{Occupied}.
This map is used only to construct sparse targets; the camera model does not predict dense occupancy.

\ifdefined\arxivpreprint\else
\ifdefined\arxivpreprint
\par\addvspace{\intextsep}
\noindent\begin{minipage}{\columnwidth}
\captionsetup{type=figure}
\else
\begin{figure}[t]
\fi
\centering
\includegraphics[width=\columnwidth]{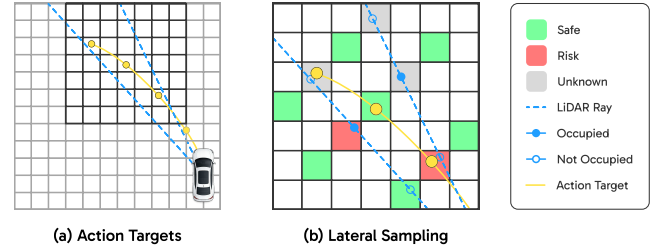}
\caption{Sparse action-target construction. LiDAR rays provide Risk and Unknown targets for action-indexed queries; ``Safe'' denotes Free evidence.}
\label{fig:soft_labels}
\ifdefined\arxivpreprint
\end{minipage}
\par\nopagebreak[4]
\else
\end{figure}
\fi

\fi

\subsection{Sparse Action Targets}

At each candidate state $\bar q_{s,t}^{\mathrm{bev}}$ we query $K$ locations laterally across the vehicle width, and $\mathcal{V}_{s,t}$ denotes the queries that stay in range.
The two targets are the fractions of those queries that fall on measured endpoints and on cells without supporting ray evidence:
\begin{equation}
r_{s,t}
=
\frac{\left|\left\{k\in\mathcal{V}_{s,t}: E_k=\textsc{Occupied}\right\}\right|}
     {\left|\mathcal{V}_{s,t}\right|},
\label{eq:risk_target}
\end{equation}
\begin{equation}
u_{s,t}
=
\frac{\left|\left\{k\in\mathcal{V}_{s,t}: E_k=\textsc{Unknown}\right\}\right|}
     {\left|\mathcal{V}_{s,t}\right|},
\label{eq:unknown_target}
\end{equation}
where $E_k$ is the evidence value at the $k$-th query position.

Risk measures observed occupancy, while unknown measures missing observations.
These continuous ratios describe current evidence, not future collision probabilities or calibrated uncertainty.
Candidate motions index the supervision locations; they are not prediction targets.
This \emph{planning-aligned} supervision places targets along rule-based ego motions rather than densely across the BEV raster.

\ifdefined\arxivpreprint
\par\addvspace{\intextsep}
\noindent\begin{minipage}{\columnwidth}
\centering
\captionsetup{type=figure}
\includegraphics[width=\columnwidth]{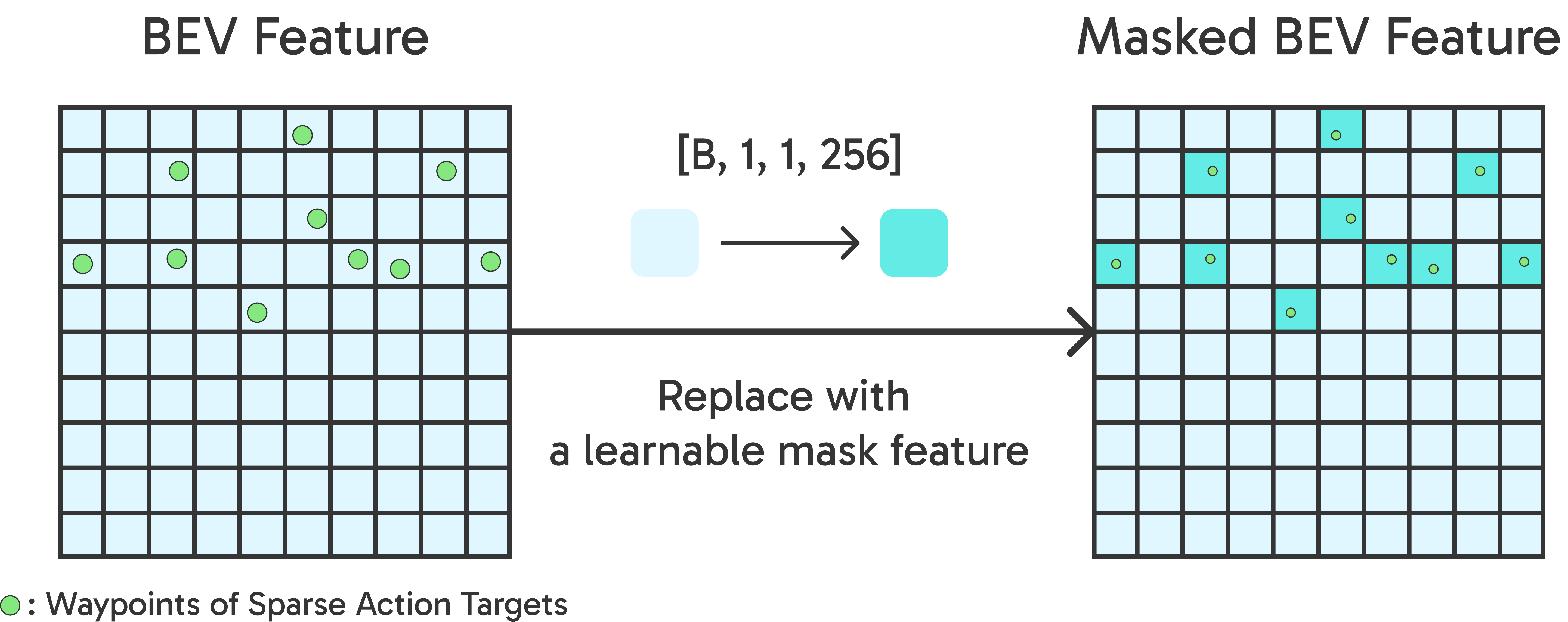}
\caption{Action-corridor feature masking. Green markers indicate lateral-query
locations. Selected BEV cells are replaced by a shared learnable token;
other cells are unchanged.}
\label{fig:action_corridor_masking}
\end{minipage}
\par\nopagebreak[4]
\fi

\subsection{Action-Corridor Feature Masking}

To limit direct access to target-associated features, we form a binary mask $M\in\{0,1\}^{B\times1\times H\times W}$ from the union of rasterized lateral-query cells.
Selected cells in a temporary feature map are replaced by a shared learnable token $m\in\mathbb{R}^{1\times C\times1\times1}$:
\begin{equation}
F^{\mathrm{mask}}=(1-M)\odot F+M\odot m.
\label{eq:feature_mask}
\end{equation}
Unselected cells retain their original features and the original representation $F$ is left unchanged.
Figure~\ref{fig:action_corridor_masking} illustrates the feature replacement.

\ifdefined\arxivpreprint\else
\begin{figure}[t]
\centering
\includegraphics[width=\columnwidth]{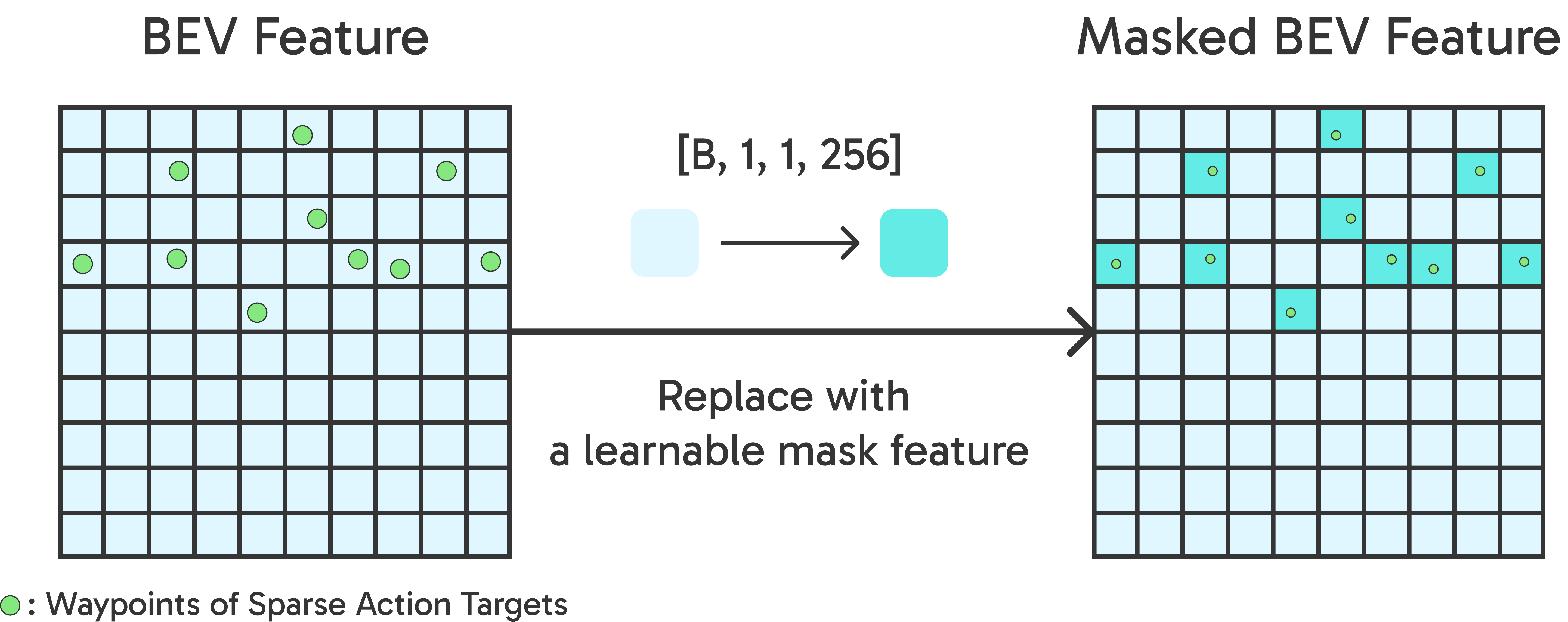}
\caption{Action-corridor feature masking. Green markers indicate lateral-query
locations. Selected BEV cells are replaced by a shared learnable token;
other cells are unchanged.}
\label{fig:action_corridor_masking}
\end{figure}
\fi

Masking removes the selected local feature vectors, but not information encoded before masking, and does not guarantee the use of neighboring cells.

\subsection{Action-Conditioned Prediction}

A pointwise $1\times1$ projector $\phi$ produces $Z=\phi(F^{\mathrm{mask}})$ without spatial aggregation.
For each action slot and horizon, we bilinearly sample $z_{s,t}$ at the transformed action center $(\bar x_{s,t},\bar y_{s,t})$ \citep{jaderberg2015stn}, concatenate the ego-frame action state, and predict two logits with the MLP $g$:
\begin{equation}
(\hat r_{s,t},\hat u_{s,t})=g\!\left(\left[z_{s,t};q_{s,t}^{\mathrm{ego}}\right]\right).
\label{eq:target_prediction}
\end{equation}
The action-center cell is masked because the lateral-offset set includes zero.
The head predicts one risk/unknown pair from a center feature, while each target pair aggregates $K$ lateral LiDAR queries.

\subsection{Pretraining Objective}

The pretraining objective is
\begin{equation}
\mathcal{L}_{\mathrm{pre}}=\lambda_r\mathcal{L}_{r}+\lambda_u\mathcal{L}_{u},
\label{eq:pretraining_objective}
\end{equation}
where $\mathcal{L}_{r}$ and $\mathcal{L}_{u}$ are logit-based binary cross-entropy losses for the continuous risk and unknown targets, averaged over valid action--horizon entries.

\ifdefined\arxivpreprint\else

\fi
Figure~\ref{fig:paver_overview} summarizes pretraining.
Only the pretrained BEV encoder is transferred; the PAVER head and target-generation components are discarded.
The downstream model, training objectives, and camera-only inference are unchanged.

The Technical Supplement provides rollout equations, coordinate conventions, target-construction details, and training settings.

\newlength{\gradcamcellwidth}
\gradcamcellwidth=0.3334\columnwidth
\newcommand{\gradCamHeaderThree}[3]{%
  \makebox[\columnwidth][c]{%
    \begin{minipage}[c]{\gradcamcellwidth}\centering\footnotesize\bfseries #1\end{minipage}%
    \begin{minipage}[c]{\gradcamcellwidth}\centering\footnotesize\bfseries #2\end{minipage}%
    \begin{minipage}[c]{\gradcamcellwidth}\centering\footnotesize\bfseries #3\end{minipage}}%
  \par\vspace{0.6pt}%
}
\newcommand{\gradCamRowThree}[3]{%
  \makebox[\columnwidth][c]{%
    \includegraphics[width=\gradcamcellwidth]{#1}%
    \includegraphics[width=\gradcamcellwidth]{#2}%
    \includegraphics[width=\gradcamcellwidth]{#3}}%
  \par%
}

\section{Experiments}

\label{sec:experiments}

\subsection{Implementation Details}

We pretrain on the 28,130 nuScenes training keyframes and evaluate on all 6,019
validation keyframes, using VAD-Tiny \citep{jiang2023vad} unless stated
otherwise. PAVER uses a single-frame temporal queue and LiDAR only for target
construction.

We pretrain for 20 epochs on 4 GPUs with AdamW \citep{loshchilov2019adamw},
cosine annealing \citep{loshchilov2017sgdr}, and a reduced image-backbone
learning rate. Downstream training follows each model's original recipe, with
the transferred BEV encoder and reinitialized task decoders. All metrics in
each row come from a single checkpoint selected for joint multi-task balance
within its training budget. Additional settings are in the Technical Supplement.

We evaluate the transferred models on nuScenes \citep{caesar2020nuscenes} using
the planning, motion-prediction, detection, and vector-map metrics adopted by
VAD \citep{jiang2023vad}. Table~\ref{tab:nuscenes_multitask} averages planning
over 1, 2, and 3 seconds; per-horizon values are reported separately.

\ifdefined\arxivpreprint
\par\addvspace{\intextsep}
\noindent\begin{minipage}{\columnwidth}
\captionsetup{type=table}
\else
\begin{table}[!htbp]
\fi
\centering
\normalsize
\renewcommand{\arraystretch}{1.10}
\tabcolsep=3.8pt
\resizebox{\columnwidth}{!}{%
\begin{tabular}{lccccccc}
\toprule
& \multicolumn{2}{c}{Planning} & \multicolumn{3}{c}{Motion} & {Detection} & {Map} \\
\cmidrule(lr){2-3}
\cmidrule(lr){4-6}
\cmidrule(lr){7-7}
\cmidrule(lr){8-8}
Target Source & L2 (m) $\downarrow$ & Col. (\%) $\downarrow$ & ADE $\downarrow$ & FDE $\downarrow$ & MR $\downarrow$ & NDS $\uparrow$ & mAP $\uparrow$ \\
\midrule
None$^{\dagger}$ & 0.662 & 0.513 & 0.905 & 1.250 & 0.135 & 0.338 & 0.419 \\
Pseudo-LiDAR & \textbf{0.563} & 0.560 & 0.831 & 1.101 & \textbf{0.121} & 0.386 & \textbf{0.439} \\
\rowcolor{oursgray}
LiDAR & 0.603 & \textbf{0.187} & \textbf{0.803} & \textbf{1.086} & 0.121 & \textbf{0.397} & 0.439 \\
\bottomrule
\end{tabular}
}
\begin{minipage}{\columnwidth}
\raggedright
\scriptsize $^{\dagger}$ Reproduced results.
\end{minipage}
\caption{Planning results with pseudo-LiDAR pretraining. Planning and motion metrics are averaged over 1, 2, and 3 seconds. None denotes the VAD-Tiny baseline.}
\label{tab:pseudo_lidar}
\ifdefined\arxivpreprint
\end{minipage}
\par\nopagebreak[4]\addvspace{\intextsep}
\else
\end{table}
\fi

\subsection{Results with Pseudo-LiDAR}

When LiDAR is unavailable, PAVER constructs targets by unprojecting per-camera
depth \citep{wang2019pseudolidar} from frozen UniK3D
\citep{piccinelli2025unik3d}. Pseudo-LiDAR improves planning L2, motion,
detection, and mapping over the baseline (Table~\ref{tab:pseudo_lidar}).
Its L2 is lower than with measured LiDAR, but collision rate exceeds both
the baseline and measured-LiDAR variants. Estimated geometry therefore
supports multi-task transfer without reproducing measured LiDAR's collision
reduction. Depth-quality analyses are provided in the Technical Supplement.

\ifdefined\arxivpreprint
\subsection{Quantitative Results on nuScenes}

PAVER matches or improves most downstream metrics on three architectures
without task annotations during pretraining (Table~\ref{tab:nuscenes_multitask}).
On VAD-Tiny and GenAD, detection gains are larger for pedestrians, motorcycles,
and bicycles than for cars (Table~\ref{tab:nuscenes_detection_classes}).
\fi

\begin{table}[h]
\centering
\small
\tabcolsep=3.0pt
\sbox{\perhorizontablebox}{%
\begin{tabular}{lccccccc}
\toprule
& \multicolumn{6}{c}{AP $\uparrow$} & \\
\cmidrule(lr){2-7}
Method & Car & Truck & Bus & Ped. & Motor. & Bike & mAP \\
\midrule
VAD-Tiny$^{\dagger}$ & 0.49 & 0.19 & 0.28 & 0.28 & 0.13 & 0.17 & 0.23 \\
\rowcolor{oursgray}
\textbf{+ Ours} & \textbf{0.51} & \textbf{0.24} & \textbf{0.41} & \textbf{0.35} & \textbf{0.22} & \textbf{0.23} & \textbf{0.28} \\
\midrule
VAD-Base$^{\dagger}$ & 0.55 & 0.24 & 0.33 & 0.38 & 0.24 & 0.27 & 0.29 \\
\rowcolor{oursgray}
\textbf{+ Ours} & \textbf{0.55} & \textbf{0.24} & \textbf{0.46} & \textbf{0.42} & \textbf{0.31} & \textbf{0.29} & \textbf{0.33} \\
\midrule
GenAD$^{\dagger}$ & 0.44 & -- & -- & 0.29 & 0.17 & 0.22 & 0.19 \\
\rowcolor{oursgray}
\textbf{+ Ours} & \textbf{0.46} & -- & -- & \textbf{0.34} & \textbf{0.23} & \textbf{0.25} & \textbf{0.21} \\
\bottomrule
\end{tabular}%
}

\begin{minipage}{\wd\perhorizontablebox}
\centering
\usebox{\perhorizontablebox}

\par
{\raggedright
\scriptsize $^{\dagger}$ Reproduced results.
\par}
\end{minipage}
\caption{Detection results per object class on nuScenes. Ped., Motor.\ and Bike
denote pedestrian, motorcycle and bicycle. A dash marks a class the reproduced
GenAD never detects; the mean still counts it as zero.}
\label{tab:nuscenes_detection_classes}
\end{table}

\begin{figure*}[!t]
\centering
\begingroup
\def\inputcol{0.470587\textwidth}
\def\bevcol{0.176471\textwidth}
\noindent
\makebox[\inputcol][c]{\footnotesize\bfseries Input Images}%
\makebox[\bevcol][c]{\footnotesize\bfseries Ground Truth}%
\makebox[\bevcol][c]{\footnotesize\bfseries VAD-Tiny}%
\makebox[\bevcol][c]{\footnotesize\bfseries + PAVER}%
\par\nointerlineskip\vspace{2pt}
\noindent
\includegraphics[width=\inputcol]{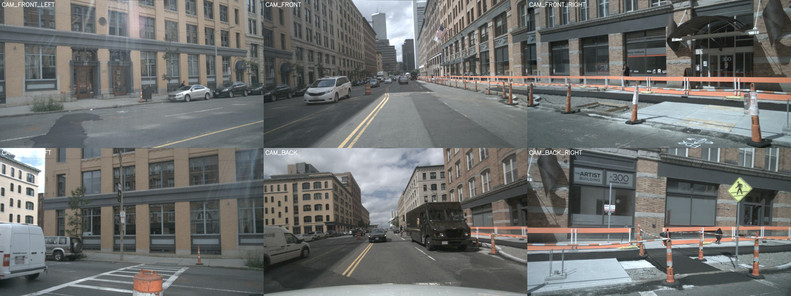}%
\includegraphics[width=\bevcol]{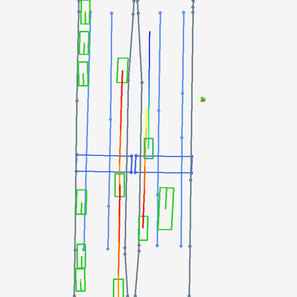}%
\includegraphics[width=\bevcol]{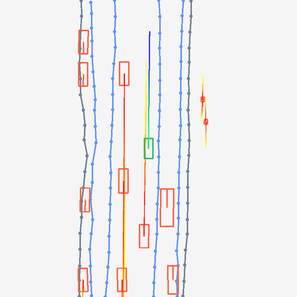}%
\includegraphics[width=\bevcol]{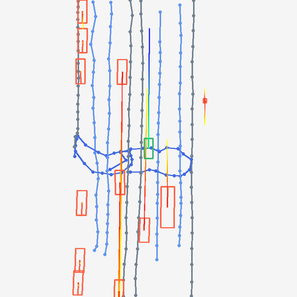}%
\par\nointerlineskip\vspace{1.2pt}
\noindent
\includegraphics[width=\inputcol]{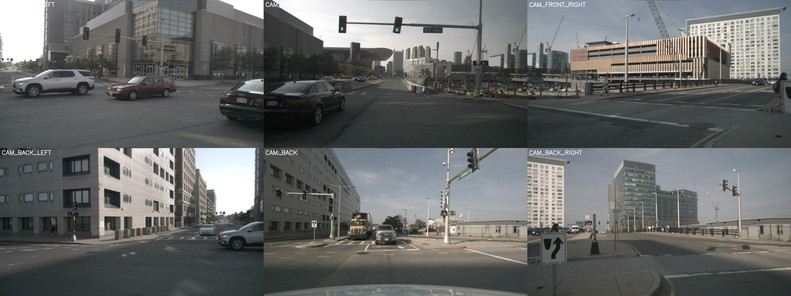}%
\includegraphics[width=\bevcol]{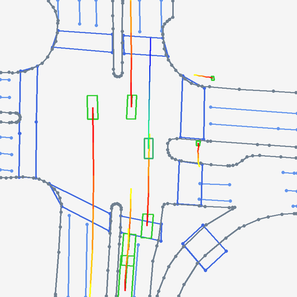}%
\includegraphics[width=\bevcol]{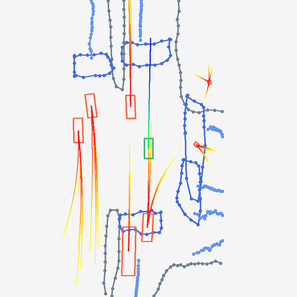}%
\includegraphics[width=\bevcol]{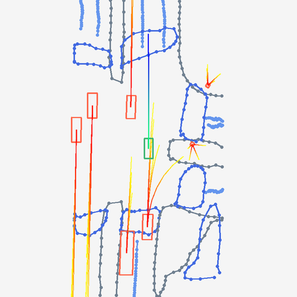}%
\par
\endgroup
\caption{VAD-Tiny qualitative comparisons on nuScenes. More results are in the
Technical Supplement.}
\label{fig:cross_model_qualitative_main}
\end{figure*}

\ifdefined\arxivpreprint\else
\subsection{Quantitative Results on nuScenes}

PAVER matches or improves most downstream metrics on three architectures
without task annotations during pretraining (Table~\ref{tab:nuscenes_multitask}).
On VAD-Tiny and GenAD, detection gains are larger for pedestrians, motorcycles,
and bicycles than for cars (Table~\ref{tab:nuscenes_detection_classes}).
\fi
\ifdefined\arxivpreprint

On VAD-Tiny, collision rate falls from 0.51 to 0.19\%, planning L2 from 0.66
to 0.60 m, and motion ADE from 0.91 to 0.80 m, while map mAP rises from 0.42
to 0.44. GenAD reduces collision from 0.37 to 0.21\% and L2 by 0.05 m.
VAD-Base reduces L2 by 0.18 m but increases collision from 0.31 to 0.40\%.
\fi

\begin{table}[t]
\centering
\small
\tabcolsep=3.0pt

\sbox{\perhorizontablebox}{%
\begin{tabular}{lcccccc}
\toprule
& \multicolumn{3}{c}{L2 (m) $\downarrow$}
& \multicolumn{3}{c}{Col. (\%) $\downarrow$} \\
\cmidrule(r){2-4}
\cmidrule(l){5-7}
Method & 1s & 2s & 3s & 1s & 2s & 3s \\
\midrule
VAD-Tiny$^{\dagger}$
& 0.35 & 0.63 & 1.01
& 0.38 & 0.45 & 0.71 \\

\rowcolor{oursgray}
\textbf{+ Ours}
& \textbf{0.32} & \textbf{0.58} & \textbf{0.92}
& \textbf{0.09} & \textbf{0.15} & \textbf{0.32} \\

\midrule
VAD-Base$^{\dagger}$
& 0.41 & 0.71 & 1.10
& \textbf{0.20} & \textbf{0.29} & \textbf{0.43} \\

\rowcolor{oursgray}
\textbf{+ Ours}
& \textbf{0.30} & \textbf{0.54} & \textbf{0.85}
& 0.24 & 0.41 & 0.55 \\

\midrule
GenAD$^{\dagger}$
& 0.33 & 0.56 & 0.88
& 0.21 & 0.34 & 0.56 \\

\rowcolor{oursgray}
\textbf{+ Ours}
& \textbf{0.28} & \textbf{0.51} & \textbf{0.82}
& \textbf{0.09} & \textbf{0.18} & \textbf{0.35} \\
\bottomrule
\end{tabular}%
}

\begin{minipage}{\wd\perhorizontablebox}
\centering
\usebox{\perhorizontablebox}

\par
{\raggedright
\small
$^{\dagger}$ Reproduced results.
\par}
\end{minipage}

\caption{Planning results by horizon on nuScenes, reported at the 1, 2, and 3 second horizons.}
\label{tab:horizon_planning}
\end{table}

VAD-Tiny and GenAD improve L2 and collision at every horizon, whereas VAD-Base
improves L2 but increases collision (Table~\ref{tab:horizon_planning}).

Compared with reconstruction-based pretraining, PAVER achieves lower L2 at
every VAD-Tiny horizon with 0.01M auxiliary parameters, compared with 6.41M
and 13.43M (Table~\ref{tab:pretraining_planning}).
Grad-CAM highlights vehicles and pedestrians (Figure~\ref{fig:gradcam_main}),
while Figure~\ref{fig:cross_model_qualitative_main} compares downstream
predictions. Additional results are in the Technical Supplement.

\newcommand{\paverPreparePretrainingPlanningTable}{%
\ifdefined\arxivpreprint
\newsavebox{\paverPretrainingPlanningBox}
\begin{lrbox}{\paverPretrainingPlanningBox}
\begin{minipage}{\columnwidth}
\captionsetup{type=table}
\else
\begin{table}[h]
\fi
\centering
\scriptsize
\renewcommand{\arraystretch}{1.02}
\tabcolsep=2.5pt

\resizebox{0.94\columnwidth}{!}{%
\begin{tabular}{llrcccccc}
\toprule
& & &
\multicolumn{3}{c}{L2 (m) $\downarrow$} &
\multicolumn{3}{c}{Col. (\%) $\downarrow$} \\
\cmidrule(lr){4-6}
\cmidrule(lr){7-9}
Method & Target & \multicolumn{1}{c}{\#Params} & 1s & 2s & 3s & 1s & 2s & 3s \\
\midrule
VAD-Tiny$^{\dagger}$
& --
& 0.00M
& 0.35 & 0.63 & 1.01
& 0.38 & 0.45 & 0.71 \\

+ UniPAD$^{\ddagger}$
& RGB, Depth
& 6.41M
& 0.45 & 0.75 & 1.11
& 0.19 & 0.31 & 0.46 \\

+ MIM4D$^{\ddagger}$
& RGB, Depth
& 13.43M
& 0.39 & 0.69 & 1.06
& 0.21 & 0.28 & 0.38 \\

\rowcolor{oursgray}
\textbf{+ Ours}
& Sparse Actions
& 0.01M
& \textbf{0.32} & \textbf{0.58} & \textbf{0.92}
& \textbf{0.09} & \textbf{0.15} & \textbf{0.32} \\

\midrule
VAD-Base$^{\dagger}$
& --
& 0.00M
& 0.41 & 0.71 & 1.10
& 0.20 & 0.29 & 0.43 \\

+ MIM4D$^{\ddagger}$
& RGB, Depth
& 13.43M
& 0.36 & 0.64 & 1.00
& \textbf{0.08} & \textbf{0.14} & \textbf{0.36} \\

\rowcolor{oursgray}
\textbf{+ Ours}
& Sparse Actions
& 0.01M
& \textbf{0.30} & \textbf{0.54} & \textbf{0.85}
& 0.24 & 0.41 & 0.55 \\
\bottomrule
\end{tabular}%
}

\begin{minipage}{0.94\columnwidth}
\raggedright
\scriptsize $^{\dagger}$ Reproduced results. $^{\ddagger}$ Reported results.
\end{minipage}

\caption{Planning performance under different pretraining methods.
RGB denotes camera-color reconstruction, and Depth denotes
LiDAR-projected sparse metric-depth reconstruction. Sparse Actions denotes
PAVER's sparse action targets. \#Params reports temporary auxiliary parameters.}
\label{tab:pretraining_planning}
\ifdefined\arxivpreprint
\end{minipage}
\end{lrbox}
\else
\end{table}
\fi
}
\ifdefined\arxivpreprint\else
\paverPreparePretrainingPlanningTable
\fi

\newcommand{\paverClosedLoopTable}{%
\ifdefined\arxivpreprint
\noindent\begin{minipage}{\columnwidth}
\captionsetup{type=table}
\else
\begin{table}[t]
\fi
\centering
\small
\tabcolsep=6.0pt
\begin{tabular}{lccc}
\toprule
Method & DS $\uparrow$ & RC $\uparrow$ & IS $\uparrow$ \\
\midrule
UniAD-Tiny & 48.45 & 60.96 & 0.85 \\
\rowcolor{oursgray}
\textbf{+ Ours} & \textbf{58.79} & \textbf{79.06} & 0.79 \\
\midrule
GenAD$^{\ddagger}$ & 34.53 & \na & \na \\
\rowcolor{oursgray}
\textbf{+ Ours} & \textbf{49.07} & 58.95 & 0.90 \\
\bottomrule
\multicolumn{4}{@{}l}{\scriptsize $^{\ddagger}$ Reported by GenAD.} \\
\end{tabular}
\caption{Closed-loop results on Bench2Drive Town05 Long. DS, RC, and IS denote
Driving Score, Route Completion, and Infraction Score, respectively; a dash
marks a metric the corresponding source does not report. Closed-loop
qualitative results are in the Technical Supplement.}
\label{tab:town05_closed_loop}
\ifdefined\arxivpreprint
\end{minipage}
\else
\end{table}
\fi
}

\ifdefined\arxivpreprint
\par\addvspace{\intextsep}
\paverClosedLoopTable
\par\nopagebreak[4]
\fi

\subsection{Closed-Loop Evaluation}

We evaluate closed-loop transfer on nine Bench2Drive Town05 Long routes
\citep{jia2024bench2drive}, with one repetition per route
(Table~\ref{tab:town05_closed_loop}). On UniAD-Tiny, Driving Score rises from
48.45 to 58.79 and Route Completion from 60.96 to 79.06, while Infraction
Score decreases from 0.85 to 0.79.
At the route level, PAVER initialization converts three UniAD-Tiny timeouts
into completions. Detailed per-route outcomes, completion timelines, and
frame-by-frame qualitative comparisons are provided in the Technical Supplement.

\ifdefined\arxivpreprint\else
\paverClosedLoopTable
\fi

\ifdefined\arxivpreprint
\paverPreparePretrainingPlanningTable
\fi

\ifdefined\arxivpreprint
\par\newpage
\noindent\begin{minipage}{\columnwidth}
\usebox{\paverPretrainingPlanningBox}
\par
\fi
\section{Ablation Study}
\label{sec:analysis}
\ifdefined\arxivpreprint
\end{minipage}
\par
\fi

\subsection{Comparison of Pretraining Supervision}

\begin{table}[t]
\centering
\scriptsize
\renewcommand{\arraystretch}{1.04}
\tabcolsep=1.4pt

\resizebox{\columnwidth}{!}{%
\begin{tabular}{cccccccccc}
\toprule
\multicolumn{4}{c}{\scriptsize Pretraining} & &
\multicolumn{2}{c}{\scriptsize Plan.} &
{\scriptsize Mot.} & {\scriptsize Det.} & {\scriptsize Map} \\
\cmidrule(lr){1-4}
\cmidrule(lr){6-7}
\cmidrule(lr){8-8}
\cmidrule(lr){9-9}
\cmidrule(lr){10-10}
Det. & Map & Occ.$^{\dagger}$ & Act. &
\#Params $\downarrow$ & L2 (m) $\downarrow$ & Col. (\%) $\downarrow$ & ADE $\downarrow$ & NDS $\uparrow$ & mAP $\uparrow$ \\
\midrule
& & & & 0.00M &
0.662 & 0.513 & 0.905 & 0.338 & 0.419 \\
\midrule
$\checkmark$ & & & & 3.01M &
0.515 & 0.240 & \underline{0.798} & 0.365 & 0.406 \\
& $\checkmark$ & & & 2.91M &
0.694 & 0.333 & 0.857 & 0.331 & 0.391 \\
& & $\checkmark$ & & 3.56M &
0.646 & 0.320 & 0.905 & 0.336 & 0.393 \\
$\checkmark$ & $\checkmark$ & & & 5.92M &
\textbf{0.504} & \underline{0.190} & 0.805 & 0.361 & 0.397 \\
$\checkmark$ & $\checkmark$ & $\checkmark$ & & 9.49M &
0.577 & 0.270 & \underline{0.798} & 0.356 & 0.420 \\
\midrule
\rowcolor{oursgray}
& & & $\checkmark$ & \textbf{0.01M} &
0.603 & \textbf{0.187} & 0.803 & \textbf{0.397} & \underline{0.439} \\
\rowcolor{oursgray}
& & & $\checkmark$ & \underline{0.03M} &
\underline{0.514} & 0.320 & 0.800 & 0.383 & \textbf{0.447} \\
\rowcolor{oursgray}
& & & $\checkmark$ & 0.09M &
0.596 & 0.220 & \textbf{0.782} & \underline{0.397} & 0.433 \\
\bottomrule
\end{tabular}%
}

\vspace{1pt}
{\raggedright
\scriptsize
$^{\dagger}$ Independently implemented.
\par}

\caption{Pretraining strategy ablation.
All pretraining runs use 20 epochs.
L2 and collision rate are averaged over 1, 2, and 3 seconds.
\#Params denotes the number of auxiliary parameters. Occ. and Act. denote
occupancy and PAVER's action conditioning. Bold and underline denote the best
and second-best results, respectively, among the pretraining runs.}
\label{tab:pretraining_strategy_ablation}
\end{table}

Detection-and-map pretraining gives the lowest planning L2, while PAVER
achieves lower collision and higher NDS than the task-supervised alternatives
(Table~\ref{tab:pretraining_strategy_ablation}).
The 0.03M variant widens the projector and action-conditioned MLP; the 0.09M
variant replaces the pointwise projector with a $3\times3$ projector.
Larger heads improve some metrics but increase collision rate.
The Technical Supplement compares downstream learning curves and multi-task
transfer across pretraining objectives.

\subsection{Feature Masking and Action State}

Variants differ only in the ablated component: dropping the concatenated $[x,y,\psi,v]$ input, or sampling the original BEV features at the same action centers.

Masking or action conditioning alone gives collision rates of 0.523\% and
0.550\%, compared with 0.513\% for the baseline. Combining them reduces
collision to 0.187\% and gives the lowest L2 and highest NDS among these
variants (Table~\ref{tab:paver_mask_action_state_ablation}). Motion and map
metrics favor the partial variants.

\begin{figure}[t]
\centering
\gradCamHeaderThree{Input Image}{VAD-Tiny}{+ PAVER}
\gradCamRowThree{}{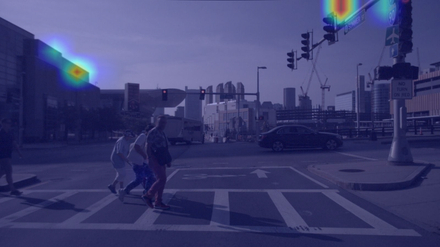}{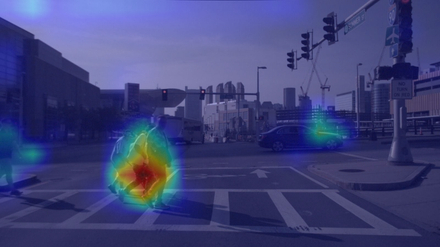}
\nointerlineskip\vspace{1.2pt}
\gradCamRowThree{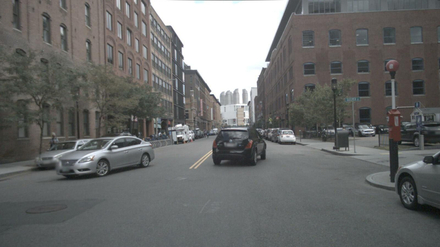}{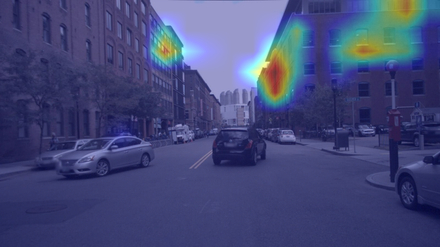}{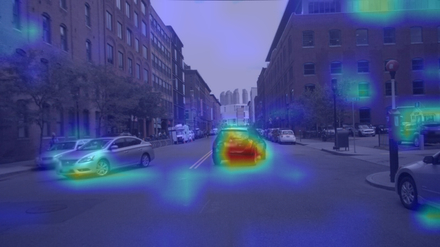}
\caption{Grad-CAM for the BEV target on nuScenes. With PAVER initialization,
activations highlight scene objects, including vehicles and pedestrians.
More results are provided in the Technical Supplement.}
\label{fig:gradcam_main}
\end{figure}

Figure~\ref{fig:gradcam_main} shows image-space attributions.
The Technical Supplement provides analyses of target coverage, frozen probes
by BEV region, and layer-wise representation similarity before and after
pretraining.

\begin{table}[!ht]
\centering
\scriptsize
\renewcommand{\arraystretch}{1.04}
\tabcolsep=2.0pt

\resizebox{\columnwidth}{!}{%
\begin{tabular}{ccccccccc}
\toprule
\multicolumn{2}{c}{} &
\multicolumn{2}{c}{Planning} &
\multicolumn{3}{c}{Motion} &
Detection &
Map \\
\cmidrule(lr){3-4}
\cmidrule(lr){5-7}
\cmidrule(lr){8-8}
\cmidrule(lr){9-9}
Mask & Action &
L2 (m) $\downarrow$ & Col. (\%) $\downarrow$ &
ADE $\downarrow$ & FDE $\downarrow$ & MR $\downarrow$ &
NDS $\uparrow$ & mAP $\uparrow$ \\
\midrule
& &
0.662 & 0.513 &
0.905 & 1.250 & 0.135 &
0.338 & 0.419 \\
\midrule
$\checkmark$ & &
0.655 & 0.523 &
0.808 & 1.084 & \textbf{0.115} &
0.396 & 0.443 \\

& $\checkmark$ &
0.705 & 0.550 &
\textbf{0.797} & \textbf{1.080} & 0.123 &
0.388 & \textbf{0.445} \\

\rowcolor{oursgray}
$\checkmark$ & $\checkmark$ &
\textbf{0.603} & \textbf{0.187} &
0.803 & 1.086 & 0.121 &
\textbf{0.397} & 0.439 \\
\bottomrule
\end{tabular}%
}

\caption{Component ablation.
Mask denotes action-corridor feature masking with a shared learnable token, and
Action denotes conditioning the prediction head on the ego-frame action state
$[x,y,\psi,v]$, where $\psi$ is the yaw heading. The variant using both is
PAVER; the empty row is the VAD-Tiny baseline.}
\label{tab:paver_mask_action_state_ablation}
\end{table}

\subsection{Pretraining and Fine-Tuning Schedules}

\begin{table}[t]
\centering
\scriptsize
\renewcommand{\arraystretch}{1.04}
\tabcolsep=2.0pt

\resizebox{\columnwidth}{!}{%
\begin{tabular}{lcccccccc}
\toprule
&
\multicolumn{3}{c}{\scriptsize Training Cost} &
\multicolumn{2}{c}{\scriptsize Planning} &
{\scriptsize Motion} &
{\scriptsize Detection} &
{\scriptsize Map} \\
\cmidrule(lr){2-4}
\cmidrule(lr){5-6}
\cmidrule(lr){7-7}
\cmidrule(lr){8-8}
\cmidrule(lr){9-9}
Method & PT & FT & Time &
L2 (m) $\downarrow$ & Col. (\%) $\downarrow$ &
ADE $\downarrow$ & NDS $\uparrow$ & mAP $\uparrow$ \\
\midrule
VAD-Tiny$^{\dagger}$
& 48$^{\ddagger}$ & 12 & \multicolumn{1}{r}{20.5h}
& 0.701 & 0.730 & 0.840 & 0.363 & 0.413 \\

VAD-Tiny$^{\dagger}$
& 0 & 60 & \multicolumn{1}{r}{21.3h}
& 0.662 & 0.513 & 0.905 & 0.338 & 0.419 \\

\rowcolor{oursgray}
\textbf{+ Ours}
& 20 & 10 & \multicolumn{1}{r}{6.5h}
& \underline{0.637} & 0.300
& 0.851 & 0.357 & 0.361 \\

\rowcolor{oursgray}
\textbf{+ Ours}
& 20 & 20 & \multicolumn{1}{r}{10.1h}
& 0.647 & \underline{0.220}
& \textbf{0.781} & \underline{0.387} & \underline{0.425} \\

\rowcolor{oursgray}
\textbf{+ Ours}
& 20 & 30 & \multicolumn{1}{r}{13.6h}
& \textbf{0.603} & \textbf{0.187}
& \underline{0.803} & \textbf{0.397} & \textbf{0.439} \\

\midrule
VAD-Base$^{\dagger}$
& 0 & 60 & \multicolumn{1}{r}{103h}
& 0.742 & \underline{0.307}
& 0.760 & 0.422 & \textbf{0.502} \\

\rowcolor{oursgray}
\textbf{+ Ours}
& 20 & 10 & \multicolumn{1}{r}{31.8h}
& \underline{0.646} & 0.530
& 0.772 & 0.371 & 0.403 \\

\rowcolor{oursgray}
\textbf{+ Ours}
& 20 & 20 & \multicolumn{1}{r}{49.0h}
& 0.757 & \textbf{0.280}
& \underline{0.698} & \underline{0.442} & 0.460 \\

\rowcolor{oursgray}
\textbf{+ Ours}
& 20 & 30 & \multicolumn{1}{r}{66.1h}
& \textbf{0.562} & 0.401
& \textbf{0.685} & \textbf{0.446} & \underline{0.498} \\
\bottomrule
\end{tabular}%
}

\begin{minipage}{\columnwidth}
\raggedright
\scriptsize
$^{\dagger}$ Reproduced results.
\\
$^{\ddagger}$ Stage 1 pretrains detection, mapping, and motion prediction without planning.
\par
\end{minipage}

\caption{Training schedule ablation. Total training time, including
pretraining, is estimated from mean iteration times on 4 NVIDIA RTX 5090 GPUs.
Bold marks the best result and underline marks the second-best result.}
\label{tab:training_schedule_ablation}
\end{table}

Twenty pretraining epochs followed by 30 fine-tuning epochs reduce estimated
total training time by about 36\%: from 21.3 to 13.6 hours for VAD-Tiny and
from 103 to 66.1 hours for VAD-Base, with the lowest L2 in each group
(Table~\ref{tab:training_schedule_ablation}). VAD-Base's collision rate is lowest
with 20 fine-tuning epochs, while its scratch baseline retains the highest map mAP.

\section{Discussion and Limitations}

\paragraph{Multi-task performance.}
PAVER improves planning L2, motion prediction, and detection on all three
nuScenes backbones; VAD-Tiny also improves mapping
(Table~\ref{tab:nuscenes_multitask}). On VAD-Tiny, motion ADE decreases from
0.91 to 0.80 m and detection NDS increases from 0.34 to 0.40.
Sparse action targets can benefit perception and prediction without
task-specific pretraining labels. Gains still vary by metric: VAD-Base's
collision rate increases, while GenAD's map mAP decreases. Closed-loop Driving
Score improves for both UniAD-Tiny and GenAD, but UniAD-Tiny's Infraction
Score decreases.

\paragraph{Explainability.}
Object returns contribute occupied evidence along candidate motions, providing
a geometric prior without object-category labels. VAD-Tiny's map mAP increases
from 0.419 to 0.439 despite the absence of map targets during pretraining
(Table~\ref{tab:training_schedule_ablation}).
Grad-CAM highlights vehicles and pedestrians (Figure~\ref{fig:gradcam_main}).
Frozen-probe analyses are provided in the Technical Supplement.

\paragraph{Limitations.}
PAVER's targets describe current geometry, not future action outcomes.
All horizons query the same sweep, so the targets do not capture obstacles
moving into or out of a candidate path. They also encode occupancy and
observability rather than traffic rules or right of way; an observed-free
corridor need not be a valid driving action. Finally, the rule-based motion
sampler leaves unsampled regions without direct target supervision during
pretraining.

\section{Conclusion}

PAVER shows that planning-relevant BEV representations can be learned through
a simple, interpretable pretraining task without dense scene reconstruction
or semantic-task supervision. Predicting risk and unknown evidence along
candidate ego motions gives the objective a clear geometric meaning:
observed obstacles and missing evidence. No driving-task annotations are
required. The compact
prediction task keeps auxiliary computation small, while the downstream
gains under shorter total training schedules suggest that this initialization
can facilitate subsequent optimization. Its benefits extend beyond planning
to perception and prediction through the shared BEV encoder. These results
highlight a route to efficient end-to-end driving: simplify the auxiliary
task while retaining relevant geometric evidence. Such supervision can
improve multi-task performance with reduced training time and without
semantic annotation costs during pretraining, while preserving the downstream
architecture and camera-only inference.

\bibliography{references}

\begin{thebibliography}{47}
\providecommand{\natexlab}[1]{#1}

\bibitem[{Amanatides and Woo(1987)}]{amanatides1987voxel}
Amanatides, J.; and Woo, A. 1987.
\newblock A Fast Voxel Traversal Algorithm for Ray Tracing.
\newblock In \emph{Eurographics}, 3--10.

\bibitem[{Caesar et~al.(2020)Caesar, Bankiti, Lang, Vora, Liong, Xu, Krishnan, Pan, Baldan, and Beijbom}]{caesar2020nuscenes}
Caesar, H.; Bankiti, V.; Lang, A.~H.; Vora, S.; Liong, V.~E.; Xu, Q.; Krishnan, A.; Pan, Y.; Baldan, G.; and Beijbom, O. 2020.
\newblock nuScenes: A Multimodal Dataset for Autonomous Driving.
\newblock In \emph{Proceedings of the IEEE/CVF Conference on Computer Vision and Pattern Recognition}, 11621--11631.

\bibitem[{Chen et~al.(2020)Chen, Zhou, Koltun, and Kr{\"a}henb{\"u}hl}]{chen2020learningbycheating}
Chen, D.; Zhou, B.; Koltun, V.; and Kr{\"a}henb{\"u}hl, P. 2020.
\newblock Learning by Cheating.
\newblock In \emph{Proceedings of the Conference on Robot Learning}, volume 100 of \emph{Proceedings of Machine Learning Research}, 66--75. PMLR.

\bibitem[{Chen et~al.(2023)Chen, Li, Zhang, Fang, Jiang, and Zhao}]{chen2022bevdistill}
Chen, Z.; Li, Z.; Zhang, S.; Fang, L.; Jiang, Q.; and Zhao, F. 2023.
\newblock BEVDistill: Cross-Modal BEV Distillation for Multi-View 3D Object Detection.
\newblock In \emph{International Conference on Learning Representations}.

\bibitem[{Deng et~al.(2009)Deng, Dong, Socher, Li, Li, and Fei-Fei}]{deng2009imagenet}
Deng, J.; Dong, W.; Socher, R.; Li, L.-J.; Li, K.; and Fei-Fei, L. 2009.
\newblock ImageNet: A Large-Scale Hierarchical Image Database.
\newblock In \emph{Proceedings of the IEEE Conference on Computer Vision and Pattern Recognition}, 248--255.

\bibitem[{Devlin et~al.(2019)Devlin, Chang, Lee, and Toutanova}]{devlin2019bert}
Devlin, J.; Chang, M.-W.; Lee, K.; and Toutanova, K. 2019.
\newblock BERT: Pre-training of Deep Bidirectional Transformers for Language Understanding.
\newblock In \emph{Proceedings of the 2019 Conference of the North American Chapter of the Association for Computational Linguistics}, 4171--4186.

\bibitem[{Doll et~al.(2024)Doll, Hanselmann, Schneider, Schulz, Cordts, Enzweiler, and Lensch}]{doll2024dualad}
Doll, S.; Hanselmann, N.; Schneider, L.; Schulz, R.; Cordts, M.; Enzweiler, M.; and Lensch, H. P.~A. 2024.
\newblock DualAD: Disentangling the Dynamic and Static World for End-to-End Driving.
\newblock In \emph{Proceedings of the IEEE/CVF Conference on Computer Vision and Pattern Recognition}, 14728--14737.

\bibitem[{Elfes(1989)}]{elfes1989occupancy}
Elfes, A. 1989.
\newblock Using Occupancy Grids for Mobile Robot Perception and Navigation.
\newblock \emph{Computer}, 22(6): 46--57.

\bibitem[{Fawcett(2006)}]{fawcett2006roc}
Fawcett, T. 2006.
\newblock An Introduction to ROC Analysis.
\newblock \emph{Pattern Recognition Letters}, 27(8): 861--874.

\bibitem[{He et~al.(2016)He, Zhang, Ren, and Sun}]{he2016resnet}
He, K.; Zhang, X.; Ren, S.; and Sun, J. 2016.
\newblock Deep Residual Learning for Image Recognition.
\newblock In \emph{Proceedings of the IEEE Conference on Computer Vision and Pattern Recognition}, 770--778.

\bibitem[{Hinton, Vinyals, and Dean(2015)}]{hinton2015distilling}
Hinton, G.; Vinyals, O.; and Dean, J. 2015.
\newblock Distilling the Knowledge in a Neural Network.
\newblock \emph{arXiv preprint arXiv:1503.02531}.

\bibitem[{Hu et~al.(2022)Hu, Chen, Wu, Li, Yan, and Tao}]{hu2022stp3}
Hu, S.; Chen, L.; Wu, P.; Li, H.; Yan, J.; and Tao, D. 2022.
\newblock ST-P3: End-to-End Vision-Based Autonomous Driving via Spatial-Temporal Feature Learning.
\newblock In \emph{European Conference on Computer Vision}, 533--549.

\bibitem[{Hu et~al.(2023)Hu, Yang, Chen, Li, Sima, Zhu, Chai, Du, Lin, Wang, Lu, Jia, Liu, Dai, Qiao, and Li}]{hu2023planning}
Hu, Y.; Yang, J.; Chen, L.; Li, K.; Sima, C.; Zhu, X.; Chai, S.; Du, S.; Lin, T.; Wang, W.; Lu, L.; Jia, X.; Liu, Q.; Dai, J.; Qiao, Y.; and Li, H. 2023.
\newblock Planning-Oriented Autonomous Driving.
\newblock In \emph{Proceedings of the IEEE/CVF Conference on Computer Vision and Pattern Recognition}, 17853--17862.

\bibitem[{Huang et~al.(2023)Huang, Zheng, Zhang, Zhou, and Lu}]{huang2023tpvformer}
Huang, Y.; Zheng, W.; Zhang, Y.; Zhou, J.; and Lu, J. 2023.
\newblock Tri-Perspective View for Vision-Based 3D Semantic Occupancy Prediction.
\newblock In \emph{Proceedings of the IEEE/CVF Conference on Computer Vision and Pattern Recognition}, 9223--9232.

\bibitem[{Jaderberg et~al.(2015)Jaderberg, Simonyan, Zisserman, and Kavukcuoglu}]{jaderberg2015stn}
Jaderberg, M.; Simonyan, K.; Zisserman, A.; and Kavukcuoglu, K. 2015.
\newblock Spatial Transformer Networks.
\newblock In \emph{Advances in Neural Information Processing Systems}, volume~28.

\bibitem[{Jia et~al.(2024)Jia, Yang, Li, Zhang, and Yan}]{jia2024bench2drive}
Jia, X.; Yang, Z.; Li, Q.; Zhang, Z.; and Yan, J. 2024.
\newblock Bench2Drive: Towards Multi-Ability Benchmarking of Closed-Loop End-to-End Autonomous Driving.
\newblock In \emph{Advances in Neural Information Processing Systems, Datasets and Benchmarks Track}.

\bibitem[{Jiang et~al.(2023)Jiang, Chen, Xu, Liao, Chen, Zhou, Zhang, Liu, Huang, and Wang}]{jiang2023vad}
Jiang, B.; Chen, S.; Xu, Q.; Liao, B.; Chen, J.; Zhou, H.; Zhang, Q.; Liu, W.; Huang, C.; and Wang, X. 2023.
\newblock VAD: Vectorized Scene Representation for Efficient Autonomous Driving.
\newblock In \emph{Proceedings of the IEEE/CVF International Conference on Computer Vision}, 8340--8350.

\bibitem[{Kim et~al.(2024)Kim, Kim, Hwang, Jeong, and Kum}]{chen2024labeldistill}
Kim, S.; Kim, Y.; Hwang, S.; Jeong, H.; and Kum, D. 2024.
\newblock LabelDistill: Label-Guided Cross-Modal Knowledge Distillation for Camera-Based 3D Object Detection.
\newblock \emph{arXiv preprint arXiv:2407.10164}.

\bibitem[{Li and Cui(2025)}]{li2025ssr}
Li, P.; and Cui, D. 2025.
\newblock Navigation-Guided Sparse Scene Representation for End-to-End Autonomous Driving.
\newblock In \emph{International Conference on Learning Representations}.

\bibitem[{Li et~al.(2025)Li, Fan, He, Wang, Chen, Zhang, and Tan}]{li2025law}
Li, Y.; Fan, L.; He, J.; Wang, Y.; Chen, Y.; Zhang, Z.; and Tan, T. 2025.
\newblock Enhancing End-to-End Autonomous Driving with Latent World Model.
\newblock In \emph{International Conference on Learning Representations}.

\bibitem[{Li et~al.(2023)Li, Ge, Yu, Yang, Wang, Shi, Sun, and Li}]{li2022bevdepth}
Li, Y.; Ge, Z.; Yu, G.; Yang, J.; Wang, Z.; Shi, Y.; Sun, J.; and Li, Z. 2023.
\newblock BEVDepth: Acquisition of Reliable Depth for Multi-View 3D Object Detection.
\newblock In \emph{Proceedings of the AAAI Conference on Artificial Intelligence}, volume~37, 1477--1485.

\bibitem[{Li et~al.(2022)Li, Wang, Li, Xie, Sima, Lu, Qiao, and Dai}]{li2022bevformer}
Li, Z.; Wang, W.; Li, H.; Xie, E.; Sima, C.; Lu, T.; Qiao, Y.; and Dai, J. 2022.
\newblock BEVFormer: Learning Bird's-Eye-View Representation from Multi-Camera Images via Spatiotemporal Transformers.
\newblock In \emph{European Conference on Computer Vision}, 1--18.

\bibitem[{Liao et~al.(2025)Liao, Chen, Yin, Jiang, Wang, Yan, Zhang, Li, Zhang, Zhang, and Wang}]{liao2025diffusiondrive}
Liao, B.; Chen, S.; Yin, H.; Jiang, B.; Wang, C.; Yan, S.; Zhang, X.; Li, X.; Zhang, Y.; Zhang, Q.; and Wang, X. 2025.
\newblock DiffusionDrive: Truncated Diffusion Model for End-to-End Autonomous Driving.
\newblock In \emph{Proceedings of the IEEE/CVF Conference on Computer Vision and Pattern Recognition}, 12037--12047.

\bibitem[{Lin et~al.(2017)Lin, Doll{\'a}r, Girshick, He, Hariharan, and Belongie}]{lin2017fpn}
Lin, T.-Y.; Doll{\'a}r, P.; Girshick, R.; He, K.; Hariharan, B.; and Belongie, S. 2017.
\newblock Feature Pyramid Networks for Object Detection.
\newblock In \emph{Proceedings of the IEEE Conference on Computer Vision and Pattern Recognition}, 2117--2125.

\bibitem[{Lin et~al.(2024)Lin, Wang, Qi, Dong, and Yang}]{lin2024bevmae}
Lin, Z.; Wang, Y.; Qi, S.; Dong, N.; and Yang, M.-H. 2024.
\newblock BEV-MAE: Bird's Eye View Masked Autoencoders for Point Cloud Pre-training in Autonomous Driving Scenarios.
\newblock In \emph{Proceedings of the AAAI Conference on Artificial Intelligence}, volume~38, 3531--3539.

\bibitem[{Loshchilov and Hutter(2017)}]{loshchilov2017sgdr}
Loshchilov, I.; and Hutter, F. 2017.
\newblock SGDR: Stochastic Gradient Descent with Warm Restarts.
\newblock In \emph{International Conference on Learning Representations}.

\bibitem[{Loshchilov and Hutter(2019)}]{loshchilov2019adamw}
Loshchilov, I.; and Hutter, F. 2019.
\newblock Decoupled Weight Decay Regularization.
\newblock In \emph{International Conference on Learning Representations}.

\bibitem[{Min et~al.(2022)Min, Xu, Zhao, Xiao, Nie, and Dai}]{min2022occupancymae}
Min, C.; Xu, X.; Zhao, D.; Xiao, L.; Nie, Y.; and Dai, B. 2022.
\newblock Self-Supervised Pre-Training Large-Scale LiDAR Point Clouds with Masked Occupancy Autoencoders.
\newblock \emph{arXiv preprint arXiv:2206.09900}.

\bibitem[{Min et~al.(2024)Min, Zhao, Xiao, Zhao, Xu, Zhu, Jin, Li, Guo, Xing, Jing, Nie, and Dai}]{min2024driveworld}
Min, C.; Zhao, D.; Xiao, L.; Zhao, J.; Xu, X.; Zhu, Z.; Jin, L.; Li, J.; Guo, Y.; Xing, J.; Jing, L.; Nie, Y.; and Dai, B. 2024.
\newblock DriveWorld: 4D Pre-trained Scene Understanding via World Models for Autonomous Driving.
\newblock In \emph{Proceedings of the IEEE/CVF Conference on Computer Vision and Pattern Recognition}, 15522--15533.

\bibitem[{Piccinelli et~al.(2025)Piccinelli, Sakaridis, Segu, Yang, Li, Abbeloos, and Van~Gool}]{piccinelli2025unik3d}
Piccinelli, L.; Sakaridis, C.; Segu, M.; Yang, Y.-H.; Li, S.; Abbeloos, W.; and Van~Gool, L. 2025.
\newblock UniK3D: Universal Camera Monocular 3D Estimation.
\newblock \emph{arXiv preprint arXiv:2503.16591}.

\bibitem[{Shi et~al.(2025)Shi, Shi, Sheng, Zhang, and Jiang}]{shi2025drivex}
Shi, C.; Shi, S.; Sheng, K.; Zhang, B.; and Jiang, L. 2025.
\newblock DriveX: Omni Scene Modeling for Learning Generalizable World Knowledge in Autonomous Driving.
\newblock In \emph{Proceedings of the IEEE/CVF International Conference on Computer Vision}, 28599--28609.

\bibitem[{Sun et~al.(2024)Sun, Lin, Shi, Zhang, Wu, and Zheng}]{sun2024sparsedrive}
Sun, W.; Lin, X.; Shi, Y.; Zhang, C.; Wu, H.; and Zheng, S. 2024.
\newblock SparseDrive: End-to-End Autonomous Driving via Sparse Scene Representation.
\newblock \emph{arXiv preprint arXiv:2405.19620}.

\bibitem[{Tang et~al.(2025)Tang, Xu, Meng, and Cheng}]{tang2025hipad}
Tang, Y.; Xu, Z.; Meng, Z.; and Cheng, E. 2025.
\newblock HiP-AD: Hierarchical and Multi-Granularity Planning with Deformable Attention for Autonomous Driving in a Single Decoder.
\newblock In \emph{Proceedings of the IEEE/CVF International Conference on Computer Vision}.

\bibitem[{Vapnik and Izmailov(2015)}]{vapnik2015lupi}
Vapnik, V.; and Izmailov, R. 2015.
\newblock Learning Using Privileged Information: Similarity Control and Knowledge Transfer.
\newblock \emph{Journal of Machine Learning Research}, 16(1): 2023--2049.

\bibitem[{Wang et~al.(2026)Wang, Yang, Bai, Zhang, Liu, Zheng, Long, Lu, and Lu}]{wang2026drivejepa}
Wang, L.; Yang, Z.; Bai, C.; Zhang, G.; Liu, X.; Zheng, X.; Long, X.-X.; Lu, C.-T.; and Lu, C. 2026.
\newblock Drive-JEPA: Video JEPA Meets Multimodal Trajectory Distillation for End-to-End Driving.
\newblock \emph{arXiv preprint arXiv:2601.22032}.

\bibitem[{Wang et~al.(2019)Wang, Chao, Garg, Hariharan, Campbell, and Weinberger}]{wang2019pseudolidar}
Wang, Y.; Chao, W.-L.; Garg, D.; Hariharan, B.; Campbell, M.; and Weinberger, K.~Q. 2019.
\newblock Pseudo-LiDAR from Visual Depth Estimation: Bridging the Gap in 3D Object Detection for Autonomous Driving.
\newblock In \emph{Proceedings of the IEEE/CVF Conference on Computer Vision and Pattern Recognition}, 8445--8453.

\bibitem[{Wang et~al.(2023)Wang, Li, Luo, Xie, and Yang}]{wang2023distillbev}
Wang, Z.; Li, D.; Luo, C.; Xie, C.; and Yang, X. 2023.
\newblock DistillBEV: Boosting Multi-Camera 3D Object Detection with Cross-Modal Knowledge Distillation.
\newblock In \emph{Proceedings of the IEEE/CVF International Conference on Computer Vision}, 8637--8646.

\bibitem[{Weng et~al.(2024)Weng, Ivanovic, Wang, Wang, and Pavone}]{weng2024paradrive}
Weng, X.; Ivanovic, B.; Wang, Y.; Wang, Y.; and Pavone, M. 2024.
\newblock PARA-Drive: Parallelized Architecture for Real-time Autonomous Driving.
\newblock In \emph{Proceedings of the IEEE/CVF Conference on Computer Vision and Pattern Recognition}, 15449--15458.

\bibitem[{Xiao et~al.(2021)Xiao, Codevilla, Pal, and Lopez}]{xiao2021actionbased}
Xiao, Y.; Codevilla, F.; Pal, C.; and Lopez, A.~M. 2021.
\newblock Action-Based Representation Learning for Autonomous Driving.
\newblock In \emph{Proceedings of the Conference on Robot Learning}, volume 155 of \emph{Proceedings of Machine Learning Research}, 232--246.

\bibitem[{Yang et~al.(2024{\natexlab{a}})Yang, Zhang, Huang, Wu, Zhu, He, Tang, Zhao, Qiu, Lin, He, and Ouyang}]{yang2024unipad}
Yang, H.; Zhang, S.; Huang, D.; Wu, X.; Zhu, H.; He, T.; Tang, S.; Zhao, H.; Qiu, Q.; Lin, B.; He, X.; and Ouyang, W. 2024{\natexlab{a}}.
\newblock UniPAD: A Universal Pre-training Paradigm for Autonomous Driving.
\newblock In \emph{Proceedings of the IEEE/CVF Conference on Computer Vision and Pattern Recognition}, 15238--15250.

\bibitem[{Yang et~al.(2024{\natexlab{b}})Yang, Chen, Sun, and Li}]{yang2024vidar}
Yang, Z.; Chen, L.; Sun, Y.; and Li, H. 2024{\natexlab{b}}.
\newblock Visual Point Cloud Forecasting Enables Scalable Autonomous Driving.
\newblock In \emph{Proceedings of the IEEE/CVF Conference on Computer Vision and Pattern Recognition}, 14673--14684.

\bibitem[{Zhang et~al.(2025)Zhang, Zhou, Zhu, Yan, Gao, Bai, Cai, Liu, Cui, and Li}]{zhang2025visionpad}
Zhang, H.; Zhou, W.; Zhu, Y.; Yan, X.; Gao, J.; Bai, D.; Cai, Y.; Liu, B.; Cui, S.; and Li, Z. 2025.
\newblock VisionPAD: A Vision-Centric Pre-training Paradigm for Autonomous Driving.
\newblock In \emph{Proceedings of the IEEE/CVF Conference on Computer Vision and Pattern Recognition}, 17165--17175.

\bibitem[{Zhang, Peng, and Zhou(2022)}]{zhang2022actionconditioned}
Zhang, Q.; Peng, Z.; and Zhou, B. 2022.
\newblock Learning to Drive by Watching YouTube Videos: Action-Conditioned Contrastive Policy Pretraining.
\newblock In \emph{European Conference on Computer Vision}, 111--128. Springer.

\bibitem[{Zheng et~al.(2024)Zheng, Song, Guo, Zhang, and Chen}]{zheng2024genad}
Zheng, W.; Song, R.; Guo, X.; Zhang, C.; and Chen, L. 2024.
\newblock GenAD: Generative End-to-End Autonomous Driving.
\newblock In \emph{European Conference on Computer Vision}, 87--104. Springer.

\bibitem[{Zhou et~al.(2023)Zhou, Liu, Hu, Zhou, and Ma}]{zhou2023unidistill}
Zhou, S.; Liu, W.; Hu, C.; Zhou, S.; and Ma, C. 2023.
\newblock UniDistill: A Universal Cross-Modality Knowledge Distillation Framework for 3D Object Detection in Bird's-Eye View.
\newblock In \emph{Proceedings of the IEEE/CVF Conference on Computer Vision and Pattern Recognition}, 5116--5125.

\bibitem[{Zhu et~al.(2026)Zhu, Xue, Zhang, Jiang, Zhou, Yan, Gao, Cai, Liu, Li, and Shen}]{zhu2026dlwm}
Zhu, Y.; Xue, Y.; Zhang, H.; Jiang, G.; Zhou, W.; Yan, X.; Gao, J.; Cai, Y.; Liu, B.; Li, Z.; and Shen, S. 2026.
\newblock DLWM: Dual Latent World Models Enable Holistic Gaussian-Centric Pre-training in Autonomous Driving.
\newblock In \emph{Proceedings of the IEEE/CVF Conference on Computer Vision and Pattern Recognition}, 39713--39723.

\bibitem[{Zou et~al.(2025)Zou, Liao, Zhang, Liu, and Wang}]{zou2025mim4d}
Zou, J.; Liao, B.; Zhang, Q.; Liu, W.; and Wang, X. 2025.
\newblock MIM4D: Masked Modeling with Multi-View Video for Autonomous Driving Representation Learning.
\newblock \emph{International Journal of Computer Vision}, 133: 6074--6087.

\end{thebibliography}

\clearpage
\onecolumn

\setcounter{topnumber}{3}
\setcounter{bottomnumber}{2}
\setcounter{totalnumber}{5}
\setcounter{dbltopnumber}{3}
\renewcommand{\topfraction}{0.92}
\renewcommand{\bottomfraction}{0.5}
\renewcommand{\dbltopfraction}{0.92}
\renewcommand{\textfraction}{0.06}
\renewcommand{\floatpagefraction}{0.5}
\renewcommand{\dblfloatpagefraction}{0.5}

\twocolumn[{
\begin{center}
{\LARGE\bfseries Planning-Aligned Pretraining of BEV Representations\\
with Sparse Action-Conditioned Targets for End-to-End Autonomous Driving\\[2pt]
{\large Technical Supplement}\par}
\end{center}
\vspace{6pt}
}]

\raggedbottom
\makeatletter
\setlength{\@fptop}{0pt plus 1fil}
\setlength{\@fpbot}{0pt plus 1fil}
\setlength{\@dblfptop}{0pt plus 1fil}
\setlength{\@dblfpbot}{0pt plus 1fil}
\setlength{\@fpsep}{12pt}
\setlength{\@dblfpsep}{12pt}
\makeatother

\section{Method and Reproducibility Details}
\label{sec:supp_reproducibility}

This section records the implementation details needed to reproduce PAVER.
Every value below is supported by the canonical configuration, implementation,
or preserved experiment artifacts.
Metric abbreviations follow their task-native evaluators. For perception, mAP is
mean average precision and NDS the nuScenes detection score; for motion, EPA is
the end-to-end prediction accuracy, ADE and FDE the average and final
displacement errors, and MR the miss rate; for occupancy, IoU and mIoU are the
completion and semantic intersection over union. The analyses additionally use
the area under the receiver operating characteristic (AUROC). The closed-loop
results use the Bench2Drive driving score (DS), route completion (RC) and
infraction score (IS).

Model names follow one convention throughout this supplement. \emph{VAD-Tiny
(scratch)} is the reproduced VAD-Tiny trained end to end for 60 downstream
epochs with no pretraining; it is the reference every VAD-Tiny comparison is
made against, and tables abbreviate it to Scratch. \emph{VAD-Tiny (staged)} is
the same architecture given 48 epochs of pretraining on the downstream tasks
followed by 12 fine-tuning epochs, so its 60-epoch total matches scratch and
isolates the effect of the pretraining objective rather than of extra
optimization; tables abbreviate it to Staged. \emph{+ PAVER} denotes 20
pretraining epochs with the PAVER objective followed by downstream fine-tuning,
with only the BEV encoder transferred. Where the auxiliary capacity varies, the
parenthesized figure is the number of trainable parameters in the temporary
PAVER head. PAVER (10K) is the canonical pointwise readout; PAVER (30K) widens
its projector and action-conditioned MLP, while PAVER (90K) replaces the wider
pointwise projector with a $3\times3$ projector that accesses local BEV context.
The exact counts are given in Table~\ref{tab:supp_auxiliary_parameters}.
\emph{VAD-Base (scratch)},
\emph{GenAD (scratch)} and \emph{UniAD-Tiny (scratch)} are used in the same
sense for the other architectures.

The experiments use the VAD BEV model \citep{jiang2023vad} on
nuScenes \citep{caesar2020nuscenes}; its spatially indexed BEV follows the
multi-camera representation used by BEVFormer \citep{li2022bevformer}.

\subsection{Model and Data Contract}

\begin{table*}[t]
\centering
\small
\renewcommand{\arraystretch}{1.06}
\begin{tabular*}{\textwidth}{@{\extracolsep{\fill}}ll@{}}
\toprule
Item & Configuration \\
\midrule
Training split & 28,130 nuScenes keyframes \\
Validation split & 6,019 nuScenes keyframes \\
Camera input & 6 synchronized surround views \\
Image backbone and neck & ResNet-50 and FPN \\
BEV representation & $[B,256,100,100]$ \\
BEV range & $x\in[-15,15)$ m, $y\in[-30,30)$ m \\
Pretraining queue & 1 frame \\
Privileged input & Current LiDAR sweep, used only for targets \\
Active pretraining outputs & Risk and Unknown logits \\
Task annotations & None \\
Downstream inference & Camera-only \\
\bottomrule
\end{tabular*}
\caption{PAVER data and model configuration.}
\label{tab:supp_model_contract}
\end{table*}

Table~\ref{tab:supp_model_contract} records the data and model configuration.
All camera pipelines use BGR normalization with mean
$(103.530,116.280,123.675)$, unit standard deviation, and no RGB conversion.
The single-frame pretraining queue prevents the objective from using a recurrent
BEV state. Detection, mapping, motion-prediction, and planning decoders are
disabled during pretraining.

\subsection{Rule-Based Action Sampling}

Let $a_s$ and $\omega_s$ denote the perturbed acceleration and yaw rate for
action slot $s$. Starting from current speed $v_0$, the rollout is
\begin{align}
v_{s,t} &= \max(0,v_0+a_s t\Delta t), \\
\psi_{s,t} &= \omega_s t\Delta t, \\
x_{s,t} &= \sum_{j=1}^{t}v_{s,j}\cos(\psi_{s,j})\Delta t, \\
y_{s,t} &= \sum_{j=1}^{t}v_{s,j}\sin(\psi_{s,j})\Delta t,
\end{align}
where $\Delta t=0.5$ s. Target construction and BEV indexing use the calibrated
sensor frame, while the prediction head receives the corresponding ego-frame
state $[x,y,\psi,v]$.

\subsection{Sparse Evidence and Action Targets}

Following occupancy-grid mapping \citep{elfes1989occupancy}, the evidence map
is initialized as Unknown. For each valid LiDAR return, samples
along the open ray segment are written as Free and the measured endpoint is then
written as Occupied. At each action state, five samples across the vehicle width
query this raster. The resulting soft ratios are supervision targets, not a
dense semantic-occupancy prediction. An observed-free cross section has target
$(r,u)=(0,0)$ and remains part of the loss.

Formally, for heading $\bar\psi_{s,t}$ the lateral unit vector is
$\ell(\bar\psi_{s,t})=(-\sin\bar\psi_{s,t},\cos\bar\psi_{s,t})$, so the $k$-th
query position is
\begin{equation}
p_{s,t,k}=(\bar x_{s,t},\bar y_{s,t})+d_k\,\ell(\bar\psi_{s,t}),
\label{eq:supp_lateral_query}
\end{equation}
with $d_k$ denoting the $k$-th symmetric lateral offset. Writing
$\mathcal{V}_{s,t}$ for the queries that survive the range check, the risk and
unknown targets used in the main paper are the indicator averages
\begin{align}
r_{s,t}&=\frac{1}{|\mathcal{V}_{s,t}|}\sum_{k\in\mathcal{V}_{s,t}}\mathbb{1}\left[E(p_{s,t,k})=\textsc{Occupied}\right],\\
u_{s,t}&=\frac{1}{|\mathcal{V}_{s,t}|}\sum_{k\in\mathcal{V}_{s,t}}\mathbb{1}\left[E(p_{s,t,k})=\textsc{Unknown}\right].
\end{align}
Out-of-range queries are dropped rather than clamped, so an action that leaves
the mapped region contributes fewer samples instead of a biased target.

This construction compresses the privileged sweep into two quantities for each
action--horizon query: whether measured endpoints intersect the vehicle width
and whether ray evidence supports that region. Unlike objectives that reconstruct
appearance, depth, or dense scene content \citep{yang2024unipad,zou2025mim4d,
zhang2025visionpad}, it does not ask the auxiliary branch to reproduce details
that the downstream planner never queries directly. The prediction problem is
small, but its coordinates are not: answering it requires the camera BEV to
place free, occupied, and unsupported evidence at the locations indexed by an
ego motion.

\subsection{PAVER Head and Objective}

The mask $M\in\{0,1\}^{B\times1\times H\times W}$ is the union of all rasterized
lateral-query cells, and a cell reached by several queries is included once.
With the shared token $m\in\mathbb{R}^{1\times C\times1\times1}$, the temporary
copy is $F^{\mathrm{mask}}=(1-M)\odot F+M\odot m$, the pointwise projector gives
$Z=\phi(F^{\mathrm{mask}})\in\mathbb{R}^{B\times C_p\times H\times W}$, and each
action--horizon entry reads a single feature
$z_{s,t}=\mathcal{S}(Z,(\bar x_{s,t},\bar y_{s,t}))\in\mathbb{R}^{C_p}$ through
bilinear sampling $\mathcal{S}$ \citep{jaderberg2015stn}. The predictor input is
the concatenation $h_{s,t}=[z_{s,t};q_{s,t}^{\mathrm{ego}}]$, so the PAVER head sees
a projected BEV feature whose corridor cells have been replaced, together with
the action state that selected it.

The action corridor is replaced only in the temporary BEV copy consumed by the
PAVER head. The original BEV representation remains unchanged. The
objective is
\begin{equation}
\mathcal{L}_{\mathrm{PAVER}}
=\mathrm{BCELogit}(\hat r,r)
+0.5\,\mathrm{BCELogit}(\hat u,u).
\end{equation}

The 10,258-parameter PAVER head is therefore only a prediction head, not the representation
being transferred. Masking removes the feature at the supervised corridor cell,
and each loss is differentiated through bilinear BEV sampling and the shared
BEV encoder. Increasing predictor capacity is unnecessary for
expressing the two targets; useful scene information must instead be organized in
the BEV encoder that remains after pretraining. This differs from masked
reconstruction, where decoder capacity participates in recovering a
high-dimensional observation \citep{devlin2019bert,lin2024bevmae,
min2022occupancymae}.

\begin{table}[t]
\centering
\small
\setlength{\tabcolsep}{5.0pt}
\begin{tabular}{lrrr}
\toprule
Learned component & 10K & 30K & 90K \\
\midrule
Mask token & 256 & 256 & 256 \\
Projector & 4,112 & 8,224 & 73,760 \\
First readout layer & 5,376 & 18,944 & 18,944 \\
Output layer & 514 & 1,026 & 1,026 \\
\midrule
Total & 10,258 & 28,450 & 93,986 \\
\bottomrule
\end{tabular}
\caption{Auxiliary parameter counts for the PAVER capacity variants. The 10K
and 30K projectors are pointwise; the 90K projector uses a $3\times3$ kernel.}
\label{tab:supp_auxiliary_parameters}
\end{table}

\subsection{Optimization and Transfer}

\begin{table}[t]
\centering
\small
\renewcommand{\arraystretch}{1.05}
\setlength{\tabcolsep}{3.0pt}
\begin{tabular}{p{0.42\columnwidth}p{0.51\columnwidth}}
\toprule
Setting & Value \\
\midrule
Pretraining duration & 20 epochs \\
Pretraining batch & 4 GPUs, 4 samples per GPU \\
Optimizer & AdamW \citep{loshchilov2019adamw} \\
Learning rate & $5\times10^{-5}$ \\
Image-backbone multiplier & 0.2 \\
Weight decay & 0.01 \\
Schedule & Cosine annealing \citep{loshchilov2017sgdr} \\
Warmup & 500 iterations, linear, ratio $1/3$ \\
Minimum learning-rate ratio & $10^{-3}$ \\
Gradient clipping & Norm 35 \\
Checkpoint interval & 1 epoch \\
Downstream learning rate & $2\times10^{-4}$ \\
Downstream queue & 3 frames \\
\bottomrule
\end{tabular}
\caption{Optimization settings for the reported PAVER experiments.}
\label{tab:supp_optimization}
\end{table}

Under the settings of Table~\ref{tab:supp_optimization}, all transferred
image-backbone, FPN, and BEV-encoder parameters are optimized during
pretraining. Afterward,
the action sampler, target builder, mask token, projector, and prediction MLP are
discarded. The transferred initialization covers the image backbone, FPN, BEV
queries and positional encodings, camera and feature-level embeddings, CAN-bus
MLP, and BEV transformer encoder. Task decoders are newly instantiated, and downstream training starts from epoch 1
with a fresh optimizer and scheduler. No PAVER-specific parameter is active at
inference.

\section{Frozen-Predictor Controls}
\label{sec:supp_frozen_controls}

\begin{figure}[t]
\centering
\includegraphics[width=\columnwidth]{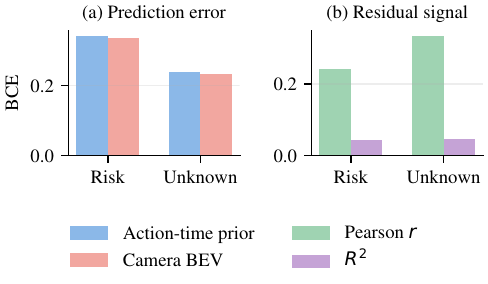}
\caption{Action--time shortcut audit. The frozen BEV predictor is compared with
a prior conditioned on control family, horizon, and perturbed action state.}
\label{fig:supp_action_time_prior}
\end{figure}

We evaluate the frozen auxiliary predictor on 192 nuScenes validation samples:
Table~\ref{tab:supp_scene_interventions} reports the interventions,
Table~\ref{tab:supp_intervention_uncertainty} their scene-bootstrap uncertainty,
Table~\ref{tab:supp_offsupport_probe} the off-support probe, and
Figure~\ref{fig:supp_action_time_prior} the action--time shortcut audit.
Each intervention preserves the weights and targets while changing only the
specified BEV or action-state input. These controls diagnose input dependence;
they are not downstream model-selection results.

\begin{table}[t]
\centering
\small
\renewcommand{\arraystretch}{1.03}
\setlength{\tabcolsep}{2.3pt}
\begin{tabular}{@{}lrrrr@{}}
\toprule
Input & Risk BCE & $\Delta$ & Unknown BCE & $\Delta$ \\
\midrule
Correct input & \textbf{0.3366} & 0.0000 & \textbf{0.2333} & 0.0000 \\
Cross-scene BEV & 0.3571 & 0.0204 & 0.2723 & 0.0390 \\
Zero context & 0.3511 & 0.0145 & 0.2784 & 0.0451 \\
Spatial permutation & 0.3691 & 0.0325 & 0.3254 & 0.0920 \\
Action permutation & 0.3879 & 0.0513 & 0.2688 & 0.0355 \\
\bottomrule
\end{tabular}
\caption{Frozen-input intervention results. Lower is better; $\Delta$ is measured from the correct-input result and computed before rounding the displayed losses.}
\label{tab:supp_scene_interventions}
\end{table}

\begin{table*}[t]
\centering
\small
\renewcommand{\arraystretch}{1.05}
\begin{tabular*}{\textwidth}{@{\extracolsep{\fill}}lcccc@{}}
\toprule
 & Cross-scene & Zero ctx. & Spatial perm. & Action perm. \\
\midrule
$\Delta$ Risk [95\% CI] & $0.0204\,[0.0150,\,0.0241]$ & $0.0145\,[0.0099,\,0.0176]$ & $0.0325\,[0.0292,\,0.0358]$ & $0.0513\,[0.0500,\,0.0535]$ \\
$\Delta$ Unknown [95\% CI] & $0.0390\,[0.0288,\,0.0464]$ & $0.0451\,[0.0390,\,0.0512]$ & $0.0920\,[0.0835,\,0.1002]$ & $0.0355\,[0.0329,\,0.0378]$ \\
\bottomrule
\end{tabular*}
\caption{Scene-bootstrap uncertainty of the frozen-input interventions.}
\label{tab:supp_intervention_uncertainty}
\end{table*}

Every intervention increases both losses, and every paired interval excludes
zero. Spatial permutation has the largest effect on Unknown, while action
permutation has the largest effect on Risk. The predictor therefore uses the
scene, its spatial arrangement, and the action--scene pairing rather than only a
fixed action and horizon prior.

The BEV predictor reduces Risk BCE by 0.0042 and Unknown BCE by 0.0061 relative
to the action--time prior. Residual $R^2$ is below 0.05 for both targets, while
Pearson correlations of 0.2410 and 0.3321 retain measurable scene dependence.
Thus, action and horizon explain substantial target variation, but do not fully
explain the learned prediction.

The sparse action targets occupy 2.66\% of the BEV and remain in the forward
half of the ego frame. Identical frozen linear probes nevertheless show AUROC
changes of $+0.0868$ in rear regions, $+0.0617$ in lateral regions, and
$+0.1488$ in front off-path regions; all paired 95\% confidence intervals
exclude zero. These measurements establish off-support representation change,
not direct supervision at those cells.

The architecture exposes a specific path for this change. All six views pass
through the same ResNet and FPN. Camera identity and calibration then determine
where each view is sampled, but the deformable-attention transformations and
output projection are shared before visible-view contributions are accumulated
in a common BEV \citep{li2022bevformer,jiang2023vad}. A sparse loss therefore
updates parameters that are reused across cameras and BEV locations even though
only a small set of output cells receives an immediate gradient. This
distinction separates spatially sparse targets from a camera-specific or
head-only optimization path.

This mechanism explains how an action-aligned objective can influence the BEV
later consumed by detection and mapping without pretraining their decoders.
Both perception endpoints improve on VAD-Tiny: map mAP rises from 0.4188 to
0.4386 and detection mAP from 0.2306 to 0.2784. Transfer is not uniform across
architectures, so the evidence supports broad representational reach rather
than direct learning of task semantics.

\begin{table}[t]
\centering
\small
\renewcommand{\arraystretch}{1.05}
\setlength{\tabcolsep}{3.0pt}
\begin{tabular}{lcccc}
\toprule
Region & Init. & PAVER & $\Delta$ & 95\% CI \\
\midrule
Rear & 0.6532 & 0.7399 & +0.0868 & [0.0598, 0.1131] \\
Lateral & 0.6455 & 0.7072 & +0.0617 & [0.0250, 0.0956] \\
Front off-path & 0.6716 & 0.8205 & +0.1488 & [0.1032, 0.1916] \\
\bottomrule
\end{tabular}
\caption{Frozen object--background probe AUROC outside direct action support. Differences are computed before rounding the displayed AUROC values.}
\label{tab:supp_offsupport_probe}
\end{table}

\providecommand{\analysisvariant}{}
\section{Comprehensive Analysis}
\label{sec:comprehensive_analysis}

The endpoint comparisons establish transfer performance but do not explain what
the pretraining objective learns or how its effect develops during downstream
optimization. Unless stated otherwise, downstream analyses use the selected
VAD-Tiny PAVER model reported in the main experiments.

The analyses test a specific account of PAVER's efficiency: dense objectives
preserve broad appearance and 3D scene content
\citep{yang2024unipad,zou2025mim4d,zhang2025visionpad}, whereas planning consumes
spatial evidence through candidate ego motions \citep{hu2023planning,jiang2023vad},
and PAVER retains only their intersection. The auxiliary head can then remain
small, because the shared BEV encoder is what must place the relevant evidence at
the queried locations.


\subsection{Changes in the Transferred BEV Representation}

\begin{figure}[tb]
\centering
\includegraphics[width=\columnwidth]{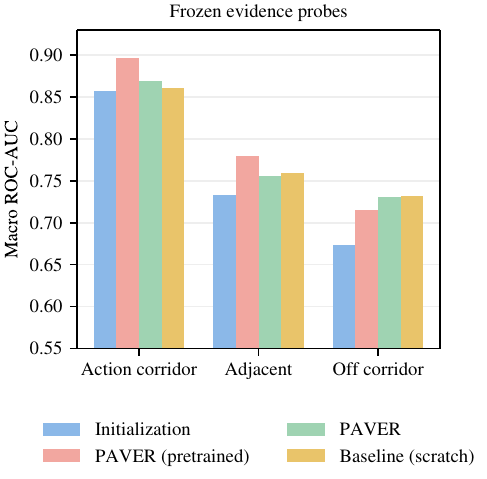}
\caption{Frozen evidence probes by BEV region. PAVER pretraining increases
linear accessibility across regions; downstream differences are smaller and
mixed.}
\label{fig:analysis_evidence_probes}
\end{figure}


Matched frozen linear probes predict Unknown, Free, and Occupied evidence from
camera-derived BEV features (Figure~\ref{fig:analysis_evidence_probes}). We use ROC-AUC to measure linear accessibility
independently of a single decision threshold \citep{fawcett2006roc}. PAVER
pretraining increases macro ROC-AUC from 0.8484 to 0.8931 in the action
corridor, from 0.7326 to 0.7755 in the adjacent ring, and from 0.6708 to
0.7175 off corridor. After downstream training, the all-region probe remains
slightly above the reproduced scratch model (0.7692 versus 0.7618), but the
regional result is mixed: corridor AUC is 0.8600 versus 0.8635, whereas
adjacent and off-corridor AUC are 0.7573 versus 0.7532 and 0.7348 versus
0.7254. The target is spatially sparse, but its representation changes are not
confined to the supervised cells.

This pattern clarifies the role of the pretraining-only auxiliary head. Its 10,258
parameters are discarded, so they cannot supply downstream capacity. What
persists is easier linear access to evidence where candidate actions query the
BEV. The objective therefore does not need to encode a complete scene in its
auxiliary output: it only needs enough capacity to expose whether the shared BEV has
organized local geometric support around an action. Comparable probe gains
outside the corridor are consistent with updating the shared image and BEV
modules used to construct the full representation, rather than adding capacity
only at queried cells.

These frozen probes measure linear accessibility, not causal use by downstream
decoders.

\subsection{Downstream Adaptation and Checkpoint Dynamics}


\begin{figure*}[!tp]
\centering
\includegraphics[width=\textwidth,trim=0bp 100bp 0bp 0bp,clip]{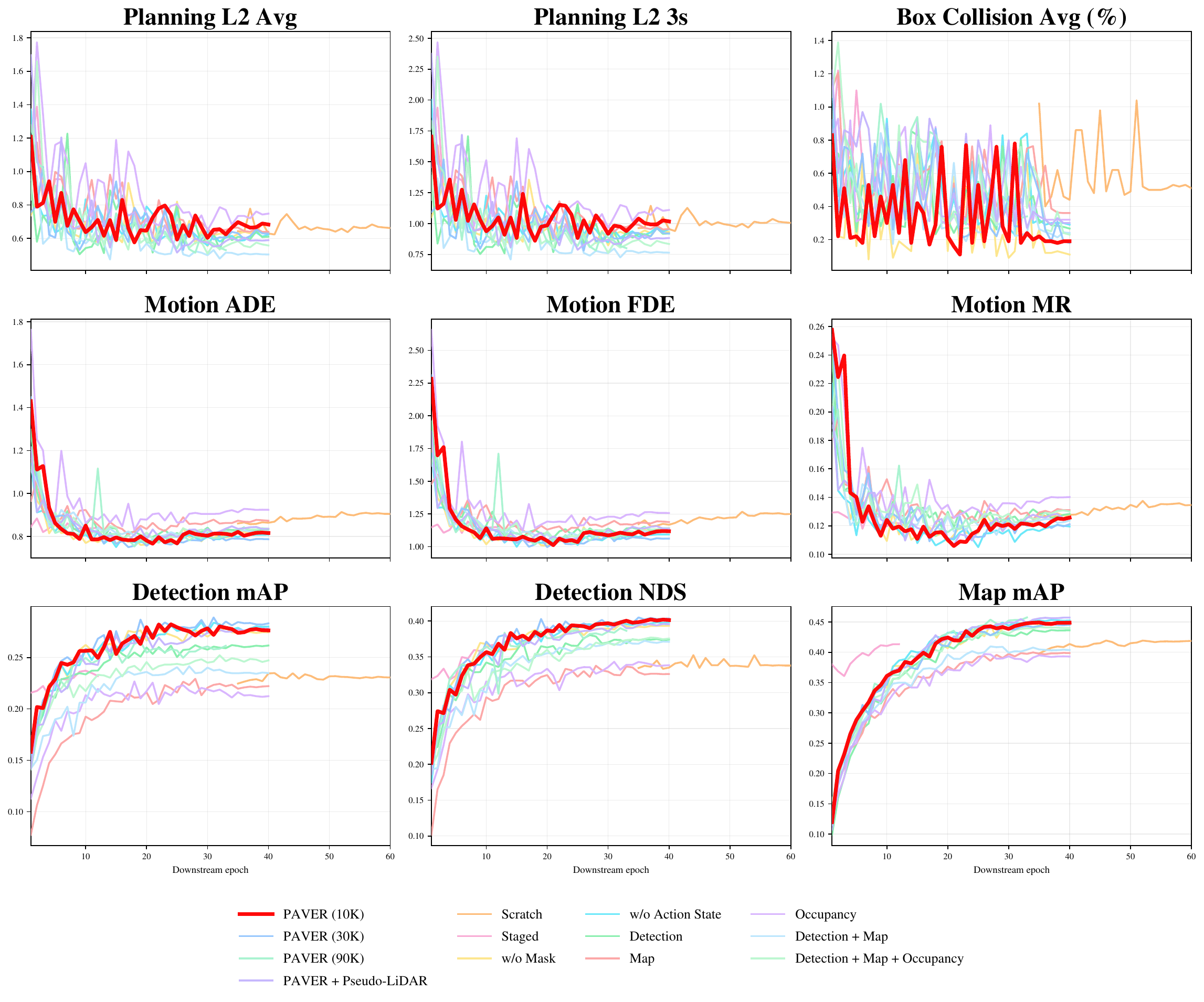}
\par\vspace{2pt}
\includegraphics[width=\textwidth]{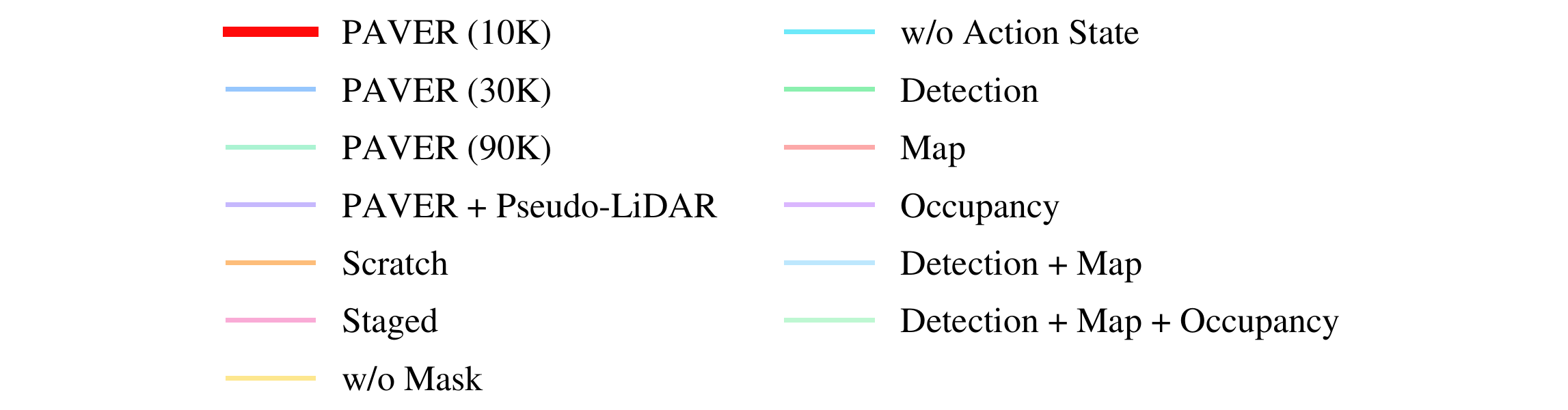}
\caption{Native per-checkpoint evaluator trajectories for 13 VAD-Tiny
experiments. Missing epochs are not interpolated; PAVER (10K) is emphasized.}
\label{fig:supp_vad_tiny_all_experiment_trajectories}
\end{figure*}

Figure~\ref{fig:supp_vad_tiny_all_experiment_trajectories} shows 13 VAD-Tiny
experiment series. Each line uses native per-checkpoint evaluator outputs only;
missing epochs are not interpolated. PAVER (10K) is emphasized for readability.
Detection and map metrics tend to improve broadly, whereas planning and
collision are noisier and non-monotonic. The trajectories are consistent with
representation refinement together with checkpoint and safety sensitivity, not
with causality or universal superiority.

\begin{figure*}[!tp]
\centering
\includegraphics[width=\textwidth]{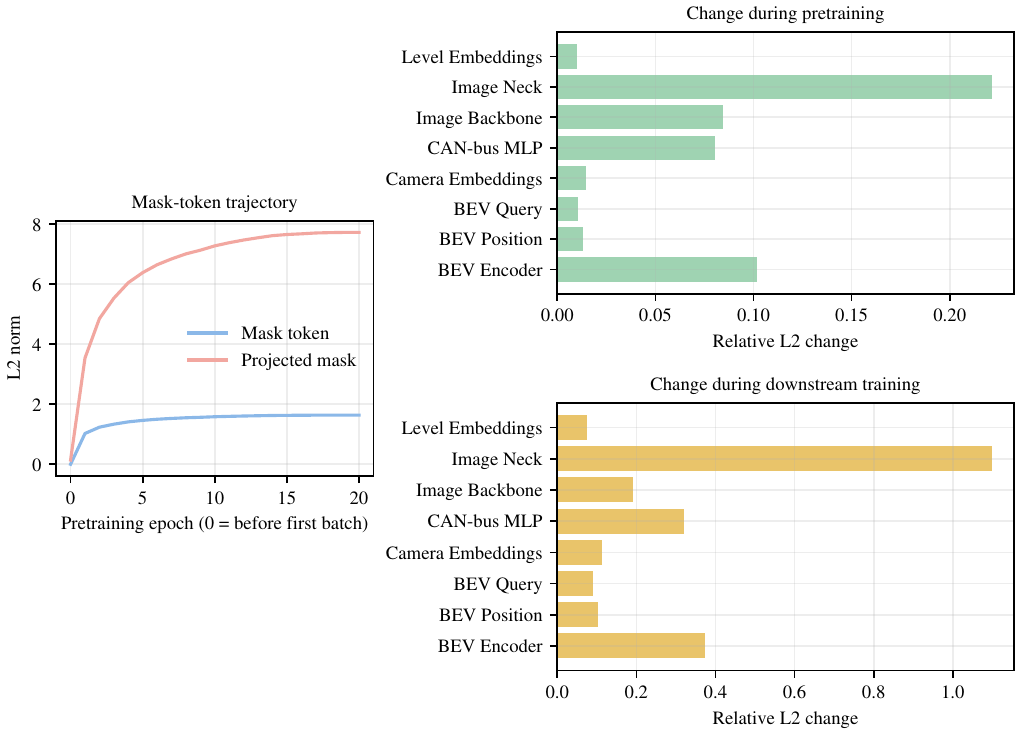}
\caption{Parameter dynamics during PAVER pretraining and downstream training.
The panels show the learned mask-token trajectory, changes from the state before
the first pretraining batch to pretraining epoch 20, and subsequent changes at
downstream epoch 30. Registered buffers are excluded.}
\label{fig:analysis_checkpoint_dynamics}
\end{figure*}

All 252 selected parameter tensors change during pretraining
(Figure~\ref{fig:analysis_checkpoint_dynamics}). Relative L2
change is 0.1017 for the BEV Encoder, 0.0845 for the Image Backbone, and 0.2215
for the Image Neck. The Image Backbone retains high directional similarity
(0.9964 cosine similarity) but has no unchanged parameter elements. This agrees
with the training configuration, which leaves the backbone trainable with a
0.2 learning-rate multiplier. The measurements establish weight-space change,
not that any one component causes the downstream gains.

\begin{figure*}[!tp]
\centering
\includegraphics[width=\textwidth]{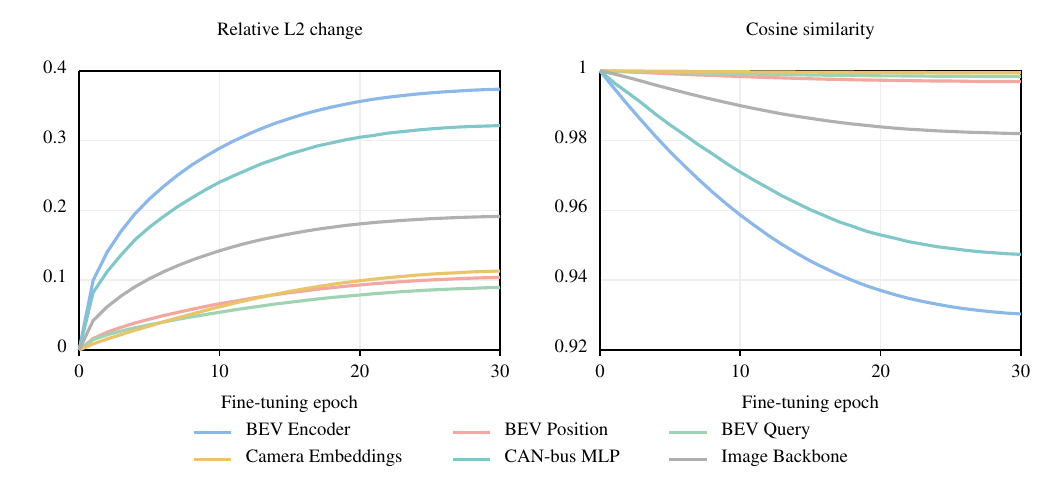}
\caption{Parameter-only changes from PAVER pretraining to downstream epoch 30
for six selected parameter groups. Left: L2 distance divided by the
pretrained parameter norm. Right: cosine similarity to the pretrained
parameters. Both panels share the centered legend; epoch 0 is the pretrained
reference. Image-neck changes are reported in the text.}
\label{fig:analysis_optimization}
\end{figure*}

Rapid early drift is compatible with useful pretraining rather than evidence
that the initialization is immediately forgotten. The encoder is free to move
because the pretraining-only auxiliary head imposes no downstream architectural constraint;
what matters is that optimization begins from a BEV in which action-relevant
evidence is already accessible.

Figure~\ref{fig:analysis_optimization} shows that, at downstream epoch 30,
relative L2 changes from the pretrained state are
0.3737 for the BEV Encoder, 0.1913 for the Image Backbone, and 1.0992 for the
Image Neck. From epochs 20 to 30, these values increase by only 0.0176, 0.0106,
and 0.0485, respectively. Thus, the transferred parameters remain trainable,
while most of their weight-space movement occurs earlier in downstream
optimization.



\subsection{Metric Structure}

The first three within-series metric components explain 92.30\% of trajectory
variance, but collision loads on a separate component and correlates only weakly
with the other utilities. Collision must therefore remain an explicit safety
axis rather than being absorbed into one balanced score.

The separate collision component helps explain why transfer is not uniform
across downstream architectures. PAVER supervises current geometric evidence,
not a future collision event; converting that evidence into a trajectory still
depends on the planner and its interaction with motion and occupancy modules
\citep{hu2022stp3,hu2023planning,jiang2023vad}. Better BEV organization can
therefore improve displacement and perception without guaranteeing the same
collision response for every decoder. We therefore retain collision explicitly
so that this decoder dependence is not hidden inside an aggregate score.

\subsection{Planning Error Tails}

Averages hide where a planner actually fails, so we rank validation samples by
their own average L2 and report the mean over the worst tenth, twentieth and
hundredth (Table~\ref{tab:planning_tails_crossmodel}). The margin PAVER holds
over each baseline widens with severity on all three architectures: on VAD-Tiny
it grows from 0.06 m across all samples to 3.47 m within the worst 1\%, on
VAD-Base from 0.18 to 3.55 m, and on GenAD from 0.05 to 1.86 m. The overall
column reproduces the planning column of the multi-task table, so the tails come
from the same checkpoints rather than from a separate evaluation.

\begin{table}[t]
\centering
\small
\renewcommand{\arraystretch}{1.02}
\tabcolsep=5.0pt
\begin{tabular}{lcccc}
\toprule
& \multicolumn{4}{c}{L2 (m) $\downarrow$} \\
\cmidrule(lr){2-5}
& All & Worst 10\% & Worst 5\% & Worst 1\% \\
\midrule
VAD-Tiny$^{\dagger}$ & 0.66 & 2.22 & 3.09 & 6.98 \\
\rowcolor{oursgray}
\textbf{+ Ours} & \textbf{0.60} & \textbf{1.71} & \textbf{2.15} & \textbf{3.51} \\
\midrule
VAD-Base & 0.74 & 2.85 & 4.28 & 10.56 \\
\rowcolor{oursgray}
\textbf{+ Ours} & \textbf{0.56} & \textbf{2.08} & \textbf{3.05} & \textbf{7.01} \\
\midrule
GenAD$^{\dagger}$ & 0.59 & 2.10 & 2.98 & 6.16 \\
\rowcolor{oursgray}
\textbf{+ Ours} & \textbf{0.54} & \textbf{1.73} & \textbf{2.30} & \textbf{4.30} \\
\bottomrule
\multicolumn{5}{@{}l}{\small $^{\dagger}$ Reproduced results.} \\
\end{tabular}
\caption{Planning results by error percentile on nuScenes. Samples are ranked by
their own average L2; each column reports the mean over that fraction of the
5,119 valid planning samples.}
\label{tab:planning_tails_crossmodel}
\end{table}

\subsection{Auxiliary Target Quality at the End of Pretraining}

\begin{table}[t]
\centering
\small
\renewcommand{\arraystretch}{1.02}
\setlength{\tabcolsep}{1.5pt}
\begin{tabular}{@{}p{0.28\columnwidth}*{5}{>{\centering\arraybackslash}p{0.12\columnwidth}}@{}}
\toprule
Pretraining target & Det. mAP & NDS & Map mAP & Occ. IoU & Occ. mIoU \\
\midrule
Detection & 0.1208 & 0.1907 & -- & -- & -- \\
Map & -- & -- & 0.0040 & -- & -- \\
Occupancy & -- & -- & -- & 0.6011 & 0.2045 \\
Det. + Map + Occ. & 0.1314 & 0.1771 & 0.0120 & 0.5579 & 0.1907 \\
\bottomrule
\end{tabular}
\caption{Auxiliary-head quality after 20 pretraining epochs. Det., Map, and Occ.\ denote pretraining on detection, on the vector map, and on occupancy, so Det.\ + Map + Occ.\ is the joint objective. Each column uses its task-native evaluator.}
\label{tab:pretraining_target_quality}
\end{table}

We evaluate each auxiliary head at pretraining epoch 20 on all 6,019
validation frames. Table~\ref{tab:pretraining_target_quality} keeps task-native
metrics separate: nuScenes detection mAP and NDS, vector-map mAP, and occupancy
completion IoU and semantic mIoU. A dash means that the corresponding head is
absent, not that its score is zero. These numbers report whether each
temporary head learned its own target at the end of pretraining.

The combined objective improves detection mAP from 0.1208 to 0.1314 and map
mAP from 0.0040 to 0.0120 relative to the corresponding single-task heads, but
its detection NDS decreases from 0.1907 to 0.1771. Occupancy completion IoU and
semantic mIoU also decrease from 0.6011 and 0.2045 to 0.5579 and 0.1907. The
mixed signs rule out the simpler explanation that adding more auxiliary tasks
uniformly improves pretraining quality. They are consistent with competition
between task losses, but do not isolate gradient interference as the cause.

\section{Target Quality and Qualitative Evidence}
\label{sec:target_qualitative_evidence}

\newsavebox{\supplementfigurebox}
\newenvironment{fitSupplementFigure}{%
  \begin{lrbox}{\supplementfigurebox}%
  \begin{minipage}{\textwidth}\centering
}{%
  \end{minipage}\end{lrbox}%
  \ifdim\dimexpr\ht\supplementfigurebox+\dp\supplementfigurebox\relax
    >\dimexpr\textheight-2pt\relax
    \resizebox*{!}{\dimexpr\textheight-2pt\relax}{\usebox{\supplementfigurebox}}%
  \else
    \usebox{\supplementfigurebox}%
  \fi
}

\newcommand{\qualitativeHeaderThree}[3]{%
  \makebox[\textwidth][c]{%
  \begin{minipage}[c]{0.48\textwidth}\centering\small\bfseries #1\end{minipage}\hfill%
  \begin{minipage}[c]{0.245\textwidth}\centering\small\bfseries #2\end{minipage}\hfill%
  \begin{minipage}[c]{0.245\textwidth}\centering\small\bfseries #3\end{minipage}}%
  \par\vspace{2pt}%
}
\newcommand{\qualitativeRowThree}[3]{%
  \makebox[\textwidth][c]{%
  \begin{minipage}[c]{0.48\textwidth}\centering\includegraphics[width=\linewidth]{#1}\end{minipage}\hfill%
  \begin{minipage}[c]{0.245\textwidth}\centering\includegraphics[width=\linewidth]{#2}\end{minipage}\hfill%
  \begin{minipage}[c]{0.245\textwidth}\centering\includegraphics[width=\linewidth]{#3}\end{minipage}}%
  \par\vspace{2pt}%
}
\newcommand{\detectionHeaderThree}[3]{%
  \makebox[\textwidth][c]{%
  \begin{minipage}[c]{0.327\textwidth}\centering\small\bfseries #1\end{minipage}\hfill%
  \begin{minipage}[c]{0.327\textwidth}\centering\small\bfseries #2\end{minipage}\hfill%
  \begin{minipage}[c]{0.327\textwidth}\centering\small\bfseries #3\end{minipage}}%
  \par\vspace{2pt}%
}
\newcommand{\detectionRowThree}[1]{%
  \makebox[\textwidth][c]{%
  \begin{minipage}[c]{0.327\textwidth}\centering\includegraphics[width=\linewidth]{#1__ground_truth}\end{minipage}\hfill%
  \begin{minipage}[c]{0.327\textwidth}\centering\includegraphics[width=\linewidth]{#1__vad_tiny}\end{minipage}\hfill%
  \begin{minipage}[c]{0.327\textwidth}\centering\includegraphics[width=\linewidth]{#1__paver}\end{minipage}}%
  \par\vspace{1.5pt}%
}
\newcommand{\qualitativeHeaderFour}[4]{%
  \makebox[\textwidth][c]{%
  \begin{minipage}[c]{0.40\textwidth}\centering\small\bfseries #1\end{minipage}\hfill%
  \begin{minipage}[c]{0.195\textwidth}\centering\small\bfseries #2\end{minipage}\hfill%
  \begin{minipage}[c]{0.195\textwidth}\centering\small\bfseries #3\end{minipage}\hfill%
  \begin{minipage}[c]{0.195\textwidth}\centering\small\bfseries #4\end{minipage}}%
  \par\vspace{2pt}%
}
\newcommand{\qualitativeRowFour}[4]{%
  \makebox[\textwidth][c]{%
  \begin{minipage}[c]{0.40\textwidth}\centering\includegraphics[width=\linewidth]{#1}\end{minipage}\hfill%
  \begin{minipage}[c]{0.195\textwidth}\centering\includegraphics[width=\linewidth]{#2}\end{minipage}\hfill%
  \begin{minipage}[c]{0.195\textwidth}\centering\includegraphics[width=\linewidth]{#3}\end{minipage}\hfill%
  \begin{minipage}[c]{0.195\textwidth}\centering\includegraphics[width=\linewidth]{#4}\end{minipage}}%
  \par\vspace{2pt}%
}
\newcommand{\qualitativeHeaderFive}[5]{%
  \makebox[\textwidth][c]{%
  \begin{minipage}[c]{0.34\textwidth}\centering\small\bfseries #1\end{minipage}\hfill%
  \begin{minipage}[c]{0.16\textwidth}\centering\small\bfseries #2\end{minipage}\hfill%
  \begin{minipage}[c]{0.16\textwidth}\centering\small\bfseries #3\end{minipage}\hfill%
  \begin{minipage}[c]{0.16\textwidth}\centering\small\bfseries #4\end{minipage}\hfill%
  \begin{minipage}[c]{0.16\textwidth}\centering\small\bfseries #5\end{minipage}}%
  \par\vspace{2pt}%
}
\newcommand{\qualitativeRowFive}[5]{%
  \makebox[\textwidth][c]{%
  \begin{minipage}[c]{0.34\textwidth}\centering\includegraphics[width=\linewidth]{#1}\end{minipage}\hfill%
  \begin{minipage}[c]{0.16\textwidth}\centering\includegraphics[width=\linewidth]{#2}\end{minipage}\hfill%
  \begin{minipage}[c]{0.16\textwidth}\centering\includegraphics[width=\linewidth]{#3}\end{minipage}\hfill%
  \begin{minipage}[c]{0.16\textwidth}\centering\includegraphics[width=\linewidth]{#4}\end{minipage}\hfill%
  \begin{minipage}[c]{0.16\textwidth}\centering\includegraphics[width=\linewidth]{#5}\end{minipage}}%
  \par\vspace{1pt}%
}
\newcommand{\detectionStackRow}[2]{%
  \makebox[\textwidth][c]{%
  \begin{minipage}[c]{0.22in}\centering\rotatebox{90}{\parbox{1.25in}{\centering\small\bfseries #1}}\end{minipage}\hspace{0.03in}%
  \begin{minipage}[c]{\dimexpr\textwidth-0.25in\relax}\centering\includegraphics[width=\linewidth]{#2}\end{minipage}}%
  \par\vspace{2.5pt}%
}
\newcommand{\detectionStackHeader}{%
  \makebox[\textwidth][c]{%
  \begin{minipage}[c]{0.22in}\mbox{}\end{minipage}\hspace{0.03in}%
  \begin{minipage}[c]{\dimexpr\textwidth-0.25in\relax}%
    \makebox[\linewidth][c]{%
      \makebox[0.3333\linewidth][c]{\small\bfseries Front Left}%
      \makebox[0.3333\linewidth][c]{\small\bfseries Front}%
      \makebox[0.3333\linewidth][c]{\small\bfseries Front Right}}%
  \end{minipage}}%
  \par\vspace{1pt}%
}
\newcommand{\detectionProjectionSample}[4]{%
  \detectionStackHeader%
  \detectionStackRow{Ground Truth}{#1}%
  \detectionStackRow{Det. Only}{#2}%
  \detectionStackRow{Det. + Map}{#3}%
  \detectionStackRow{Det. + Map + Occ.}{#4}%
}
\newcommand{\balancedRowFour}[3]{%
  \qualitativeRowFour{#1__6_cameras}{#1__ground_truth}{#1__#2}{#1__#3}%
}
\newcommand{\balancedRowFourCompact}[3]{%
  \makebox[\textwidth][c]{%
  \begin{minipage}[c]{0.40\textwidth}\centering\includegraphics[width=\linewidth]{#1__6_cameras}\end{minipage}\hfill%
  \begin{minipage}[c]{0.195\textwidth}\centering\includegraphics[width=\linewidth]{#1__ground_truth}\end{minipage}\hfill%
  \begin{minipage}[c]{0.195\textwidth}\centering\includegraphics[width=\linewidth]{#1__#2}\end{minipage}\hfill%
  \begin{minipage}[c]{0.195\textwidth}\centering\includegraphics[width=\linewidth]{#1__#3}\end{minipage}}%
  \par\vspace{1.5pt}%
}
\newcommand{\balancedRowFive}[4]{%
  \qualitativeRowFive{#1__6_cameras}{#1__ground_truth}{#1__#2}{#1__#3}{#1__#4}%
}
\newcommand{\gradCamCameraHeader}{%
  \makebox[\textwidth][c]{%
  \begin{minipage}[c]{0.20in}\mbox{}\end{minipage}\hspace{0.01in}%
  \begin{minipage}[c]{1.636in}\centering\small\bfseries Front Left\end{minipage}\hspace{0.04in}%
  \begin{minipage}[c]{1.636in}\centering\small\bfseries Front\end{minipage}\hspace{0.04in}%
  \begin{minipage}[c]{1.636in}\centering\small\bfseries Front Right\end{minipage}}%
  \par\vspace{1pt}%
}
\newcommand{\gradCamRowLabel}[1]{%
  \fontsize{10.4}{10.4}\selectfont\bfseries\mbox{#1}%
}
\newcommand{\gradCamGridRow}[4]{%
  \makebox[\textwidth][c]{%
  \begin{minipage}[c]{0.20in}\centering\rotatebox{90}{\gradCamRowLabel{#1}}\end{minipage}\hspace{0.01in}%
  \begin{minipage}[c]{1.636in}\centering\includegraphics[height=0.92in]{#2}\end{minipage}\hspace{0.04in}%
  \begin{minipage}[c]{1.636in}\centering\includegraphics[height=0.92in]{#3}\end{minipage}\hspace{0.04in}%
  \begin{minipage}[c]{1.636in}\centering\includegraphics[height=0.92in]{#4}\end{minipage}}%
  \par\vspace{0.5pt}%
}
\newcommand{\gradCamInputRowThree}[3]{%
  \gradCamGridRow{Input Image}{#1}{#2}{#3}%
}
\newcommand{\gradCamAttributionRowThree}[4]{%
  \gradCamGridRow{#1}{#2}{#3}{#4}%
}
\newcommand{\occHeaderCompact}[4]{%
  \makebox[\textwidth][c]{%
  \begin{minipage}[c]{0.351\textwidth}\centering\small\bfseries #1\end{minipage}\hspace{0.010\textwidth}%
  \begin{minipage}[c]{0.171\textwidth}\centering\small\bfseries #2\end{minipage}\hspace{0.010\textwidth}%
  \begin{minipage}[c]{0.171\textwidth}\centering\small\bfseries #3\end{minipage}\hspace{0.010\textwidth}%
  \begin{minipage}[c]{0.171\textwidth}\centering\small\bfseries #4\end{minipage}}%
  \par\vspace{1.5pt}%
}
\newcommand{\occRowCompact}[4]{%
  \makebox[\textwidth][c]{%
  \begin{minipage}[c]{0.351\textwidth}\centering\includegraphics[width=\linewidth]{#1}\end{minipage}\hspace{0.010\textwidth}%
  \begin{minipage}[c]{0.171\textwidth}\centering\includegraphics[width=\linewidth]{#2}\end{minipage}\hspace{0.010\textwidth}%
  \begin{minipage}[c]{0.171\textwidth}\centering\includegraphics[width=\linewidth]{#3}\end{minipage}\hspace{0.010\textwidth}%
  \begin{minipage}[c]{0.171\textwidth}\centering\includegraphics[width=\linewidth]{#4}\end{minipage}}%
  \par\vspace{1pt}%
}
\newcommand{\pseudoLidarGroupedHeader}{%
  \makebox[\textwidth][c]{%
  \begin{minipage}[c]{0.34\textwidth}\mbox{}\end{minipage}\hfill%
  \begin{minipage}[c]{0.325\textwidth}\centering\small\bfseries Top-down View\end{minipage}\hfill%
  \begin{minipage}[c]{0.325\textwidth}\centering\small\bfseries Oblique View\end{minipage}}%
  \par\vspace{1pt}%
  \qualitativeHeaderFive{Input Images}{LiDAR}{Pseudo-LiDAR}{LiDAR}{Pseudo-LiDAR}%
}
\newcommand{\pseudoLidarRowFive}[5]{%
  \qualitativeRowFive{#1}{#2}{#3}{#4}{#5}%
}

\subsection{Pseudo-LiDAR Target Quality}

\begin{table}[t]
\centering
\small
\renewcommand{\arraystretch}{1.05}
\setlength{\tabcolsep}{2.5pt}
\begin{tabular}{cccccc}
\toprule
AbsRel $\downarrow$ & RMSE $\downarrow$ & SILog $\downarrow$
& $\delta_1$ $\uparrow$ & $\delta_2$ $\uparrow$ & $\delta_3$ $\uparrow$ \\
\midrule
0.1286 & 5.5592 & 21.7424 & 0.9238 & 0.9593 & 0.9739 \\
\bottomrule
\end{tabular}
\caption{Pseudo-LiDAR depth quality on a 1,000-frame training audit. RMSE is in meters. The 95\% intervals are $[0.1268,0.1305]$ for AbsRel and $[0.9225,0.9250]$ for $\delta_1$.}
\label{tab:pseudo_lidar_depth_quality}
\end{table}

\newcommand{\paverClasswiseMapTable}[1]{%
\begin{table*}[#1]
\centering
\small
\renewcommand{\arraystretch}{1.02}
\setlength{\tabcolsep}{4.0pt}
\begin{tabular}{llrrrr}
\toprule
Model & Method / Pretraining & Divider & Ped. Crossing & Boundary & mAP \\
\midrule
VAD-Tiny & Scratch & 0.476 & 0.317 & 0.463 & 0.419 \\
VAD-Tiny & Staged & 0.445 & 0.352 & 0.443 & 0.413 \\
\midrule
VAD-Tiny & PAVER (10K) & 0.483 & 0.339 & 0.494 & 0.439 \\
VAD-Tiny & PAVER (30K) & 0.480 & 0.383 & 0.479 & 0.447 \\
VAD-Tiny & PAVER (90K) & 0.469 & 0.329 & 0.501 & 0.433 \\
VAD-Tiny & w/o Mask & 0.479 & 0.372 & 0.484 & 0.445 \\
VAD-Tiny & w/o Action State & 0.465 & 0.375 & 0.488 & 0.443 \\
VAD-Tiny & Pseudo-LiDAR & 0.475 & 0.378 & 0.465 & 0.439 \\
\midrule
VAD-Tiny & Detection & 0.444 & 0.336 & 0.437 & 0.406 \\
VAD-Tiny & Map & 0.402 & 0.336 & 0.434 & 0.391 \\
VAD-Tiny & Occupancy & 0.429 & 0.318 & 0.434 & 0.393 \\
VAD-Tiny & Detection + Map & 0.423 & 0.329 & 0.438 & 0.397 \\
VAD-Tiny & Detection + Map + Occupancy & 0.439 & 0.372 & 0.450 & 0.420 \\
\midrule
VAD-Base & Scratch & 0.530 & 0.440 & 0.535 & 0.502 \\
VAD-Base & PAVER & 0.532 & 0.438 & 0.523 & 0.498 \\
\midrule
GenAD & Scratch & 0.501 & 0.376 & 0.488 & 0.455 \\
GenAD & PAVER & 0.473 & 0.354 & 0.484 & 0.437 \\
\bottomrule
\end{tabular}
\caption{Vector map prediction results, aggregate and class-wise.}
\label{tab:classwise_map_transparency}
\end{table*}
}
\ifdefined\arxivpreprint
\begingroup
\edef\paverSavedTableCounter{\number\value{table}}
\stepcounter{table}
\paverClasswiseMapTable{!t}
\setcounter{table}{\paverSavedTableCounter}
\endgroup
\fi

The pseudo-LiDAR audit measures the metric-depth source used by the UniK3D
variant before any downstream training. We draw 1,000 training frames by
scene-stratified sampling, evaluate all six cameras, and pool projected points
within each frame. Predictions are evaluated in camera-depth coordinates over
1--80 m and within a 20 m LiDAR radius. No scale or shift alignment is applied.
Table~\ref{tab:pseudo_lidar_depth_quality} therefore evaluates the raw metric
predictions rather than a fitted visualization. The audit contains 6,000 camera
images and 15,425,270 valid projected points.

\begin{figure*}[t]
\centering
\pseudoLidarGroupedHeader
\pseudoLidarRowFive{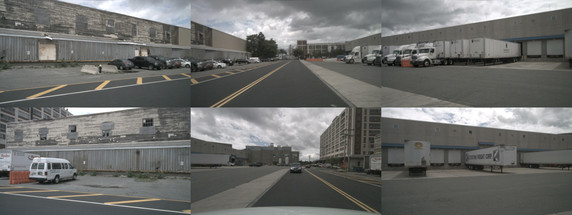}{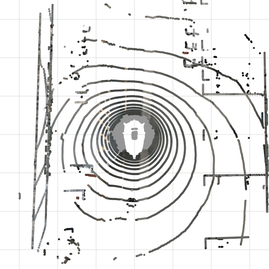}{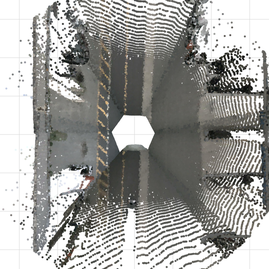}{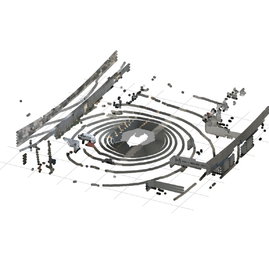}{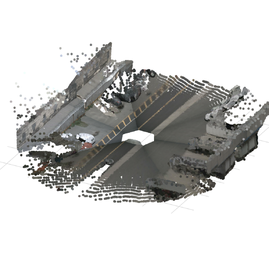}
\pseudoLidarRowFive{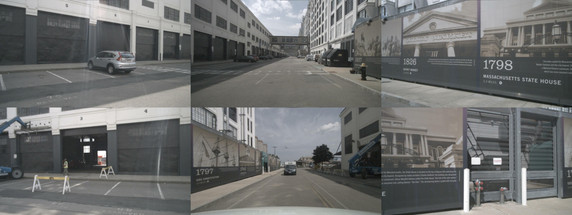}{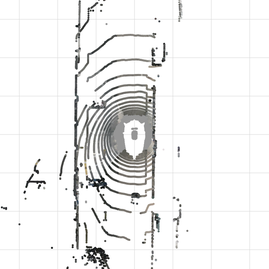}{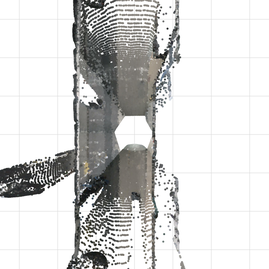}{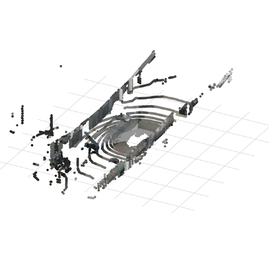}{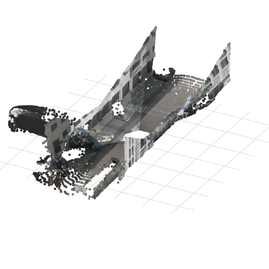}
\pseudoLidarRowFive{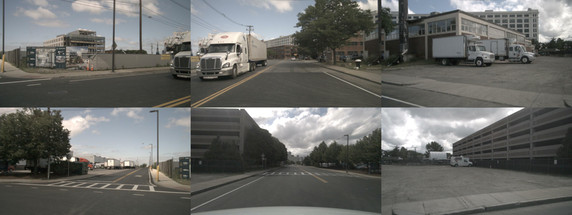}{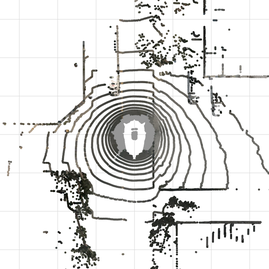}{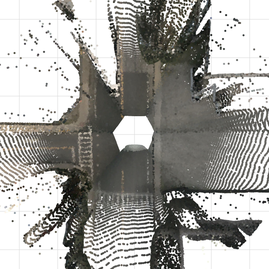}{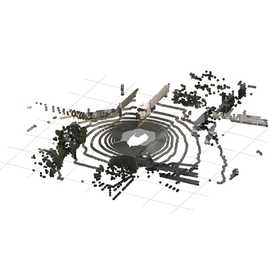}{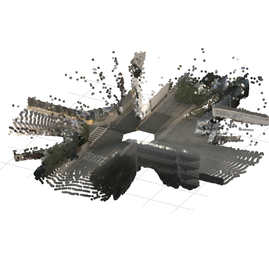}
\pseudoLidarRowFive{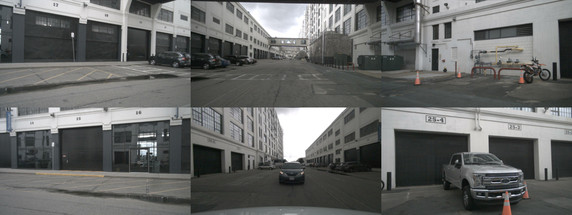}{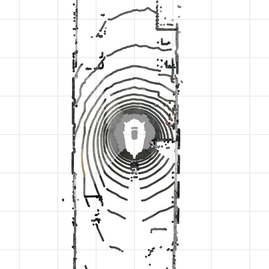}{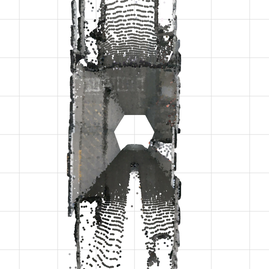}{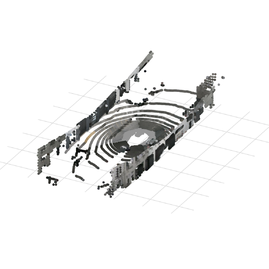}{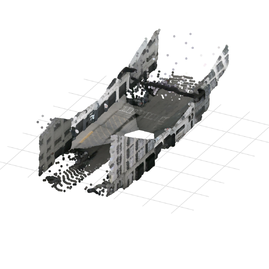}
\caption{Pseudo-LiDAR quality, set 1. Measured LiDAR against the UniK3D
pseudo-LiDAR used as the pretraining target, in top-down and oblique views.}
\label{fig:pseudo_lidar_gallery_a}
\end{figure*}

\begin{figure*}[t]
\centering
\pseudoLidarGroupedHeader
\pseudoLidarRowFive{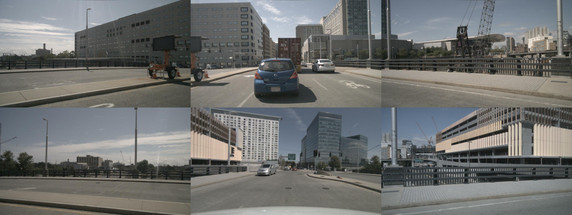}{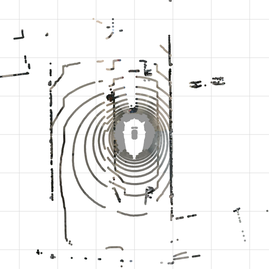}{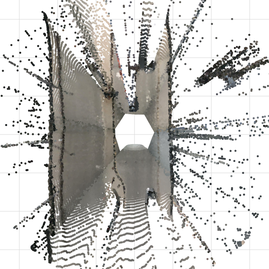}{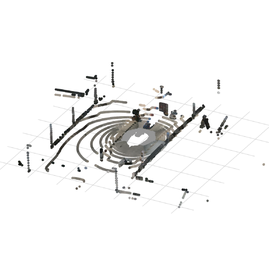}{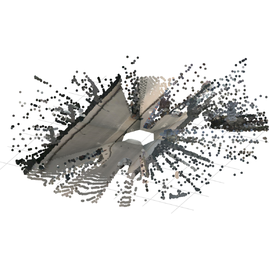}
\pseudoLidarRowFive{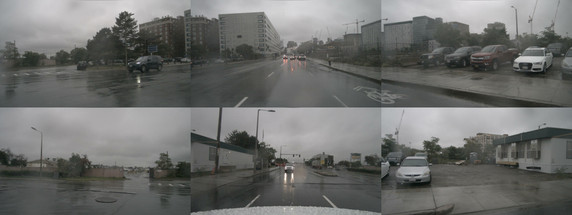}{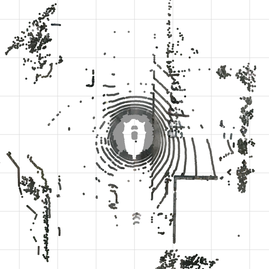}{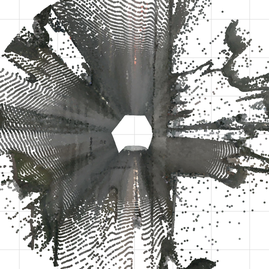}{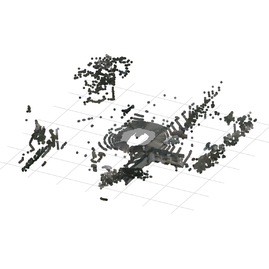}{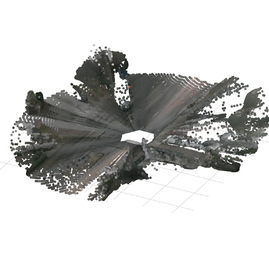}
\pseudoLidarRowFive{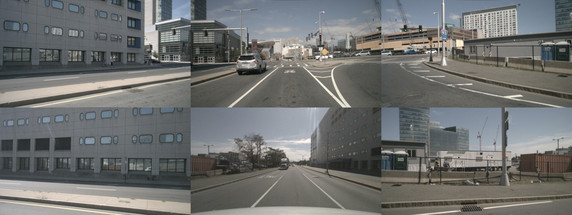}{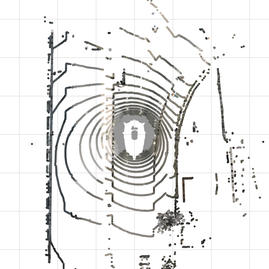}{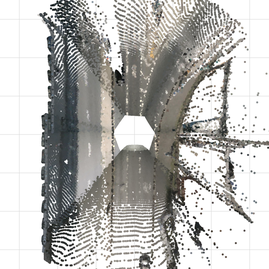}{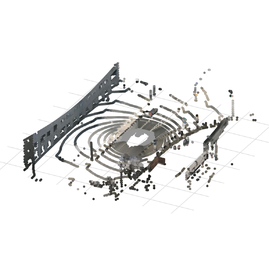}{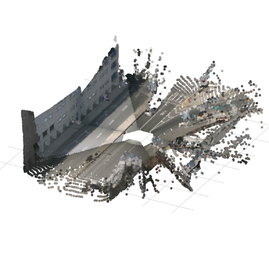}
\pseudoLidarRowFive{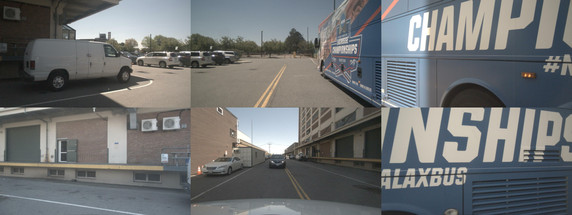}{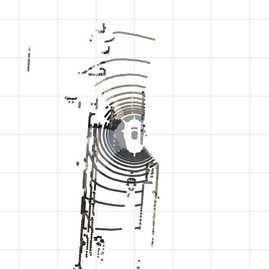}{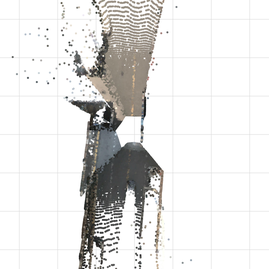}{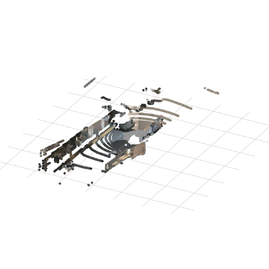}{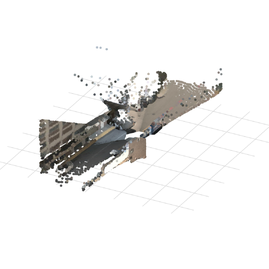}
\pseudoLidarRowFive{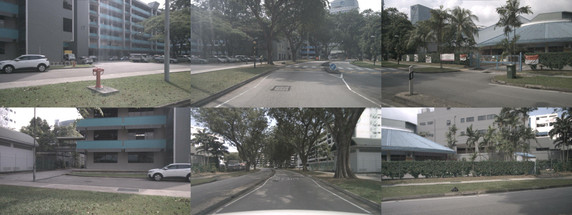}{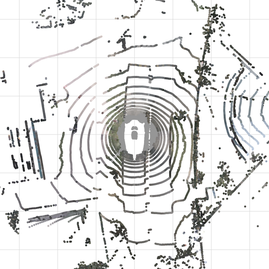}{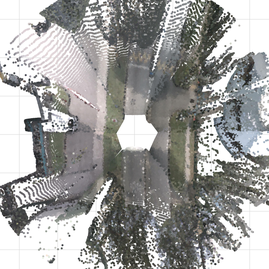}{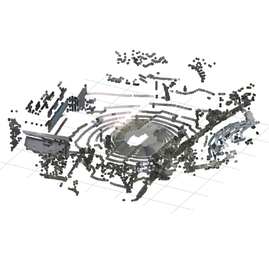}{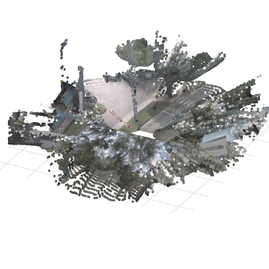}
\pseudoLidarRowFive{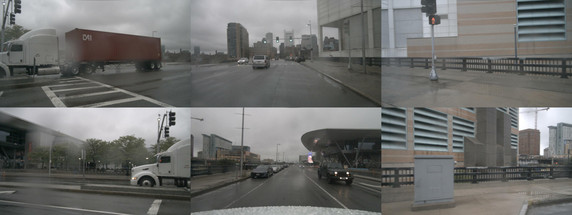}{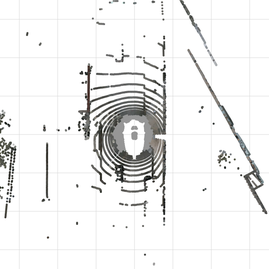}{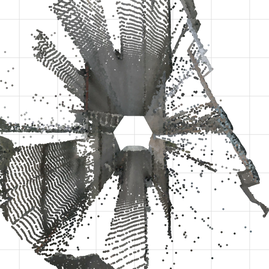}{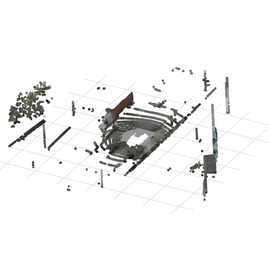}{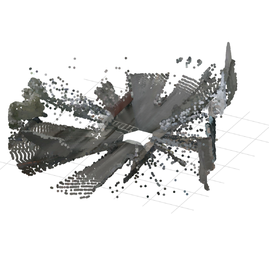}
\caption{Pseudo-LiDAR quality, set 2.}
\label{fig:pseudo_lidar_gallery_b}
\end{figure*}

The high $\delta_1$ coverage shows that most projected points preserve useful
metric scale, while the larger RMSE and SILog expose non-negligible long-tail
depth errors. This distinction matters for interpretation: pseudo-LiDAR can
provide broad geometric support without being equivalent to measured LiDAR.
Figures~\ref{fig:pseudo_lidar_gallery_a} and~\ref{fig:pseudo_lidar_gallery_b}
compare RGB-colored measured and pseudo point clouds from the same camera
frames in top-down and oblique views. We rank all audited frames by
$\delta_1$, retain the top 30, and choose 10 from that fixed pool to cover
distinct road layouts, lighting conditions, and traffic compositions.

\subsection{Capacity and Pretraining-Target Variants}

\begin{table*}[t]
\centering
\small
\renewcommand{\arraystretch}{1.02}
\setlength{\tabcolsep}{0.8pt}
\begin{tabular*}{\textwidth}{@{\extracolsep{\fill}}ll*{9}{c}@{}}
\toprule
& & \multicolumn{2}{c}{Planning} & \multicolumn{4}{c}{Motion}
& \multicolumn{2}{c}{Detection} & Map \\
\cmidrule(lr){3-4}\cmidrule(lr){5-8}\cmidrule(lr){9-10}\cmidrule(lr){11-11}
Model & Method / Pretraining & L2 $\downarrow$ & Col. (\%) $\downarrow$
& EPA $\uparrow$ & ADE $\downarrow$ & FDE $\downarrow$ & MR $\downarrow$
& mAP $\uparrow$ & NDS $\uparrow$ & mAP $\uparrow$ \\
\midrule
VAD-Tiny & Scratch & 0.662 & 0.513 & 0.458 & 0.905 & 1.250 & 0.135 & 0.231 & 0.338 & 0.419 \\
VAD-Tiny & Staged & 0.701 & 0.730 & 0.455 & 0.840 & 1.151 & 0.122 & 0.232 & 0.363 & 0.413 \\
\midrule
VAD-Tiny & PAVER (10K) & 0.603 & 0.187 & 0.498 & 0.803 & 1.086 & 0.121 & 0.278 & 0.397 & 0.439 \\
VAD-Tiny & PAVER (30K) & 0.514 & 0.320 & 0.492 & 0.800 & 1.090 & 0.125 & 0.256 & 0.383 & 0.447 \\
VAD-Tiny & PAVER (90K) & 0.596 & 0.220 & 0.509 & 0.782 & 1.056 & 0.120 & 0.280 & 0.397 & 0.433 \\
VAD-Tiny & w/o Mask & 0.705 & 0.550 & 0.499 & 0.797 & 1.080 & 0.123 & 0.276 & 0.388 & 0.445 \\
VAD-Tiny & w/o Action State & 0.655 & 0.523 & 0.508 & 0.808 & 1.084 & 0.115 & 0.274 & 0.396 & 0.443 \\
VAD-Tiny & Pseudo-LiDAR & 0.563 & 0.560 & 0.495 & 0.831 & 1.101 & 0.121 & 0.268 & 0.386 & 0.439 \\
\midrule
VAD-Tiny & Detection & 0.515 & 0.240 & 0.455 & 0.798 & 1.067 & 0.119 & 0.254 & 0.365 & 0.406 \\
VAD-Tiny & Map & 0.694 & 0.333 & 0.429 & 0.857 & 1.153 & 0.123 & 0.213 & 0.331 & 0.391 \\
VAD-Tiny & Occupancy & 0.646 & 0.320 & 0.438 & 0.905 & 1.230 & 0.136 & 0.219 & 0.336 & 0.393 \\
VAD-Tiny & Detection + Map & 0.504 & 0.190 & 0.432 & 0.805 & 1.073 & 0.117 & 0.234 & 0.361 & 0.397 \\
VAD-Tiny & Detection + Map + Occupancy & 0.577 & 0.270 & 0.457 & 0.798 & 1.082 & 0.122 & 0.243 & 0.356 & 0.420 \\
\midrule
VAD-Base & Scratch & 0.742 & 0.307 & 0.562 & 0.760 & 1.027 & 0.111 & 0.291 & 0.422 & 0.502 \\
VAD-Base & PAVER & 0.562 & 0.401 & 0.592 & 0.685 & 0.902 & 0.093 & 0.327 & 0.446 & 0.498 \\
\midrule
GenAD & Scratch & 0.591 & 0.370 & 0.413 & 0.868 & 1.184 & 0.140 & 0.188 & 0.263 & 0.455 \\
GenAD & PAVER & 0.537 & 0.207 & 0.462 & 0.800 & 1.053 & 0.114 & 0.210 & 0.277 & 0.437 \\
\bottomrule
\end{tabular*}
\caption{Planning, motion, detection, and map results on nuScenes. Planning values are 3-horizon averages, collision is in percent, and motion uses the 2.0 m matching threshold.}
\label{tab:all_fixed_checkpoint_metrics}
\end{table*}

\label{sec:downstream_variant_qualitative}

\begin{figure*}[t]
\centering
\includegraphics[width=\textwidth]{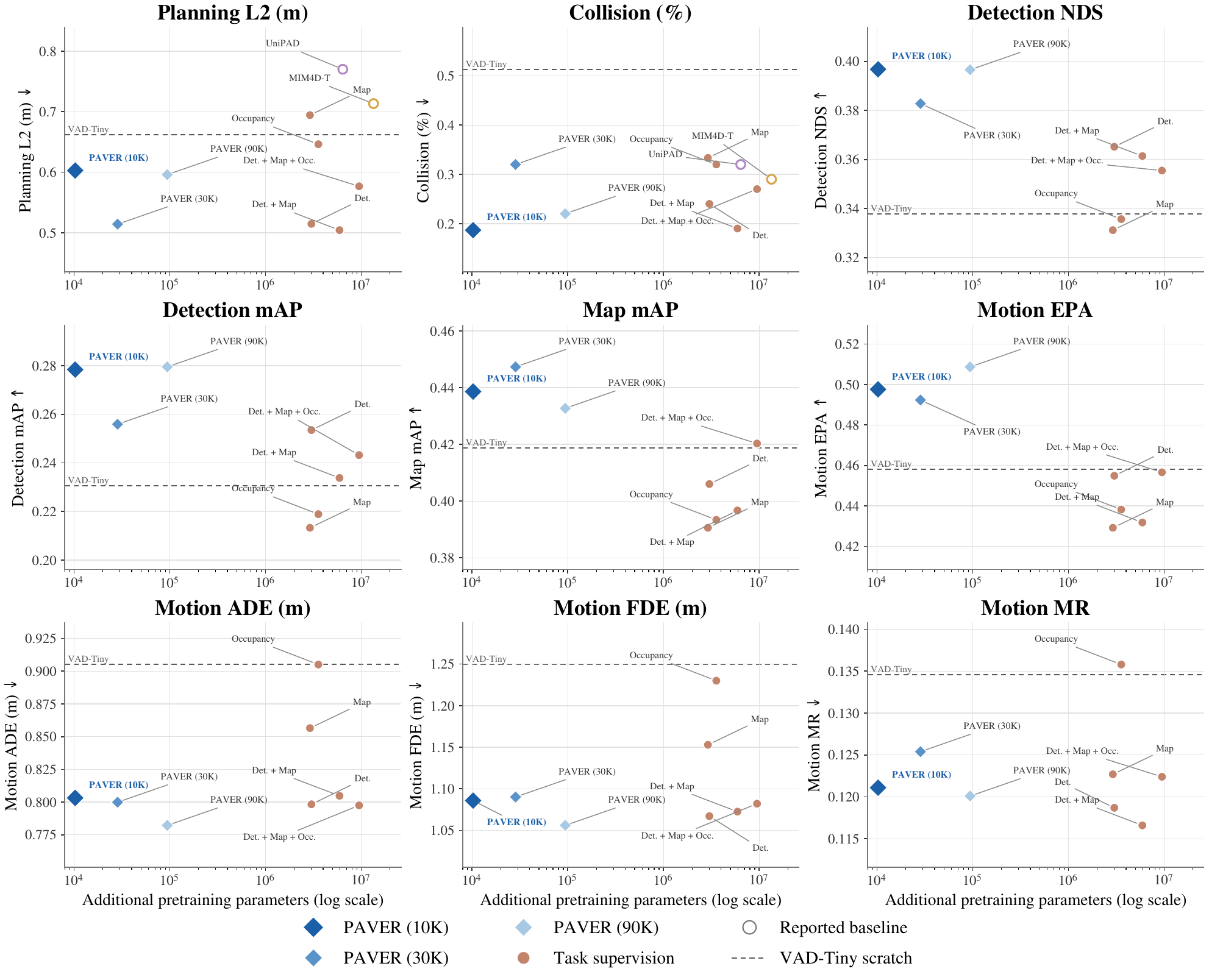}
\caption{Auxiliary capacity against every native metric. PAVER, the
task-supervised controls, and the reported UniPAD and MIM4D baselines are shown
against the VAD-Tiny scratch reference.}
\label{fig:capacity_metric_grid}
\end{figure*}

\begin{figure*}[t]
\centering
\qualitativeHeaderFive{Input Images}{Ground Truth}{VAD-Tiny}{PAVER (10K)}{PAVER (30K)}
\balancedRowFive{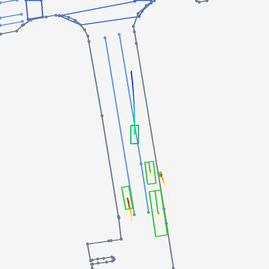}{vad_tiny}{paver_10k}{paver_30k}
\balancedRowFive{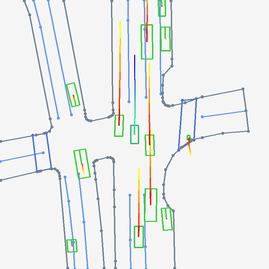}{vad_tiny}{paver_10k}{paver_30k}
\balancedRowFive{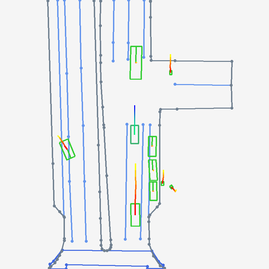}{vad_tiny}{paver_10k}{paver_30k}
\balancedRowFive{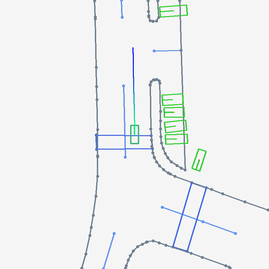}{vad_tiny}{paver_10k}{paver_30k}
\balancedRowFive{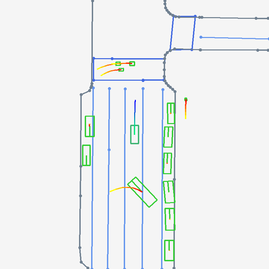}{vad_tiny}{paver_10k}{paver_30k}
\balancedRowFive{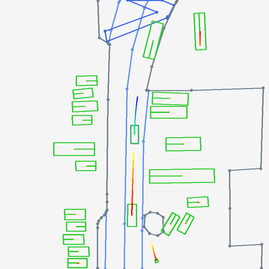}{vad_tiny}{paver_10k}{paver_30k}
\caption{Auxiliary capacity quality, PAVER (30K). After ground truth, the columns are VAD-Tiny trained from scratch and PAVER pretrained with 10K and with 30K auxiliary parameters.}
\label{fig:paver_30k_gallery_a}
\end{figure*}

\begin{figure*}[t]
\begin{fitSupplementFigure}
\centering
\qualitativeHeaderFour{Input Images}{Ground Truth}{VAD-Tiny}{PAVER (90K)}
\balancedRowFour{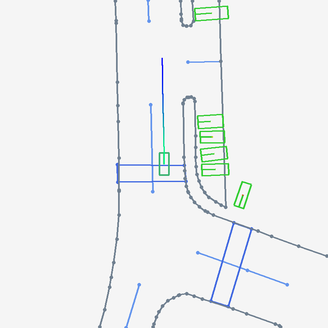}{vad_tiny}{paver_90k}
\balancedRowFour{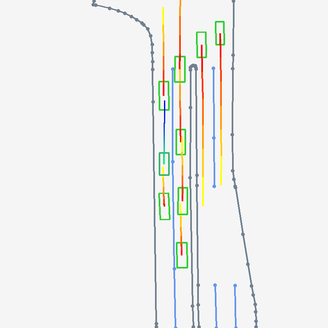}{vad_tiny}{paver_90k}
\balancedRowFour{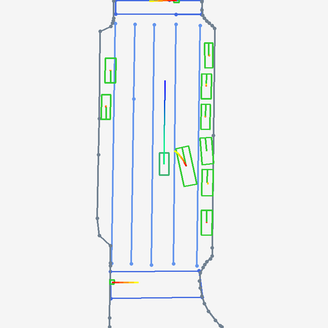}{vad_tiny}{paver_90k}
\balancedRowFour{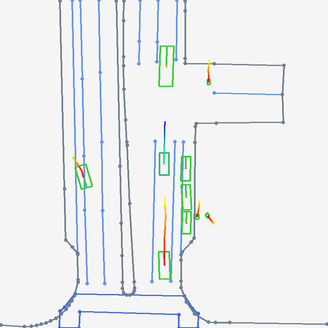}{vad_tiny}{paver_90k}
\balancedRowFour{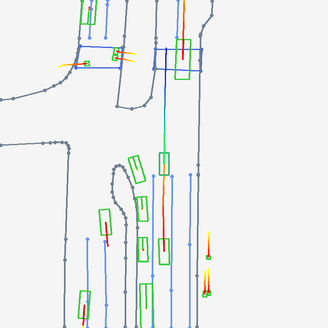}{vad_tiny}{paver_90k}
\balancedRowFour{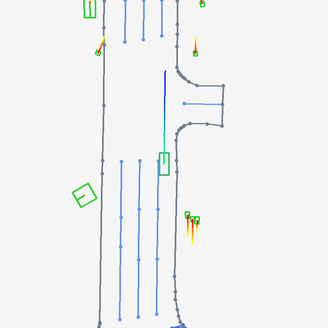}{vad_tiny}{paver_90k}
\caption{Auxiliary capacity quality, PAVER (90K). After ground truth, the columns are VAD-Tiny trained from scratch and PAVER pretrained with 90K auxiliary parameters.}
\label{fig:paver_90k_gallery_a}
\end{fitSupplementFigure}
\end{figure*}

\begin{figure*}[t]
\begin{fitSupplementFigure}
\centering
\qualitativeHeaderFour{Input Images}{Ground Truth}{VAD-Tiny}{PAVER (Pseudo-LiDAR)}
\balancedRowFour{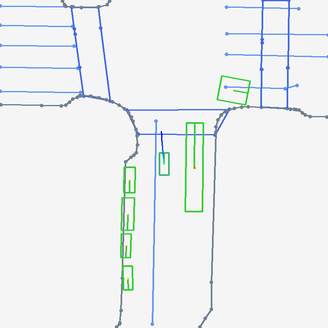}{vad_tiny}{paver_pseudo_lidar}
\balancedRowFour{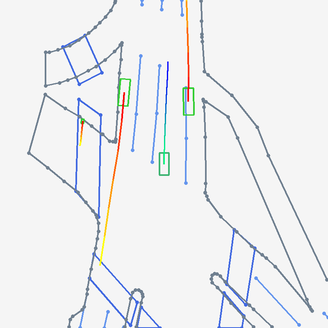}{vad_tiny}{paver_pseudo_lidar}
\balancedRowFour{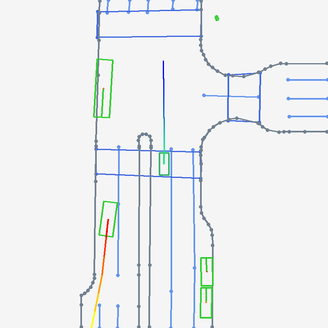}{vad_tiny}{paver_pseudo_lidar}
\balancedRowFour{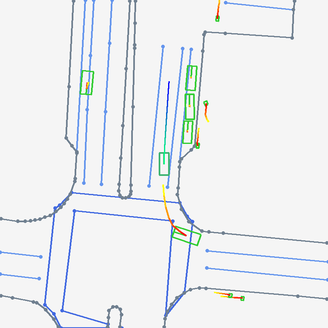}{vad_tiny}{paver_pseudo_lidar}
\balancedRowFour{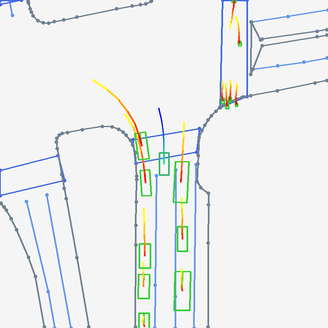}{vad_tiny}{paver_pseudo_lidar}
\balancedRowFour{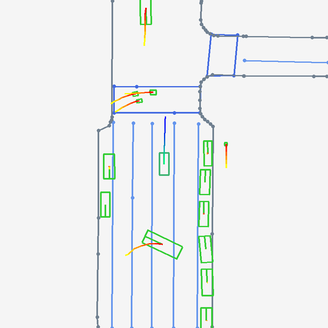}{vad_tiny}{paver_pseudo_lidar}
\caption{Pretraining-target quality, pseudo-LiDAR. After ground truth, the columns are VAD-Tiny trained from scratch and PAVER pretrained with pseudo-LiDAR as the target source. Pseudo-LiDAR replaces measured LiDAR as the input to the target builder, while the sparse action-target formulation and pretraining objective remain unchanged.}
\label{fig:paver_pseudolidar_gallery_a}
\end{fitSupplementFigure}
\end{figure*}

\begin{figure*}[t]
\begin{fitSupplementFigure}
\centering
\qualitativeHeaderFour{Input Images}{Ground Truth}{VAD-Tiny}{Det. + Map + Occ.}
\balancedRowFour{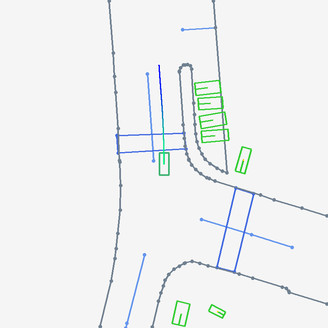}{vad_tiny}{det_map_occ}
\balancedRowFour{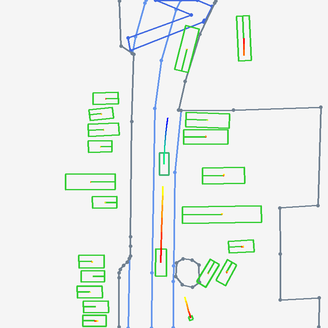}{vad_tiny}{det_map_occ}
\balancedRowFour{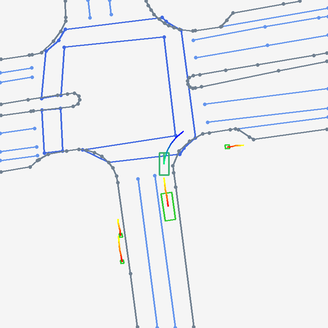}{vad_tiny}{det_map_occ}
\balancedRowFour{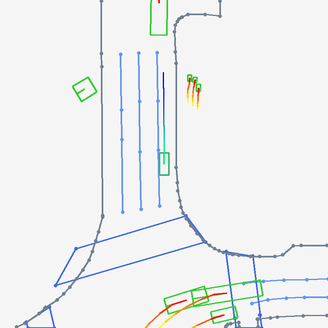}{vad_tiny}{det_map_occ}
\balancedRowFour{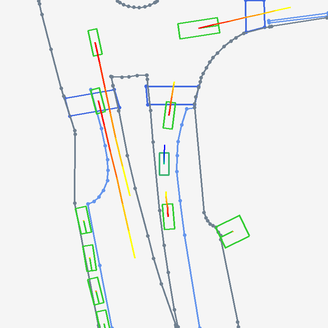}{vad_tiny}{det_map_occ}
\balancedRowFour{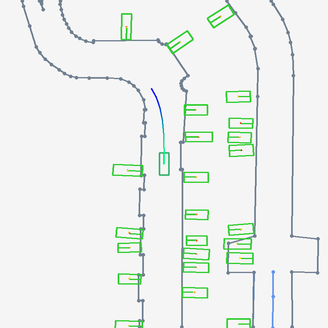}{vad_tiny}{det_map_occ}
\caption{Pretraining-target quality, detection, map and occupancy. After ground truth, the columns are VAD-Tiny trained from scratch and a model pretrained on detection, vector map, and occupancy jointly.}
\label{fig:det_map_occ_gallery_a}
\end{fitSupplementFigure}
\end{figure*}

Table~\ref{tab:all_fixed_checkpoint_metrics} reports the fixed checkpoints used
for the capacity and target comparisons,
Figure~\ref{fig:capacity_metric_grid} plots auxiliary capacity against every
native metric, and Figures~\ref{fig:paver_30k_gallery_a},
\ref{fig:paver_90k_gallery_a}, \ref{fig:paver_pseudolidar_gallery_a}
and~\ref{fig:det_map_occ_gallery_a} show the corresponding downstream
predictions. Increasing the auxiliary capacity does
not produce a monotonic multi-task gain. PAVER (30K) gives the lowest planning
L2 and the highest map mAP in this group. PAVER (90K) has the best ADE and
detection mAP, whereas PAVER (10K) has the lowest collision and the highest
exact NDS (0.3968 versus 0.3966 for PAVER (90K), although both display as
0.397). Pseudo-LiDAR and joint detection--map--occupancy targets also reduce
L2 relative to scratch training, but their motion, collision, and perception
metrics move in different directions. These mixed signs rule out a simple
explanation based on auxiliary capacity or target complexity alone.

\subsection{Cross-Model Transfer Evidence}
\label{sec:cross_model_qualitative}

\begin{figure*}[t]
\begin{fitSupplementFigure}
\centering
\qualitativeHeaderFour{Input Images}{Ground Truth}{VAD-Base}{+ PAVER}
\balancedRowFour{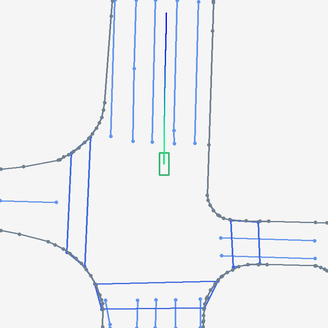}{vad_base}{paver}
\balancedRowFour{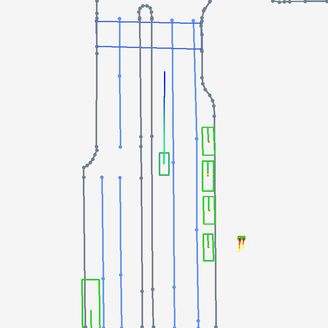}{vad_base}{paver}
\balancedRowFour{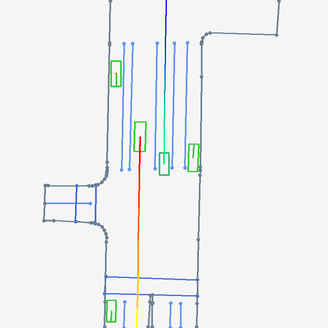}{vad_base}{paver}
\balancedRowFour{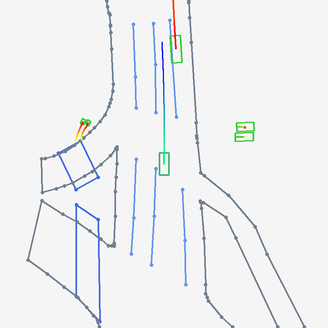}{vad_base}{paver}
\balancedRowFour{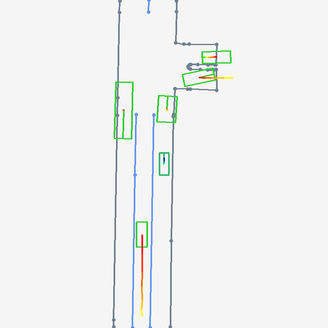}{vad_base}{paver}
\balancedRowFour{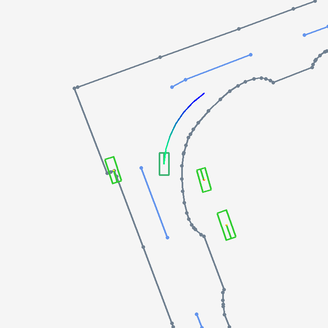}{vad_base}{paver}
\caption{Cross-model quality, VAD-Base. After ground truth, the columns are the
reproduced epoch-60 model and the same architecture initialized with PAVER pretraining.}
\label{fig:vad_base_gallery_a}
\end{fitSupplementFigure}
\end{figure*}

\begin{figure*}[t]
\begin{fitSupplementFigure}
\centering
\qualitativeHeaderFour{Input Images}{Ground Truth}{GenAD}{+ PAVER}
\balancedRowFour{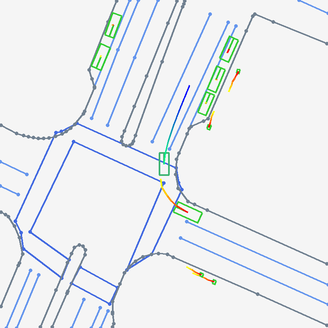}{genad}{paver}
\balancedRowFour{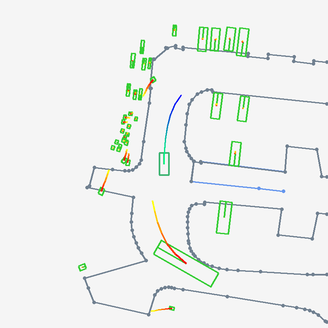}{genad}{paver}
\balancedRowFour{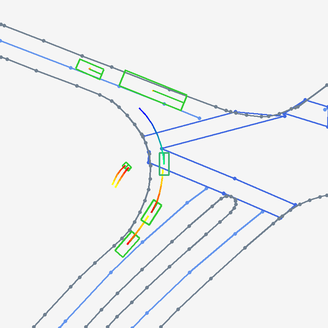}{genad}{paver}
\balancedRowFour{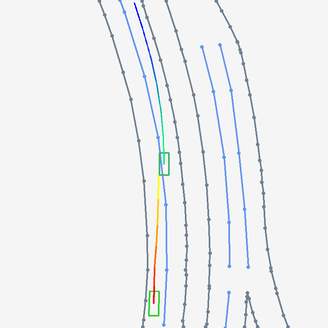}{genad}{paver}
\balancedRowFour{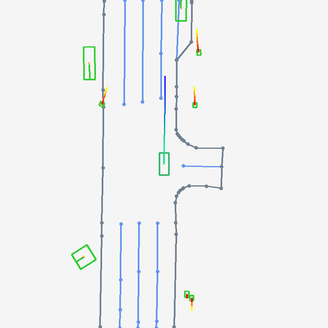}{genad}{paver}
\balancedRowFour{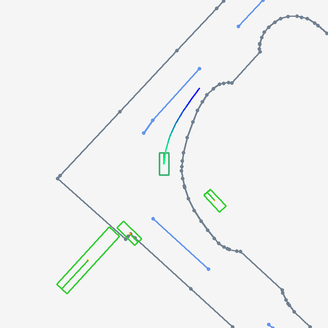}{genad}{paver}
\caption{Cross-model quality, GenAD. After ground truth, the columns are GenAD trained from scratch and the same model initialized with PAVER pretraining.}
\label{fig:genad_gallery_a}
\end{fitSupplementFigure}
\end{figure*}

Table~\ref{tab:all_fixed_checkpoint_metrics} reports the fixed VAD-Tiny,
VAD-Base, and GenAD checkpoints used in the main paper, and
Figures~\ref{fig:vad_base_gallery_a} and~\ref{fig:genad_gallery_a} show the
VAD-Base and GenAD predictions with and without PAVER initialization. On VAD-Tiny, PAVER improves planning,
collision, motion, detection, NDS, and map mAP. On VAD-Base, PAVER reduces average L2,
improves motion and detection, raises collision, and leaves map mAP essentially
unchanged at the displayed precision.
On GenAD, it improves planning, collision, motion, detection, and NDS, but map
mAP decreases. The different metric signs are consistent with a shared BEV
initialization whose downstream effect remains architecture dependent; they do
not support a uniform improvement claim.

\subsection{Class-wise Perception Results}

\ifdefined\arxivpreprint
\stepcounter{table}
\else
\paverClasswiseMapTable{t}
\fi

\begin{table*}[t]
\centering
\small
\renewcommand{\arraystretch}{1.02}
\setlength{\tabcolsep}{0.8pt}
\begin{tabular*}{\textwidth}{@{\extracolsep{\fill}}ll*{12}{c}@{}}
\toprule
Model & Method / Pretraining & mAP & NDS & Car & Truck & C.Veh. & Bus & Trail. & Barr.
& Motor. & Bicy. & Ped. & Cone \\
\midrule
VAD-Tiny & Scratch & 0.231 & 0.338 & 0.492 & 0.189 & 0.060 & 0.283 & 0.005 & 0.338 & 0.126 & 0.171 & 0.275 & 0.367 \\
VAD-Tiny & Staged & 0.232 & 0.363 & 0.485 & 0.168 & 0.059 & 0.338 & 0.004 & 0.336 & 0.126 & 0.162 & 0.299 & 0.345 \\
\midrule
VAD-Tiny & PAVER (10K) & 0.278 & 0.397 & 0.509 & 0.237 & 0.109 & 0.414 & 0.010 & 0.323 & 0.217 & 0.227 & 0.346 & 0.391 \\
VAD-Tiny & PAVER (30K) & 0.256 & 0.383 & 0.514 & 0.245 & 0.060 & 0.335 & 0.009 & 0.298 & 0.213 & 0.215 & 0.334 & 0.335 \\
VAD-Tiny & PAVER (90K) & 0.280 & 0.397 & 0.527 & 0.220 & 0.099 & 0.369 & 0.013 & 0.318 & 0.258 & 0.237 & 0.335 & 0.419 \\
VAD-Tiny & w/o Mask & 0.276 & 0.388 & 0.506 & 0.217 & 0.081 & 0.394 & 0.012 & 0.316 & 0.236 & 0.268 & 0.344 & 0.388 \\
VAD-Tiny & w/o Action State & 0.274 & 0.396 & 0.513 & 0.213 & 0.130 & 0.332 & 0.009 & 0.312 & 0.248 & 0.243 & 0.340 & 0.399 \\
VAD-Tiny & Pseudo-LiDAR & 0.268 & 0.386 & 0.503 & 0.227 & 0.081 & 0.323 & 0.018 & 0.303 & 0.232 & 0.281 & 0.340 & 0.377 \\
\midrule
VAD-Tiny & Detection & 0.254 & 0.365 & 0.500 & 0.197 & 0.041 & 0.359 & 0.005 & 0.334 & 0.157 & 0.243 & 0.315 & 0.385 \\
VAD-Tiny & Map & 0.213 & 0.331 & 0.475 & 0.180 & 0.043 & 0.228 & 0.005 & 0.289 & 0.155 & 0.182 & 0.254 & 0.322 \\
VAD-Tiny & Occupancy & 0.219 & 0.336 & 0.457 & 0.189 & 0.033 & 0.240 & 0.007 & 0.289 & 0.122 & 0.214 & 0.297 & 0.339 \\
VAD-Tiny & Detection + Map & 0.234 & 0.361 & 0.502 & 0.146 & 0.061 & 0.314 & 0.003 & 0.312 & 0.171 & 0.176 & 0.284 & 0.369 \\
VAD-Tiny & Detection + Map + Occ. & 0.243 & 0.356 & 0.513 & 0.184 & 0.078 & 0.260 & 0.004 & 0.298 & 0.170 & 0.246 & 0.301 & 0.379 \\
\midrule
VAD-Base & Scratch & 0.291 & 0.422 & 0.548 & 0.240 & 0.081 & 0.332 & 0.017 & 0.358 & 0.244 & 0.267 & 0.378 & 0.442 \\
VAD-Base & PAVER & 0.327 & 0.446 & 0.550 & 0.243 & 0.124 & 0.460 & 0.029 & 0.385 & 0.311 & 0.293 & 0.423 & 0.452 \\
\midrule
GenAD & Scratch & 0.188 & 0.263 & 0.443 & 0.000 & 0.077 & 0.000 & 0.000 & 0.298 & 0.169 & 0.224 & 0.290 & 0.373 \\
GenAD & PAVER & 0.210 & 0.277 & 0.459 & 0.000 & 0.102 & 0.000 & 0.000 & 0.314 & 0.226 & 0.249 & 0.342 & 0.412 \\
\bottomrule
\end{tabular*}
\caption{3D object detection results, aggregate and class-wise. C.Veh., Trail., Barr., Motor., Bicy., Ped., and Cone denote construction vehicle, trailer, barrier, motorcycle, bicycle, pedestrian, and traffic cone. A dash marks a class the evaluator did not report.}
\label{tab:classwise_detection_transparency}
\end{table*}

\label{sec:classwise_detection_transparency}

Table~\ref{tab:classwise_detection_transparency} reports every available
nuScenes detection class AP for the fixed checkpoints used in the component,
capacity, auxiliary-target, pseudo-LiDAR, and cross-model comparisons. The
VAD-Tiny (scratch) values are taken from the preserved native epoch-60
detection summary, including its per-class evaluator output. For every complete row, the mean
of the displayed class AP values agrees with the reported mAP within 0.0006;
the residual difference is source rounding.

The class-level results show redistribution rather than uniform gains. PAVER
(90K) gives the strongest VAD-Tiny aggregate detection result, while component
removal and auxiliary-target changes favor different object classes. Map-only
pretraining is weakest in aggregate detection. VAD-Base changes are mixed,
whereas GenAD improves aggregate detection and every class with nonzero AP
under PAVER initialization.

\begin{figure*}[t]
\centering
\detectionProjectionSample{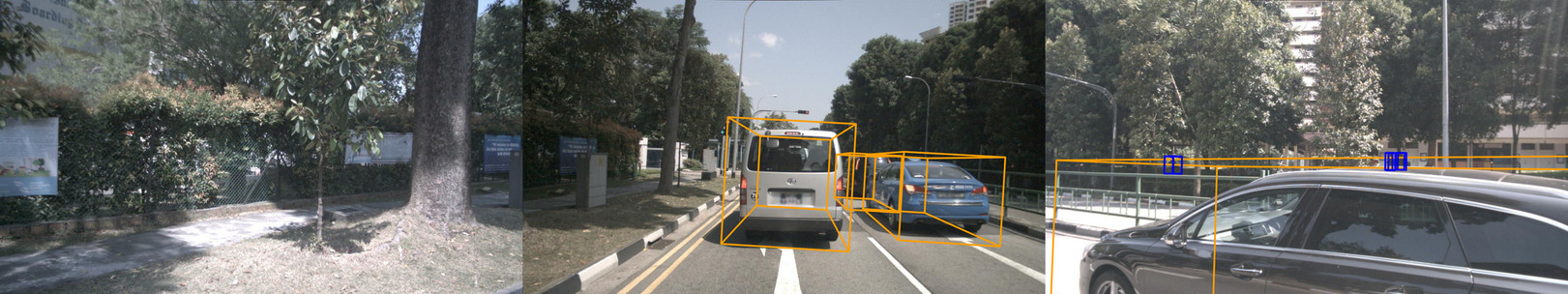}{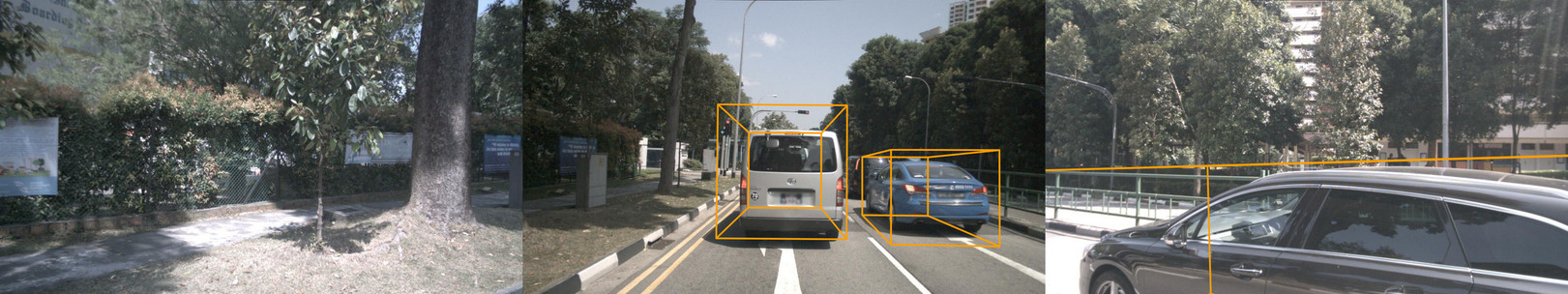}{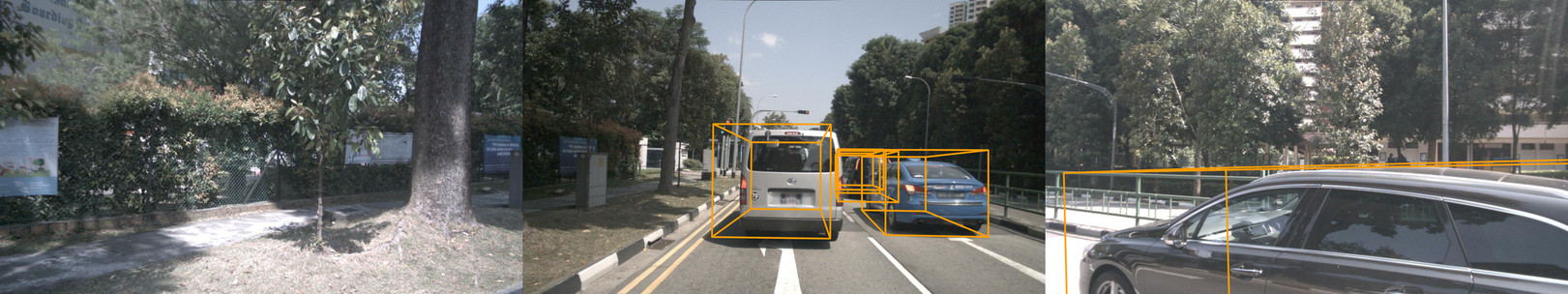}{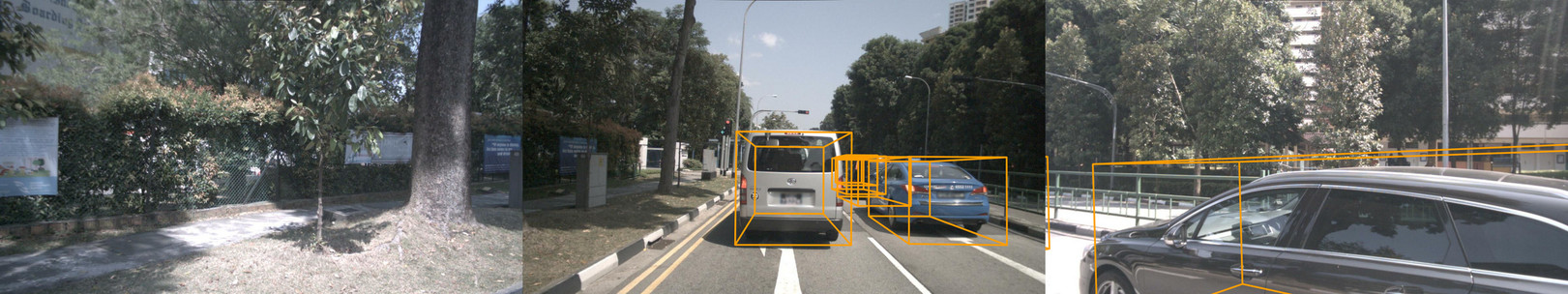}
\caption{Detection pretraining quality. Each panel concatenates the front-left, front, and front-right camera views. Ground-truth and predicted 3D boxes are shown for detection-only, detection--map, and detection--map--occupancy pretraining.}
\label{fig:pretraining_detection_gallery_a}
\end{figure*}

\begin{figure*}[t]
\centering
\detectionProjectionSample{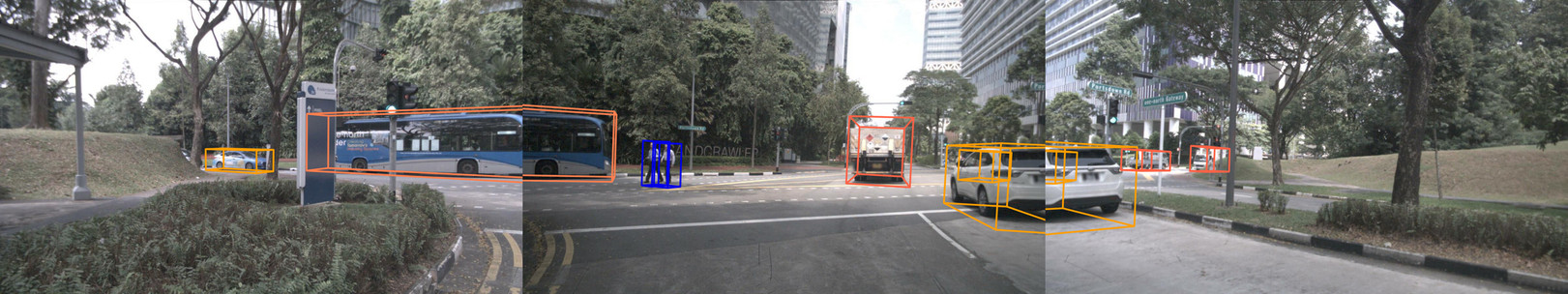}{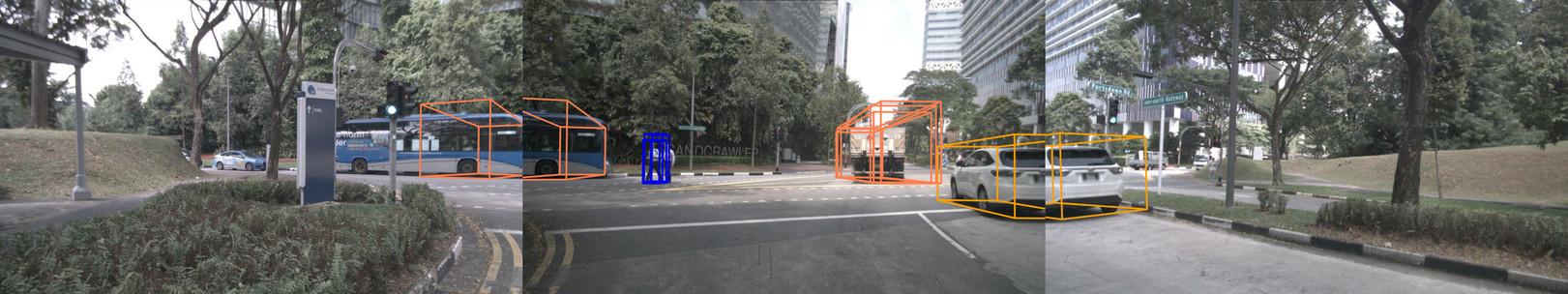}{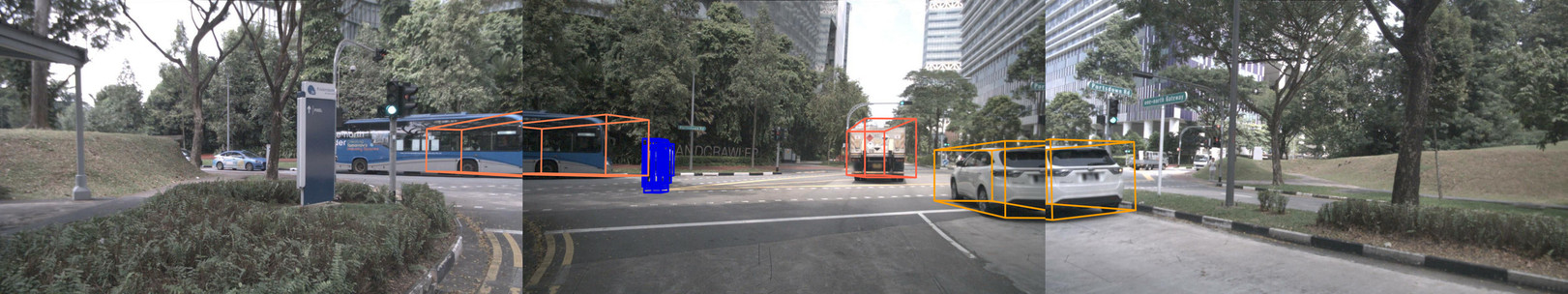}{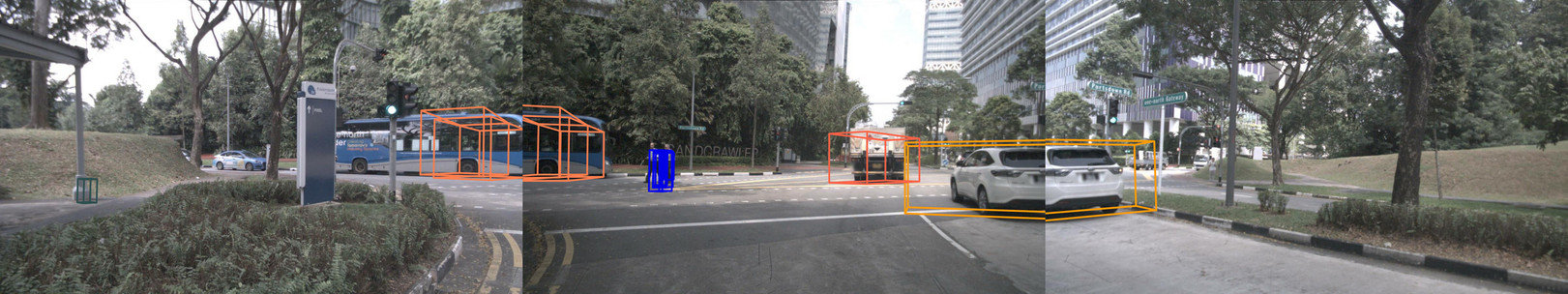}
\caption{Detection pretraining quality, sample 4.}
\label{fig:pretraining_detection_gallery_d}
\end{figure*}

\begin{figure*}[t]
\centering
\detectionProjectionSample{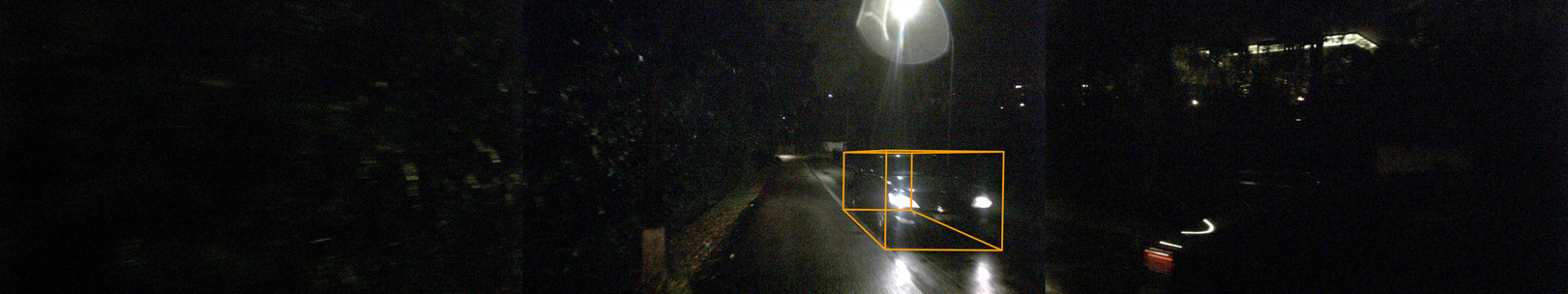}{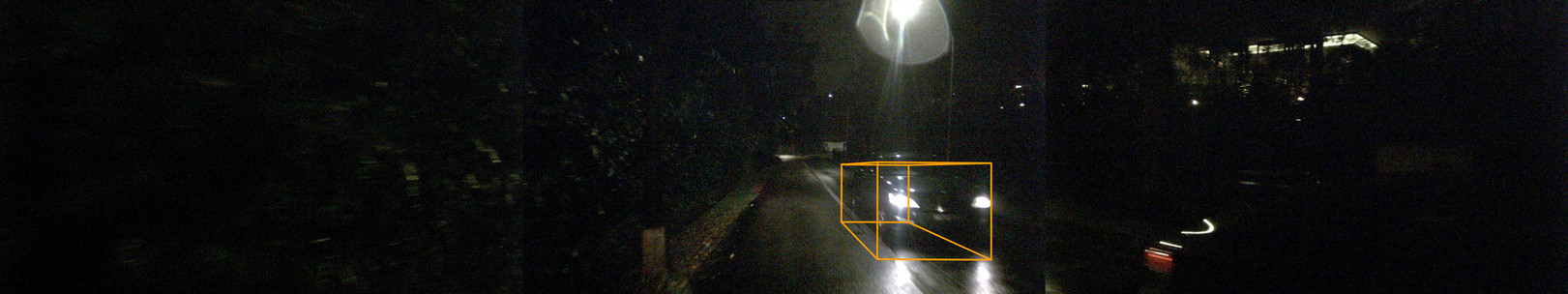}{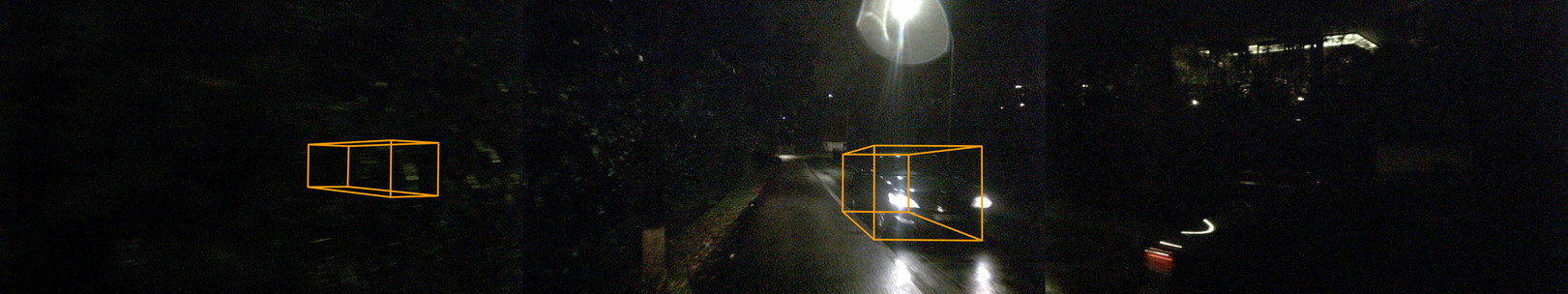}{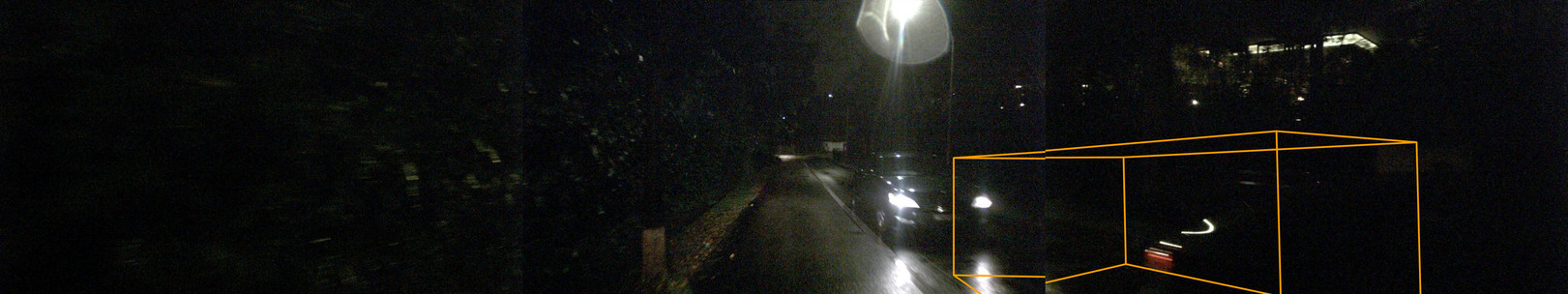}
\caption{Detection pretraining quality, sample 5.}
\label{fig:pretraining_detection_gallery_e}
\end{figure*}

\begin{figure*}[t]
\centering
\detectionProjectionSample{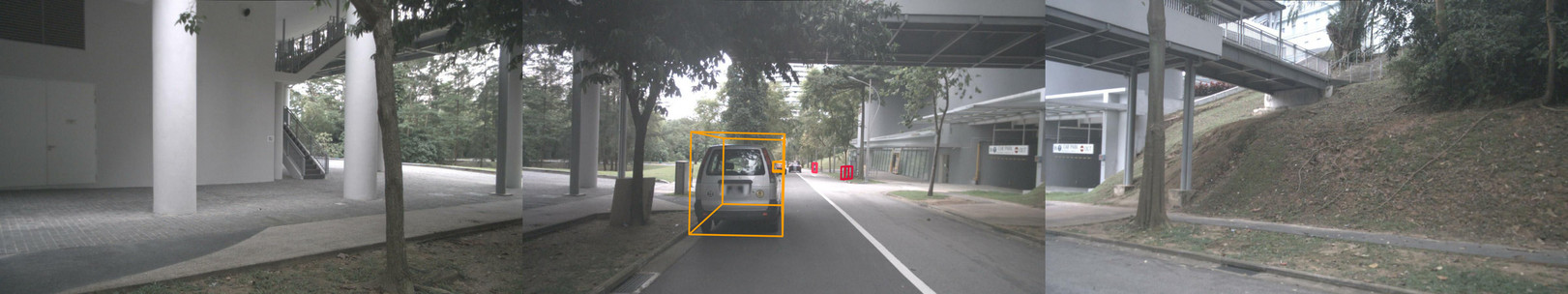}{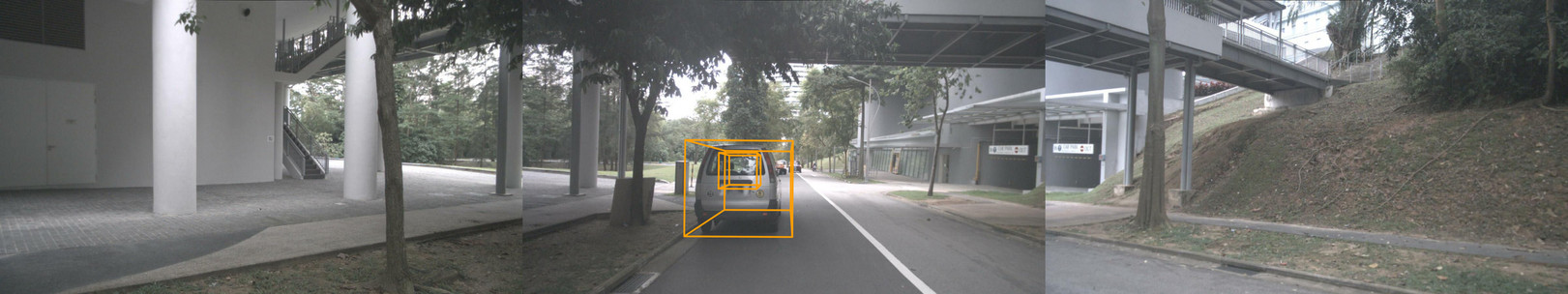}{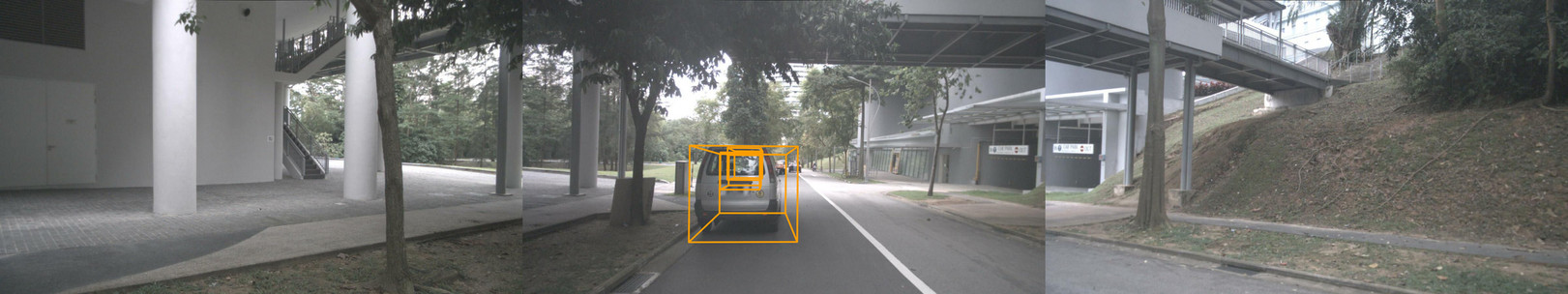}{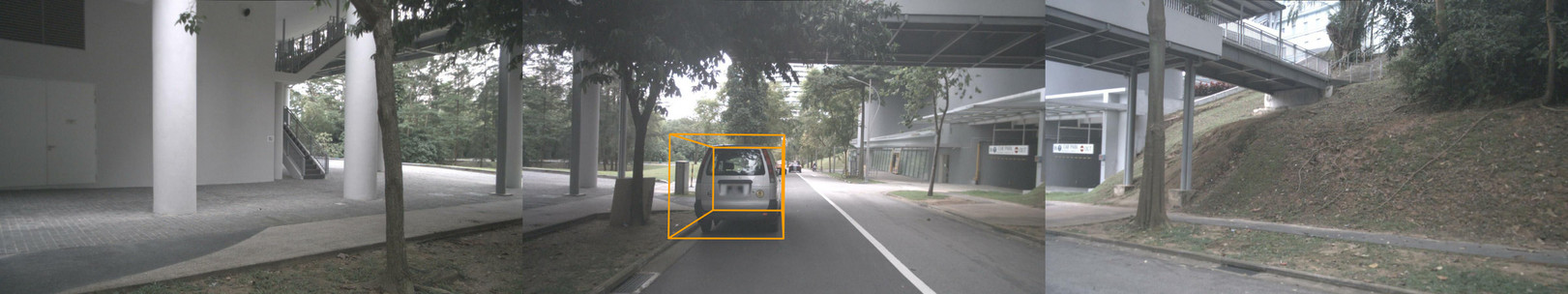}
\caption{Detection pretraining quality, sample 6.}
\label{fig:pretraining_detection_gallery_f}
\end{figure*}

\begin{figure*}[t]
\centering
\detectionProjectionSample{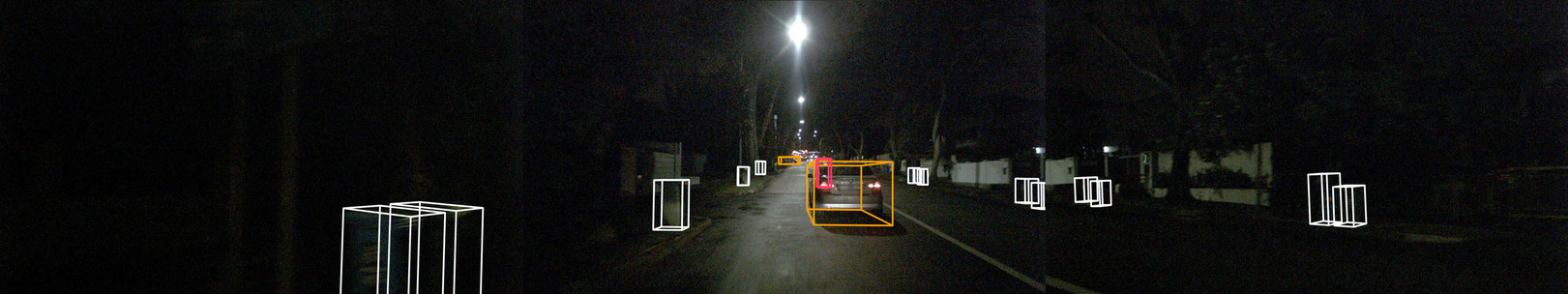}{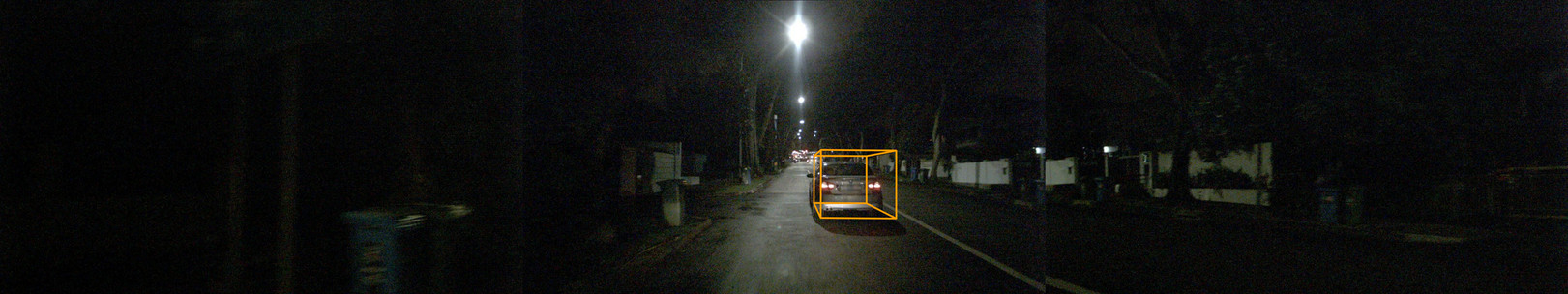}{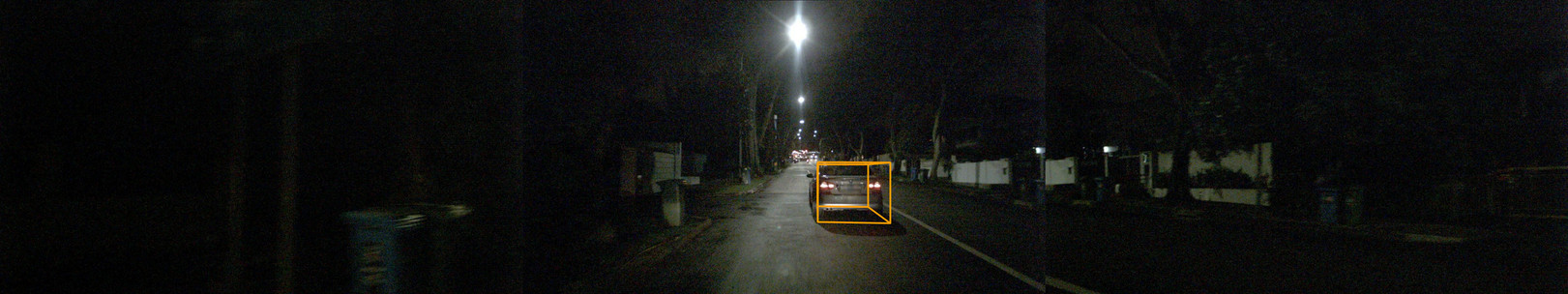}{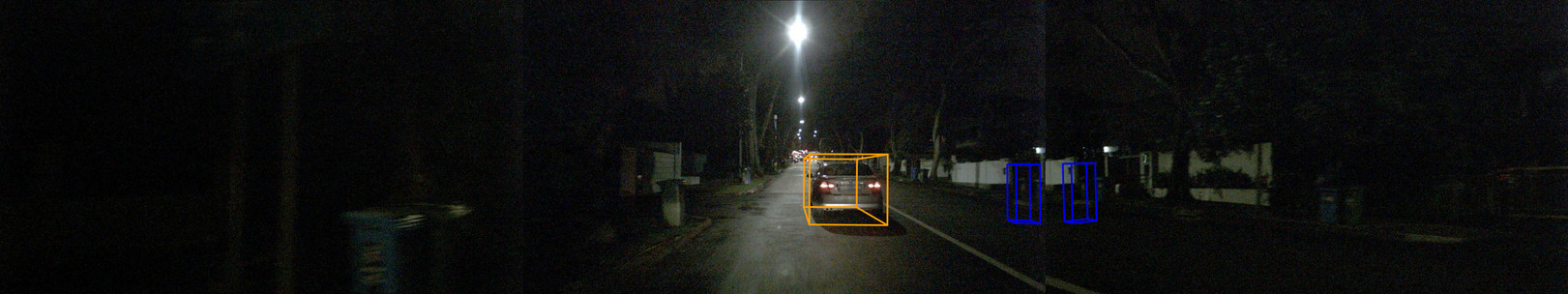}
\caption{Detection pretraining quality, sample 7.}
\label{fig:pretraining_detection_gallery_g}
\end{figure*}

\begin{figure*}[t]
\centering
\qualitativeHeaderFive{Input Images}{Ground Truth}{Map Only}{Det. + Map}{Det. + Map + Occ.}
\qualitativeRowFive{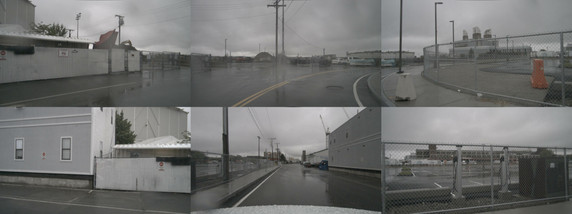}{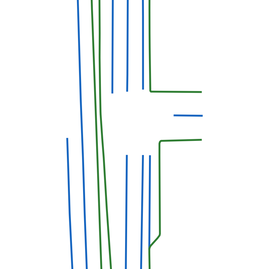}{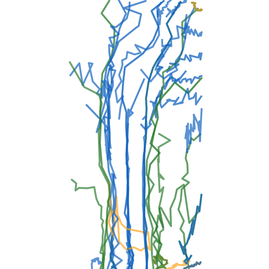}{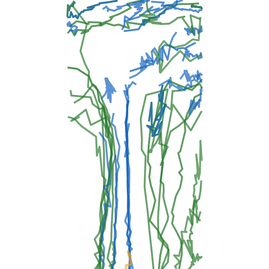}{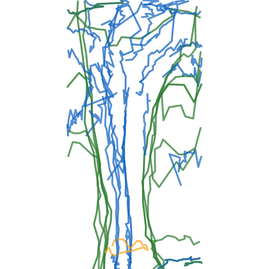}
\qualitativeRowFive{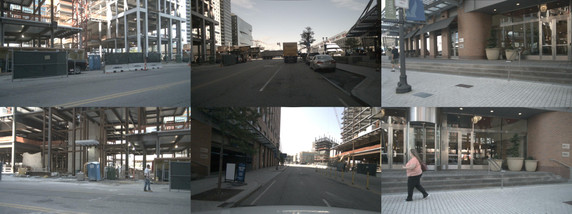}{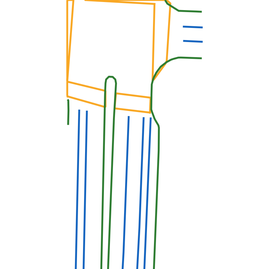}{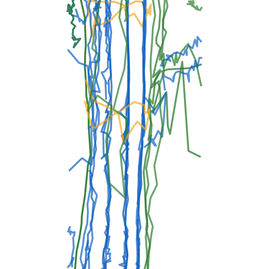}{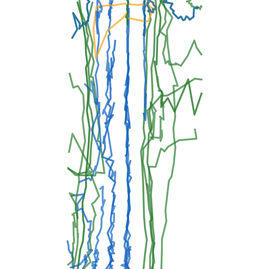}{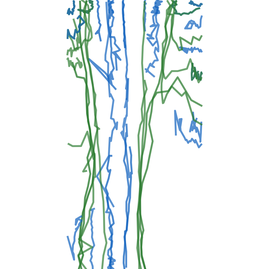}
\qualitativeRowFive{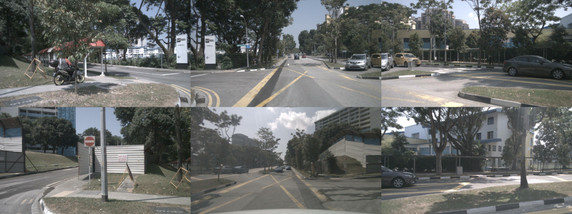}{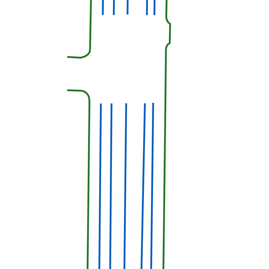}{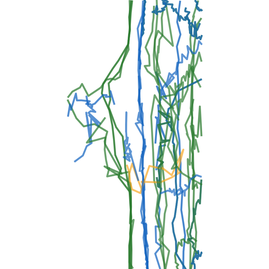}{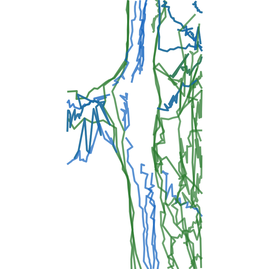}{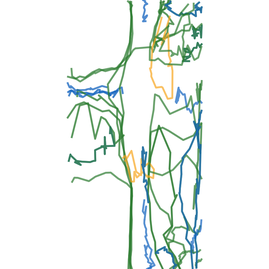}
\qualitativeRowFive{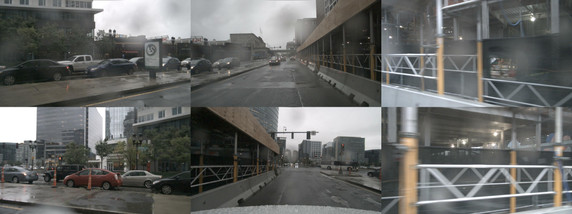}{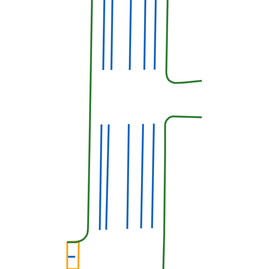}{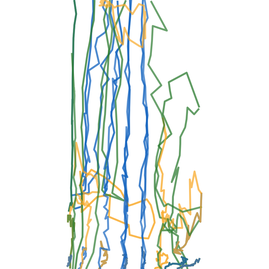}{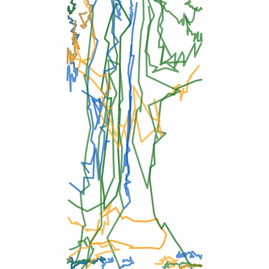}{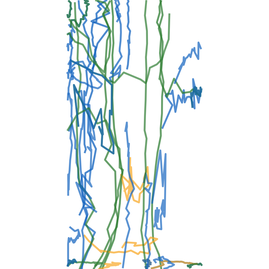}
\qualitativeRowFive{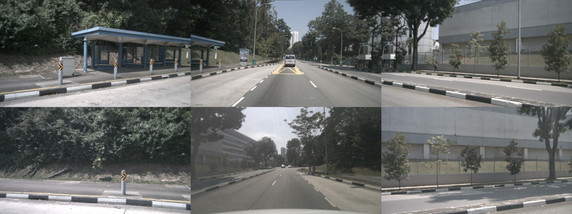}{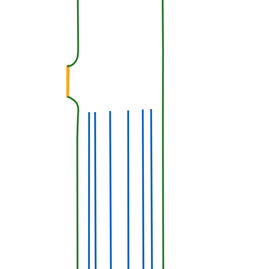}{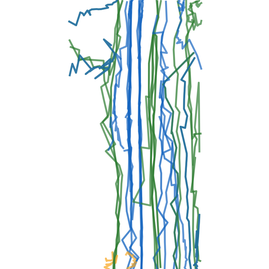}{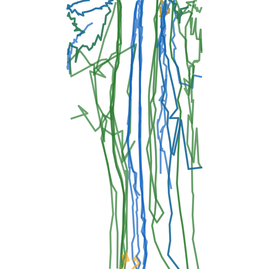}{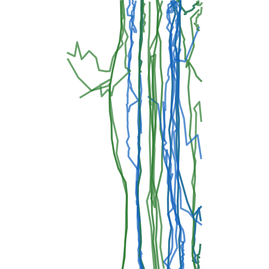}
\qualitativeRowFive{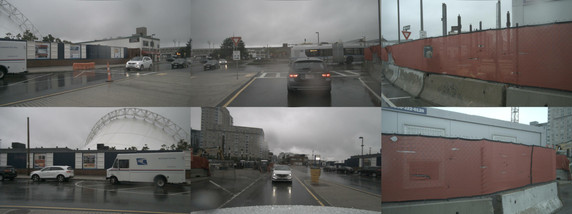}{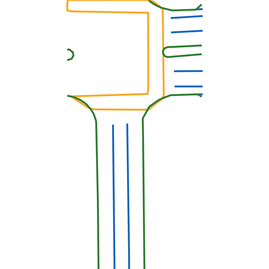}{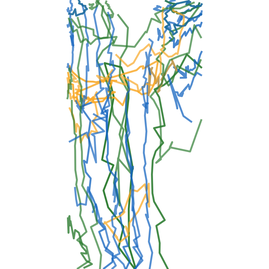}{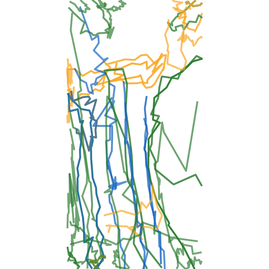}{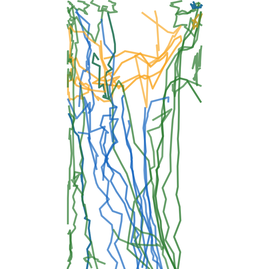}
\qualitativeRowFive{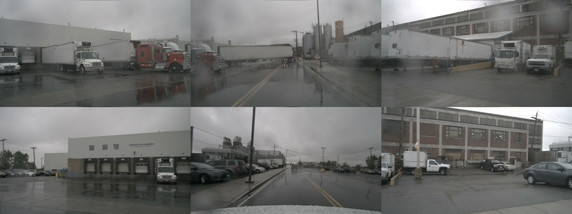}{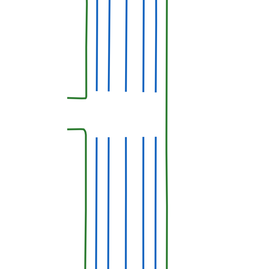}{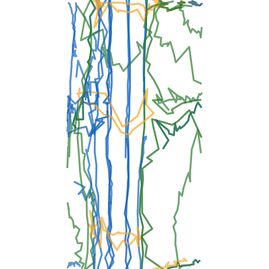}{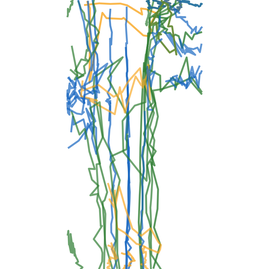}{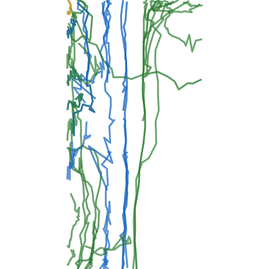}
\caption{Map pretraining quality after the shared 20-epoch budget. Columns
compare map-only, detection--map, and detection--map--occupancy pretraining
against ground truth. The fixed 20-epoch budget is not sufficient for the map
task to converge or reach its best attainable quality, so these outputs should
be interpreted as intermediate pretraining states.}
\label{fig:pretraining_map_gallery_a}
\end{figure*}

Table~\ref{tab:classwise_map_transparency} applies the same fixed-checkpoint
policy to vector-map prediction. The VAD-Tiny (scratch) class values come
from the preserved native epoch-60 map evaluation. Increasing the
temporary PAVER capacity does not yield a monotonic map trend: PAVER (30K)
has the highest VAD-Tiny map mAP among the three capacities, whereas PAVER (90K)
is lower than PAVER (10K). The component and auxiliary-target rows likewise
move the three map classes differently, which argues against attributing transfer
to uniformly better vector-map reconstruction.
Figures~\ref{fig:pretraining_detection_gallery_a}--\ref{fig:pretraining_detection_gallery_g},
\ref{fig:pretraining_map_gallery_a},
\ifdefined\arxivpreprint
and~\ref{fig:pretraining_occupancy_gallery_a}
\else
\ref{fig:pretraining_occupancy_gallery_a} and~\ref{fig:pretraining_occupancy_gallery_b}
\fi
show what each auxiliary head reconstructs at the end of its 20-epoch budget.

\subsection{Component Ablation Metrics}
\label{sec:component_ablation_metrics}

\begin{figure*}[t]
\centering
\detectionHeaderThree{Ground Truth}{VAD-Tiny}{+ PAVER}
\detectionRowThree{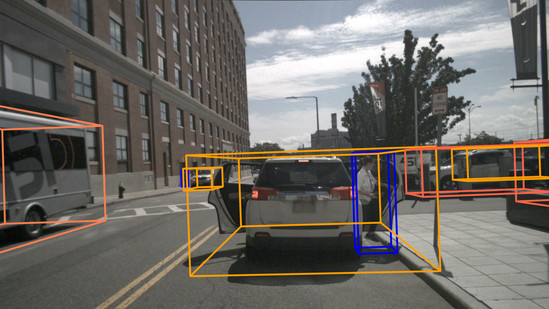}
\detectionRowThree{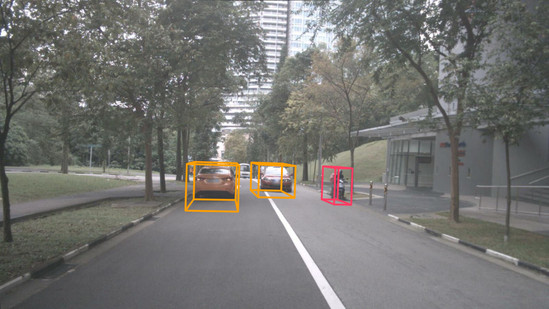}
\detectionRowThree{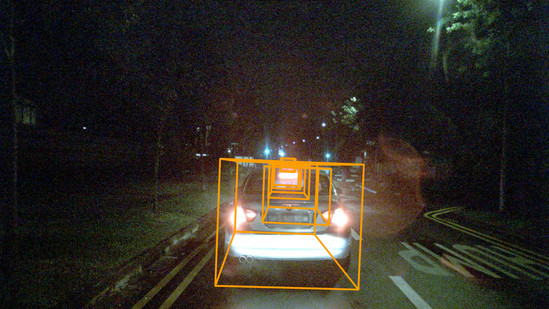}
\detectionRowThree{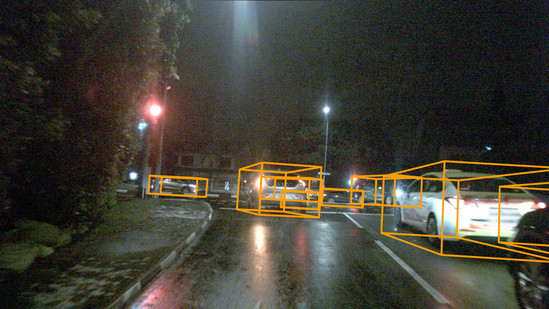}
\detectionRowThree{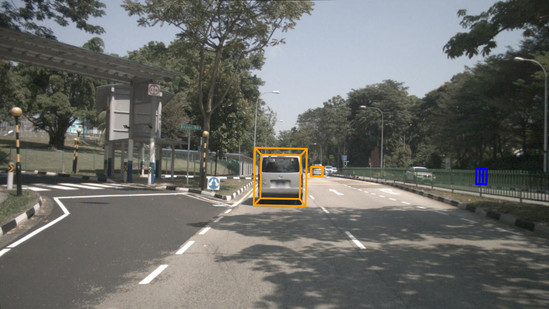}
\detectionRowThree{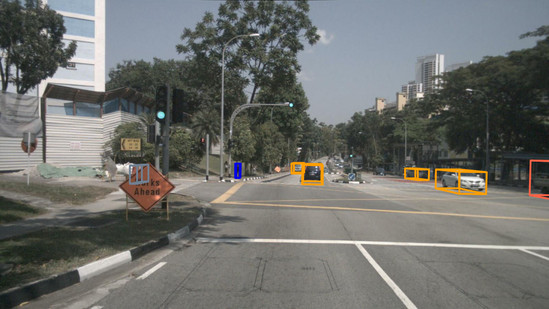}
\caption{Downstream detection quality on nuScenes. All detection classes are projected onto the front camera at a common score threshold.}
\label{fig:vad_tiny_detection_gallery}
\end{figure*}

\begin{figure*}[t]
\centering
\detectionHeaderThree{Ground Truth}{VAD-Tiny}{+ PAVER}
\detectionRowThree{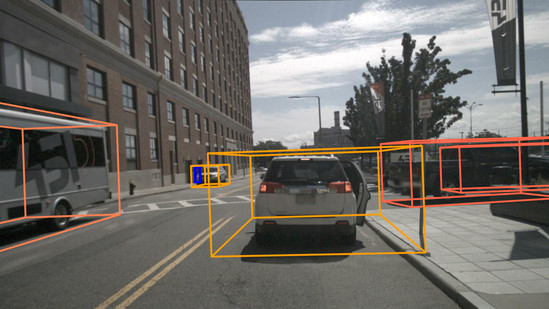}
\detectionRowThree{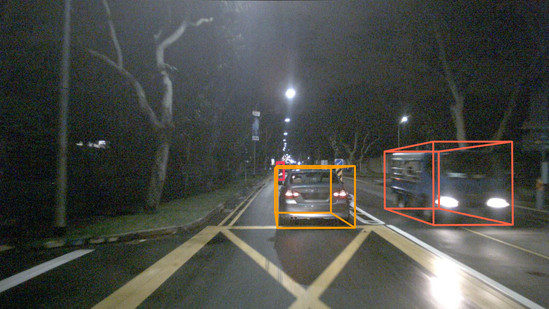}
\detectionRowThree{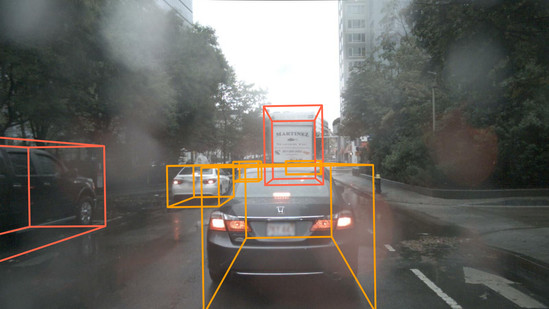}
\detectionRowThree{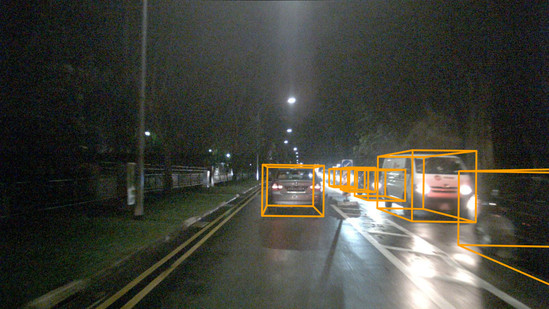}
\detectionRowThree{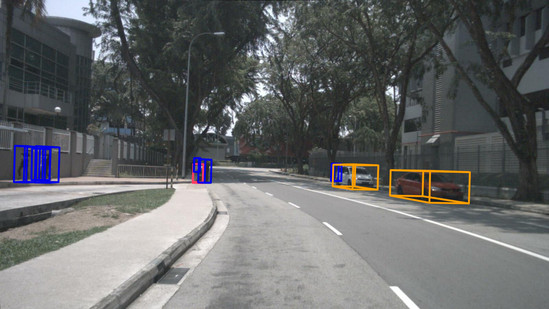}
\detectionRowThree{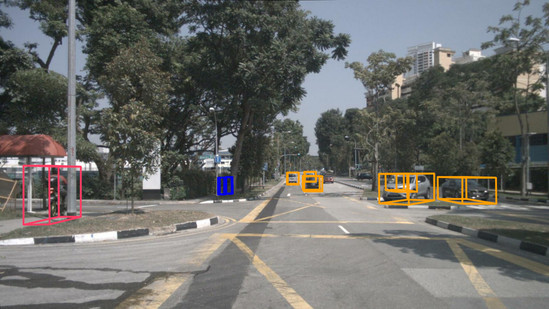}
\caption{Additional downstream detection examples on nuScenes.}
\label{fig:vad_tiny_detection_gallery_b}
\end{figure*}

\begin{figure*}[t]
\begin{fitSupplementFigure}
\centering
\qualitativeHeaderFour{Input Images}{Ground Truth}{VAD-Tiny}{+ PAVER}
\balancedRowFour{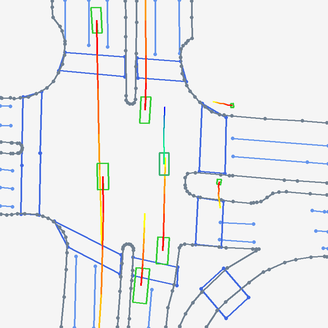}{vad_tiny}{paver_10k}
\balancedRowFour{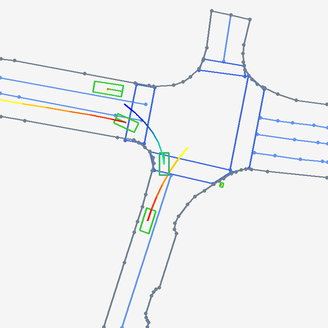}{vad_tiny}{paver_10k}
\balancedRowFour{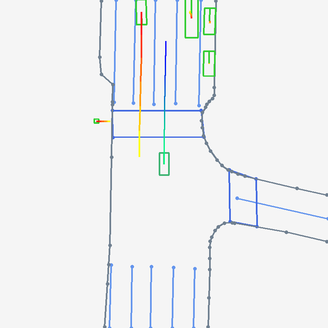}{vad_tiny}{paver_10k}
\balancedRowFour{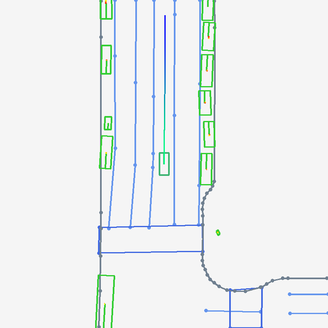}{vad_tiny}{paver_10k}
\balancedRowFour{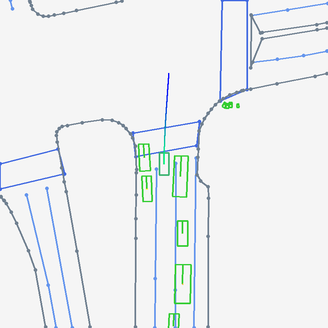}{vad_tiny}{paver_10k}
\balancedRowFour{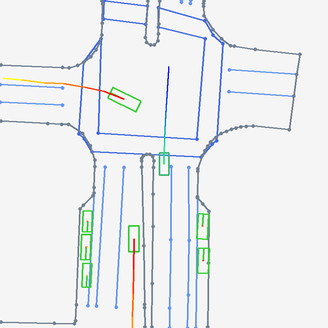}{vad_tiny}{paver_10k}
\caption{Downstream planning quality. After ground truth, the columns are VAD-Tiny trained from scratch and the same model initialized with PAVER pretraining.}
\label{fig:vad_tiny_downstream_gallery_a}
\end{fitSupplementFigure}
\end{figure*}

\begin{figure*}[t]
\centering
\qualitativeHeaderFour{Input Images}{Ground Truth}{VAD-Tiny}{+ PAVER}
\balancedRowFourCompact{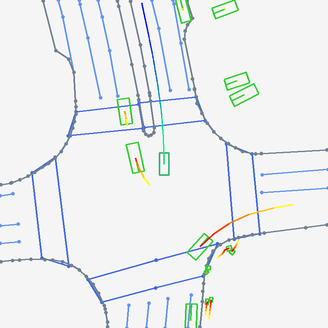}{vad_tiny}{paver_10k}
\balancedRowFourCompact{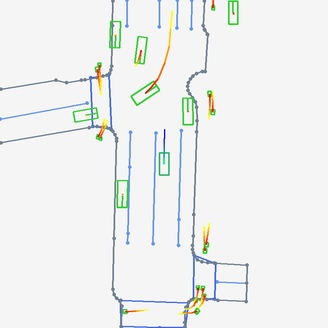}{vad_tiny}{paver_10k}
\balancedRowFourCompact{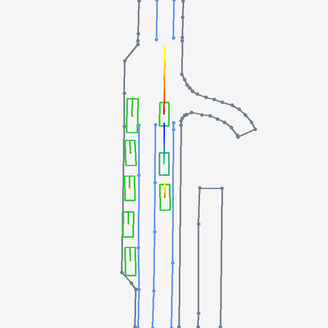}{vad_tiny}{paver_10k}
\balancedRowFourCompact{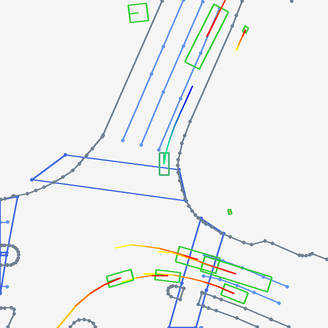}{vad_tiny}{paver_10k}
\balancedRowFourCompact{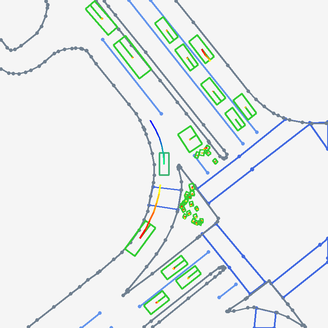}{vad_tiny}{paver_10k}
\balancedRowFourCompact{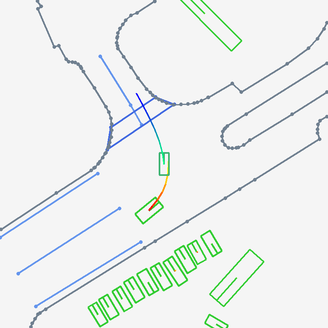}{vad_tiny}{paver_10k}
\caption{Joint map--planning and curved-driving cases. The first three rows improve both planning L2 and vector-map F1; the remaining rows show curved trajectories with lower planning L2.}
\label{fig:vad_tiny_downstream_gallery_b}
\end{figure*}

\begin{figure*}[t]
\centering
\qualitativeHeaderFive{Input Images}{Ground Truth}{VAD-Tiny}{w/o Action State}{w/ Action State}
\balancedRowFive{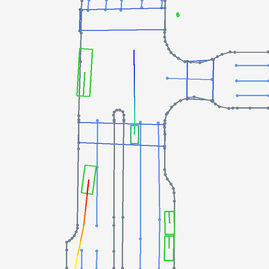}{vad_tiny}{w_o_action_state}{w_action_state}
\balancedRowFive{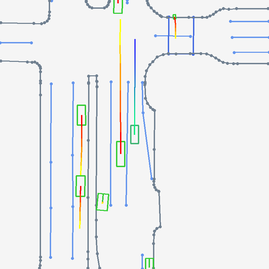}{vad_tiny}{w_o_action_state}{w_action_state}
\balancedRowFive{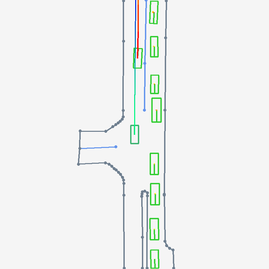}{vad_tiny}{w_o_action_state}{w_action_state}
\balancedRowFive{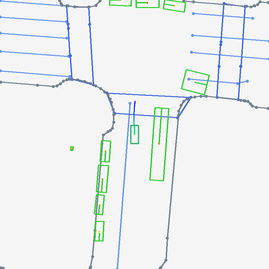}{vad_tiny}{w_o_action_state}{w_action_state}
\balancedRowFive{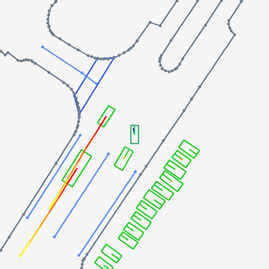}{vad_tiny}{w_o_action_state}{w_action_state}
\balancedRowFive{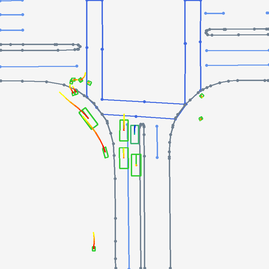}{vad_tiny}{w_o_action_state}{w_action_state}
\caption{Action-state ablation. After ground truth, the columns are VAD-Tiny trained from scratch, PAVER pretrained without action-state conditioning, and full PAVER.}
\label{fig:no_action_gallery_a}
\end{figure*}

\begin{figure*}[t]
\centering
\qualitativeHeaderFive{Input Images}{Ground Truth}{VAD-Tiny}{w/o Mask}{w/ Mask}
\balancedRowFive{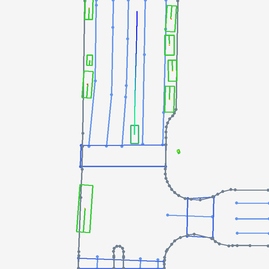}{vad_tiny}{w_o_mask}{w_mask}
\balancedRowFive{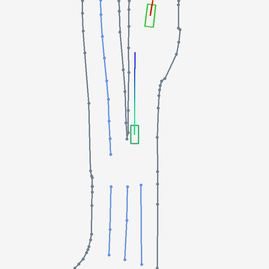}{vad_tiny}{w_o_mask}{w_mask}
\balancedRowFive{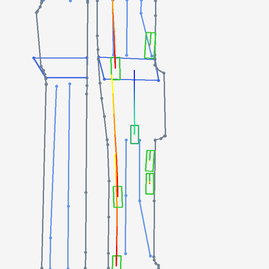}{vad_tiny}{w_o_mask}{w_mask}
\balancedRowFive{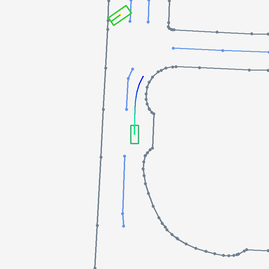}{vad_tiny}{w_o_mask}{w_mask}
\balancedRowFive{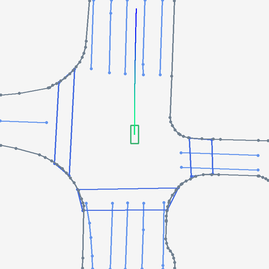}{vad_tiny}{w_o_mask}{w_mask}
\balancedRowFive{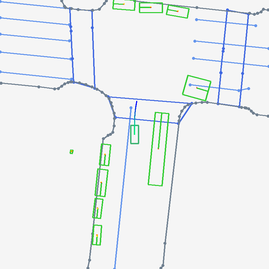}{vad_tiny}{w_o_mask}{w_mask}
\caption{Action-corridor masking ablation. After ground truth, the columns are VAD-Tiny trained from scratch, PAVER pretrained without action-corridor masking, and full PAVER.}
\label{fig:unmasked_gallery_a}
\end{figure*}

The component rows of Table~\ref{tab:all_fixed_checkpoint_metrics} carry the
aggregate metrics for this comparison, and
Figures~\ref{fig:no_action_gallery_a} and~\ref{fig:unmasked_gallery_a} show the
two ablations against the selected VAD-Tiny model of
Figures~\ref{fig:vad_tiny_downstream_gallery_a},
\ref{fig:vad_tiny_downstream_gallery_b}, \ref{fig:vad_tiny_detection_gallery}
and~\ref{fig:vad_tiny_detection_gallery_b}. Removing action conditioning keeps
the action-corridor mask, whereas removing the mask retains action
conditioning. Full PAVER has the lowest planning L2 and collision at every
horizon, as well as the highest exact NDS (0.3968). The partial variants retain
better individual motion or map metrics: removing action conditioning improves
FDE, MR, EPA, and map mAP, while removing the mask improves ADE, FDE, EPA, and
map mAP. Because each removal costs planning accuracy and safety while trading
away a different auxiliary metric, both components are load-bearing.

\subsection{Closed-Loop Route Outcomes on Bench2Drive}

\begin{figure*}[p]
\centering
\ifdefined\arxivpreprint
\begin{fitSupplementFigure}
\fi
\includegraphics[width=\textwidth]{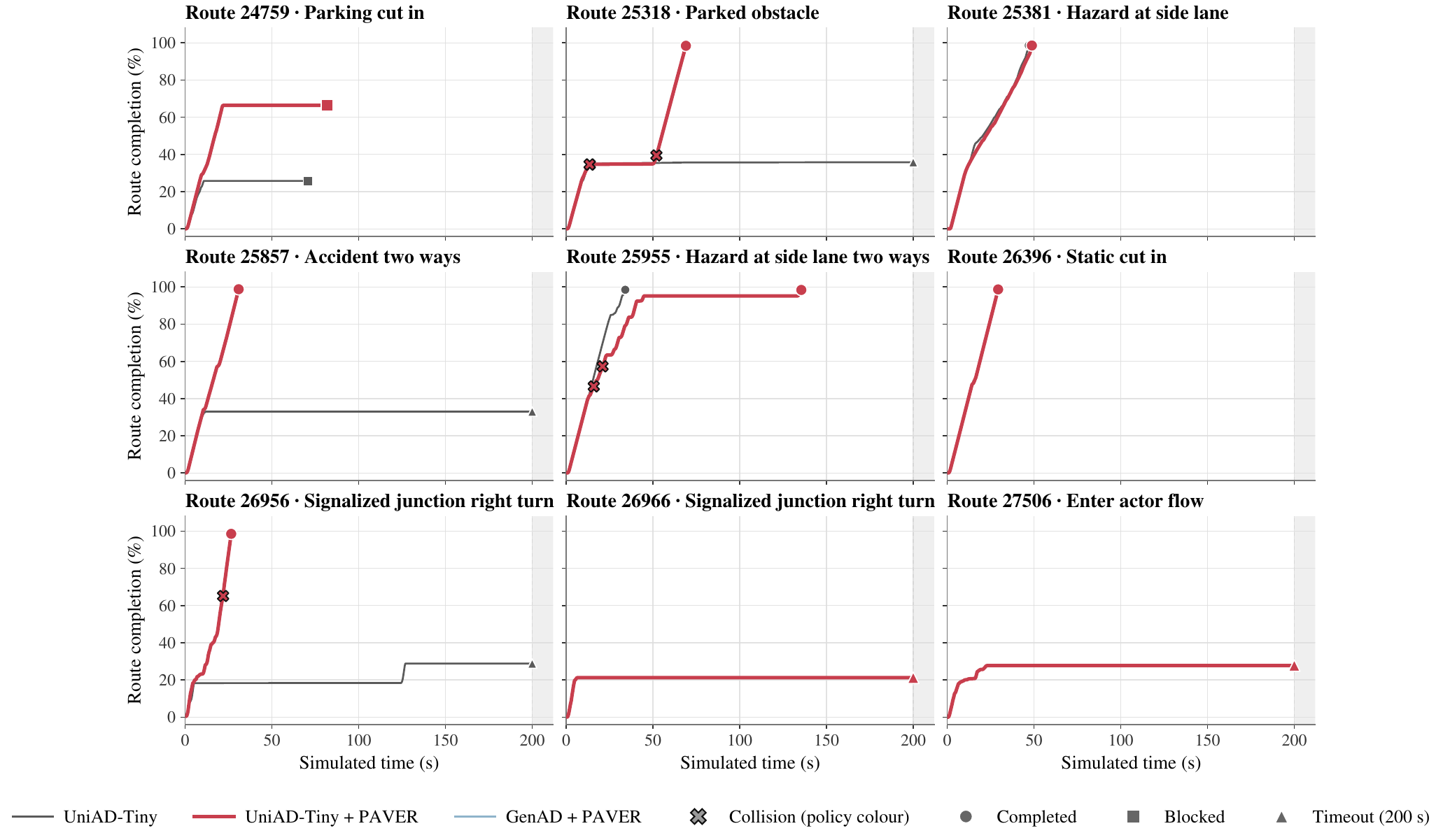}
\caption{Bench2Drive Town05 Long per-route timelines. Curves show recorded
route completion against simulated time; markers indicate collisions,
completion, or timeout. A missing completion trace is not a zero-completion
trajectory.}
\label{fig:b2d_route_timelines}
\ifdefined\arxivpreprint
\par\bigskip
\begin{minipage}{\textwidth}
\setlength{\multicolsep}{0pt}
\begin{multicols}{2}
\input{sections/supplement/03_summary}
\end{multicols}
\end{minipage}
\end{fitSupplementFigure}
\fi
\end{figure*}

Table~\ref{tab:b2d_route_audit} audits every Bench2Drive Town05 Long route
behind the closed-loop results of the main paper, with one repetition per route.
Timeout marks an exhausted 200\,s simulated-time budget. PAVER initialization
converts three UniAD-Tiny timeouts into completions.
Figure~\ref{fig:b2d_route_timelines} shows every route as a timeline. Three
UniAD-Tiny routes carry a frame-by-frame figure: the two whose outcome changes
and whose baseline frames were recorded, and one where both policies stop but at
very different points. Routes 26956, 26966 and 27506 recorded no baseline
frames, so they appear in the audit only. The GenAD policy ran the same nine
routes and two of them are shown frame by frame as well.

\begin{figure*}[tp]
\centering
\includegraphics[width=\textwidth]{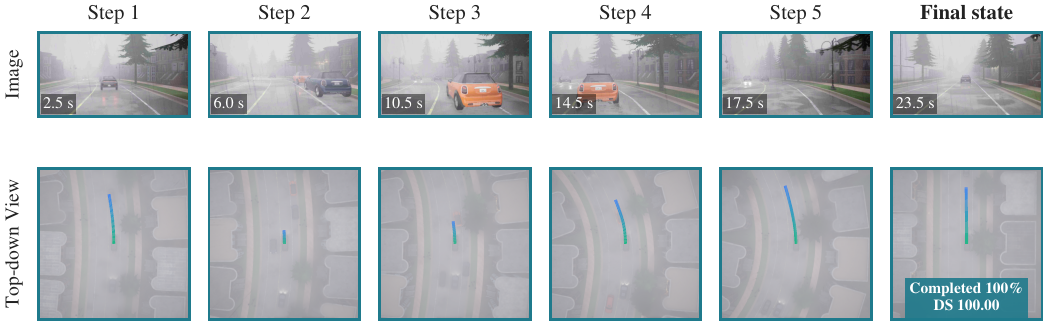}
\caption{Bench2Drive Town05 Long, route 24759: parking cut in, for GenAD
initialized with PAVER. The camera image is paired with the simulator's
top-down view, on which the planned path is drawn as a band running from its
near end to its far end. This run has no completion trace, so the columns are
spaced by elapsed time rather than by route completion, and the campaign
contains no scratch GenAD counterpart, so the figure reports single-policy
behaviour. The policy passes the cut-in vehicle and completes the route.}
\label{fig:b2d_filmstrip_genad}
\end{figure*}

\begin{figure*}[tp]
\centering
\includegraphics[width=\textwidth]{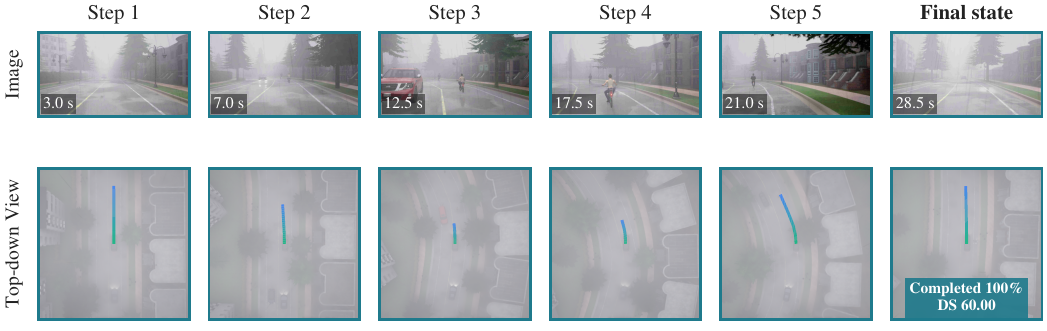}
\caption{Bench2Drive Town05 Long, route 25955: hazard at side lane two ways,
for GenAD initialized with PAVER. The policy
passes a stopped vehicle and a cyclist in fog and reaches the end of the route;
the driving score of 60.00 reflects one recorded collision on an otherwise
completed route (Table~\ref{tab:b2d_route_audit}).}
\label{fig:b2d_filmstrip_genad_25955}
\end{figure*}

The three filmstrips show a shared baseline failure mode: the policy reaches
the scenario obstacle, slows, and does not resume. With PAVER initialization,
UniAD-Tiny passes the parked obstacle at night (route 25318) and the accident
blocking the oncoming lane (route 25857), completing both routes. In the
parking cut-in scenario (route 24759), both policies remain blocked, but PAVER
increases route completion from 25.42\% to 66.91\%
(Table~\ref{tab:b2d_route_audit}).
The two completed routes require manoeuvres around static blockers rather
than lane following, consistent with the larger open-loop gains in the error
tail than at the median.
Figures~\ref{fig:b2d_filmstrip}, \ref{fig:b2d_filmstrip_25318}
and~\ref{fig:b2d_filmstrip_24759} show these three cases frame by frame.
Figures~\ref{fig:b2d_filmstrip_genad} and~\ref{fig:b2d_filmstrip_genad_25955}
show the GenAD policy on routes 24759 and 25955. Neither GenAD run has a
recorded completion trace, so both are stepped in elapsed time rather than
matched on route completion, and the campaign holds no scratch GenAD run to
pair them with; they document how a second architecture behaves after the same
pretraining rather than a controlled comparison.

\begin{table*}[p]
\centering
\small
\setlength{\tabcolsep}{6pt}
\begin{tabular}{llllccc}
\toprule
Route & Scenario & Policy & Outcome & RC $\uparrow$ & DS $\uparrow$ & Collisions $\downarrow$ \\
\midrule
24759 & Parking cut in & UniAD-Tiny & Blocked & 25.42 & 25.42 & 0 \\
\rowcolor{oursgray}
 &  & \textbf{+ PAVER} & Blocked & 66.91 & 52.43 & 0 \\
 &  & GenAD + PAVER & Completed$^{\dagger}$ & 100.00 & 100.00 & 0 \\
\midrule
25318 & Parked obstacle & UniAD-Tiny & Timeout & 35.69 & 21.41 & 1 \\
\rowcolor{oursgray}
 &  & \textbf{+ PAVER} & Completed & 100.00 & 36.00 & 2 \\
 &  & GenAD + PAVER & Timeout$^{\dagger}$ & 31.86 & 31.86 & 0 \\
\midrule
25381 & Hazard at side lane & UniAD-Tiny & Completed & 100.00 & 100.00 & 0 \\
\rowcolor{oursgray}
 &  & \textbf{+ PAVER} & Completed & 100.00 & 100.00 & 0 \\
 &  & GenAD + PAVER & Completed$^{\dagger}$ & 100.00 & 51.09 & 1 \\
\midrule
25857 & Accident two ways & UniAD-Tiny & Timeout & 32.30 & 32.30 & 0 \\
\rowcolor{oursgray}
 &  & \textbf{+ PAVER} & Completed & 100.00 & 100.00 & 0 \\
 &  & GenAD + PAVER & Timeout$^{\dagger}$ & 33.83 & 33.83 & 0 \\
\midrule
25955 & Hazard at side lane two ways & UniAD-Tiny & Completed & 100.00 & 100.00 & 0 \\
\rowcolor{oursgray}
 &  & \textbf{+ PAVER} & Completed & 100.00 & 36.00 & 2 \\
 &  & GenAD + PAVER & Completed$^{\dagger}$ & 100.00 & 60.00 & 1 \\
\midrule
26396 & Static cut in & UniAD-Tiny & Completed & 100.00 & 100.00 & 0 \\
\rowcolor{oursgray}
 &  & \textbf{+ PAVER} & Completed & 100.00 & 100.00 & 0 \\
 &  & GenAD + PAVER & Completed$^{\dagger}$ & 100.00 & 100.00 & 0 \\
\midrule
26956 & Signalized junction right turn & UniAD-Tiny & Timeout & 26.72 & 26.72 & 0 \\
\rowcolor{oursgray}
 &  & \textbf{+ PAVER} & Completed & 100.00 & 60.00 & 1 \\
 &  & GenAD + PAVER & Timeout$^{\dagger}$ & 22.64 & 22.64 & 0 \\
\midrule
26966 & Signalized junction right turn & UniAD-Tiny & Timeout & 28.48 & 28.48 & 0 \\
\rowcolor{oursgray}
 &  & \textbf{+ PAVER} & Timeout & 18.15 & 18.15 & 0 \\
 &  & GenAD + PAVER & Timeout$^{\dagger}$ & 20.73 & 20.73 & 0 \\
\midrule
27506 & Enter actor flow & UniAD-Tiny & Completed & 100.00 & 1.74 & 7 \\
\rowcolor{oursgray}
 &  & \textbf{+ PAVER} & Timeout & 26.52 & 26.52 & 0 \\
 &  & GenAD + PAVER & Timeout$^{\dagger}$ & 21.50 & 21.50 & 0 \\
\bottomrule
\end{tabular}
\caption{Bench2Drive Town05 Long, per-route outcome audit. One repetition per
route; RC is route completion, DS the composed driving score. Timeout marks
an exhausted 200\,s simulated-time budget. $^{\dagger}$ scored without a
completion trace, so it cannot be matched on route completion.}
\label{tab:b2d_route_audit}
\end{table*}

\newcommand{\bTwoPolicyFilmstrip}[1]{%
  \includegraphics[width=\textwidth,trim=0 574pt 0 0,clip]{#1}%
  \par\nointerlineskip
  \includegraphics[width=\textwidth,trim=0 0 0 24pt,clip]{#1}%
}

\begin{figure*}[tp]
\centering
\bTwoPolicyFilmstrip{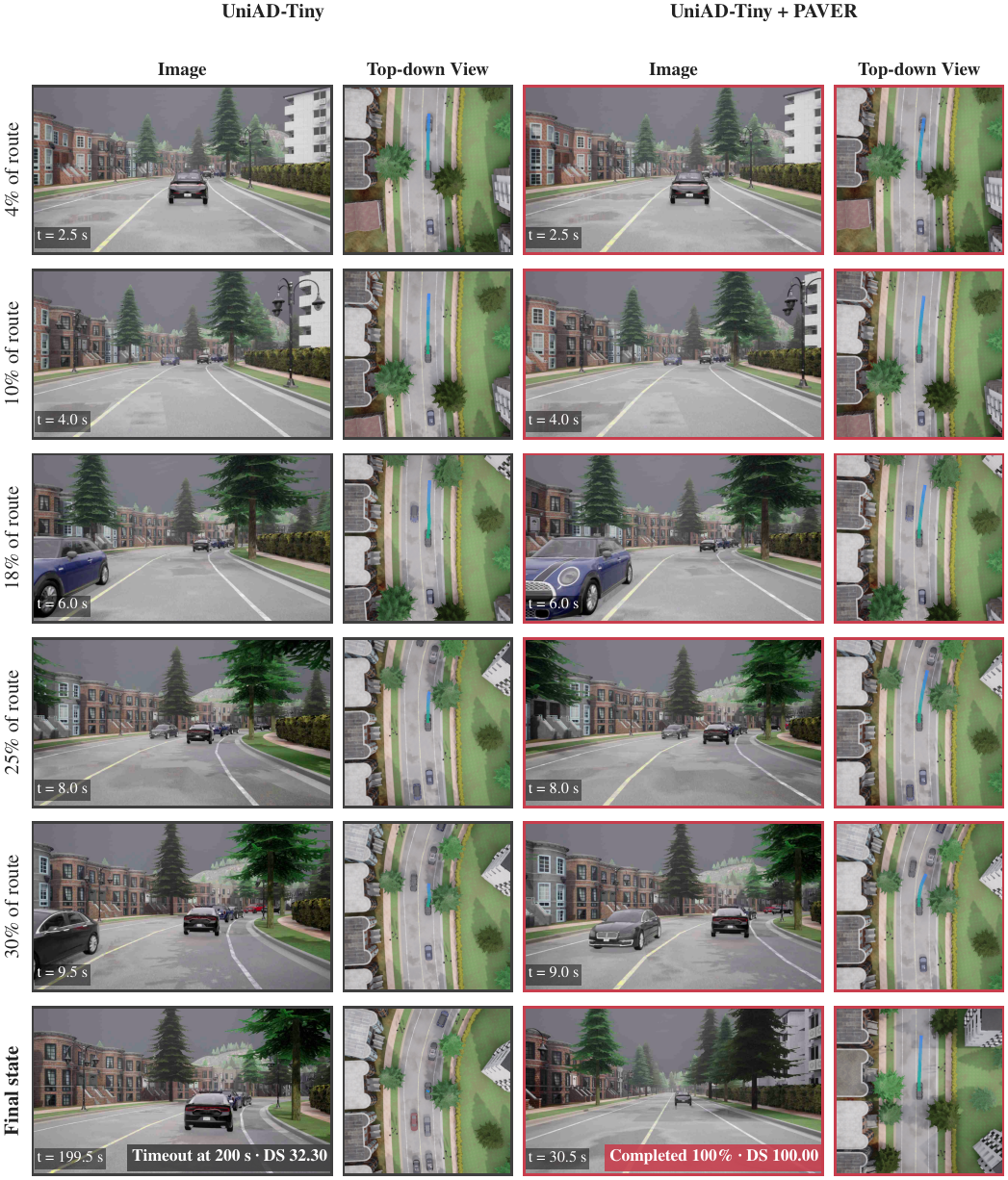}
\caption{Bench2Drive Town05 Long, route 25857: accident two ways. Rows are
matched on route completion, so both policies are shown at the same physical
point; each camera image is paired with the simulator's top-down view.
UniAD-Tiny exhausts the 200\,s budget at 32.30 route completion, while the same
architecture initialized with PAVER completes the route
(Table~\ref{tab:b2d_route_audit}).}
\label{fig:b2d_filmstrip}
\end{figure*}

\begin{figure*}[tp]
\centering
\bTwoPolicyFilmstrip{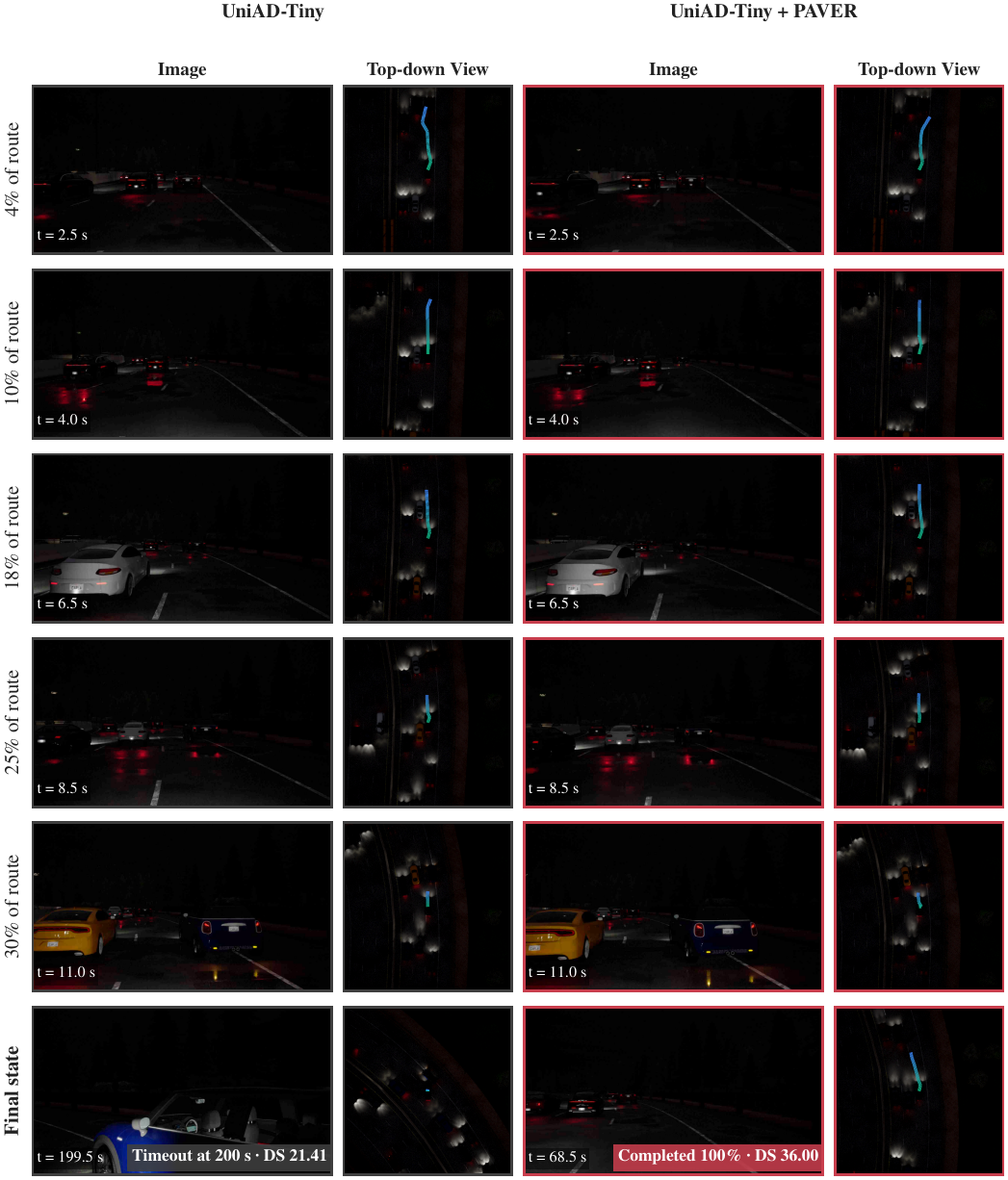}
\caption{Bench2Drive Town05 Long, route 25318: parked obstacle at night.
UniAD-Tiny exhausts the time budget behind the obstacle; with PAVER
initialization the same architecture passes it and reaches the end of the
route.}
\label{fig:b2d_filmstrip_25318}
\end{figure*}

\begin{figure*}[tp]
\centering
\bTwoPolicyFilmstrip{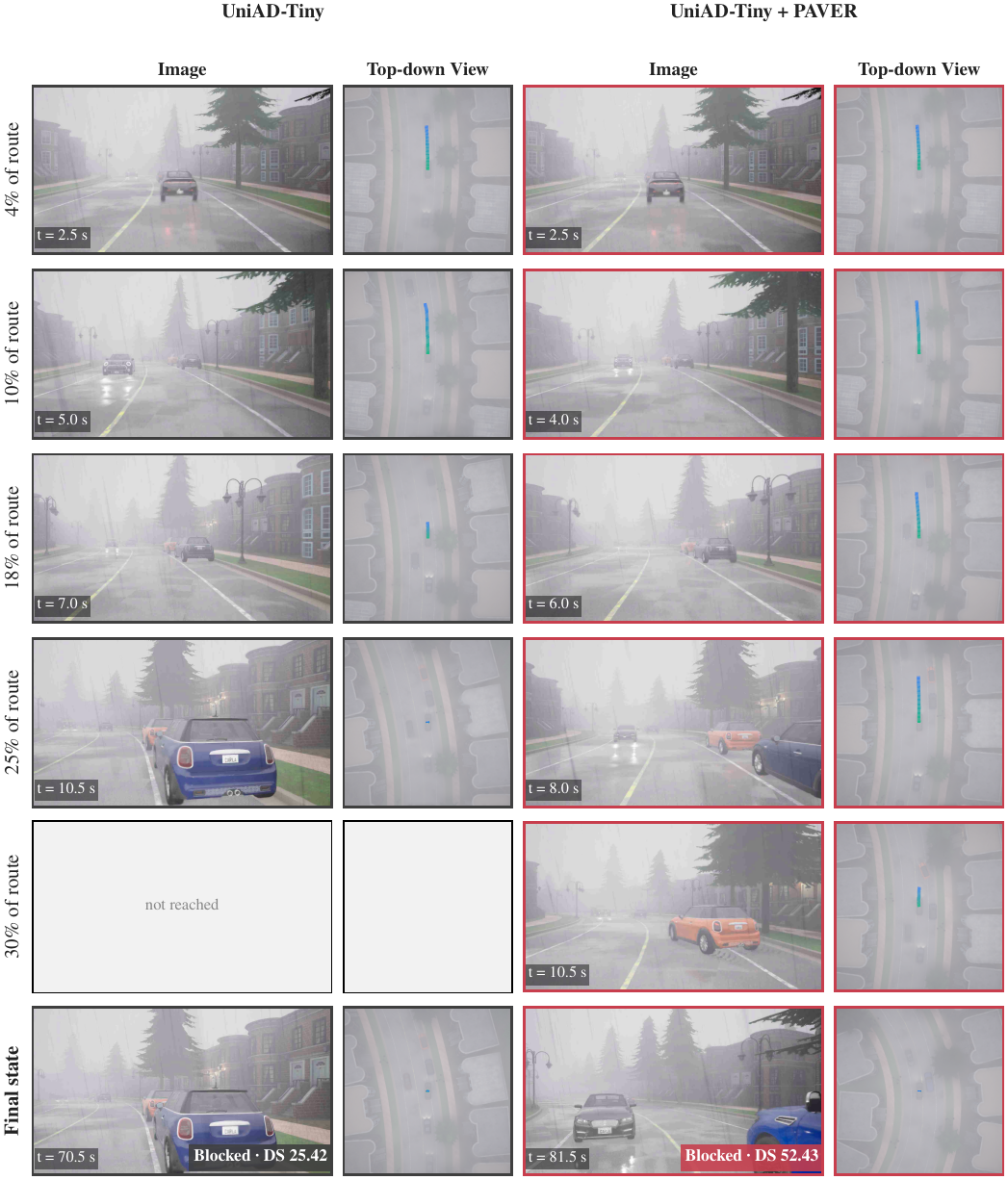}
\caption{Bench2Drive Town05 Long, route 24759: parking cut in. Both policies
are blocked by the cut-in vehicle, but the pretrained model travels more than
twice as far before it stops (Table~\ref{tab:b2d_route_audit}).}
\label{fig:b2d_filmstrip_24759}
\end{figure*}

\ifdefined\arxivpreprint\else
\section{Mechanistic Summary}
\label{sec:supp_mechanistic_summary}

The analyses support a 4-part explanation for the observed transfer. First,
PAVER compresses privileged LiDAR evidence into two ratios instead of reconstructing
a high-dimensional scene. Second, rule-based actions determine where those
ratios are queried, aligning the supervision with spatial evidence later used by
planning.
\ifdefined\arxivpreprint
Third, supervision is spatially sparse, but frozen probes detect representation
changes beyond the sampled corridor.
\else
Third, the target coordinates are sparse but the optimization path is
not: gradients update the shared BEV encoder, and frozen probes detect changes
beyond the sampled corridor.
\fi
Fourth, the pretraining-only auxiliary head is removed, while
the transferred encoder remains adaptable; early downstream weight drift and
multi-task learning curves show refinement rather than dependence on a retained
pretraining decoder.

This combination explains why parameter count and target difficulty need not
track transfer quality. A large decoder is useful when pretraining must recover
RGB, depth, dense occupancy, or future scene state
\citep{yang2024unipad,zou2025mim4d,min2024driveworld,yang2024vidar}. PAVER asks
a smaller question whose coordinates carry the planning prior, so the 10K
parameters only map an already shared BEV feature and action state to two outputs.
The useful capacity is the downstream BEV encoder being organized, not the
pretraining-only auxiliary head being enlarged. The same account also predicts the observed
limit: because the targets summarize current geometric support, transfer can
weaken when occluded intent or future agent dynamics dominate.

\section{Evidence Scope and Limitations}
\label{sec:supp_evidence_limits}

The comprehensive analyses use preserved checkpoints and matched nuScenes
samples.
\ifdefined\arxivpreprint\else
Learning-curve advantage and threshold-crossing summaries use downstream
epochs 1--30.
\fi
The native trajectories in
Figure~\ref{fig:supp_vad_tiny_all_experiment_trajectories} extend to the
available checkpoints, including epoch 40 for pretraining variants and
epoch 60 for scratch. Each configuration is trained once; scene-cluster
bootstrap intervals measure variation across sampled scenes, not across
independent training runs.
\ifdefined\arxivpreprint\else
The gradient audit instead uses eight fixed training frames.
\fi
Correlated checkpoints are not treated as independent trials.
\ifdefined\arxivpreprint\else
The route, kinematic, environment, complexity, and interaction strata are
observational; they identify heterogeneity rather than causal robustness effects.
\fi

The frozen-predictor controls establish dependence on BEV content, spatial
arrangement, and action state, but do not isolate every architectural mechanism.
The closed-loop routes are one repetition each on nine routes, which indicates
transfer rather than a ranking.
\ifdefined\arxivpreprint\else
Camera-dropout tests cover sustained missing views represented by mean-valued
images; they do not cover blur, calibration error, exposure shift,
partial occlusion, or intermittent packet loss.
\fi
These results remain diagnostic evidence and should not be extrapolated
to general driving safety.

\fi

\clearpage

\clearpage

\ifdefined\arxivpreprint\else
\begin{figure*}[t]
\centering
\detectionProjectionSample{figures/qualitative/pretraining/matched/detection/rank_02_7c07e754d1e243ce82fcc01ef1ee292f__gt_3cam.png}{figures/qualitative/pretraining/matched/detection/rank_02_7c07e754d1e243ce82fcc01ef1ee292f__detection_only_3cam.png}{figures/qualitative/pretraining/matched/detection/rank_02_7c07e754d1e243ce82fcc01ef1ee292f__detection_map_3cam.png}{figures/qualitative/pretraining/matched/detection/rank_02_7c07e754d1e243ce82fcc01ef1ee292f__detection_map_occupancy_3cam.png}
\caption{Detection pretraining quality, sample 2.}
\label{fig:pretraining_detection_gallery_b}
\end{figure*}

\begin{figure*}[t]
\centering
\detectionProjectionSample{figures/qualitative/pretraining/matched/detection/rank_03_20622852ae754f25bb55b8fa466b5455__gt_3cam.png}{figures/qualitative/pretraining/matched/detection/rank_03_20622852ae754f25bb55b8fa466b5455__detection_only_3cam.png}{figures/qualitative/pretraining/matched/detection/rank_03_20622852ae754f25bb55b8fa466b5455__detection_map_3cam.png}{figures/qualitative/pretraining/matched/detection/rank_03_20622852ae754f25bb55b8fa466b5455__detection_map_occupancy_3cam.png}
\caption{Detection pretraining quality, sample 3.}
\label{fig:pretraining_detection_gallery_c}
\end{figure*}
\fi

\clearpage

\ifdefined\arxivpreprint
\begin{figure*}[t]
\begin{fitSupplementFigure}
\centering
\occHeaderCompact{Input Images}{Ground Truth}{Occ. Only}{Det. + Map + Occ.}
\occRowCompact{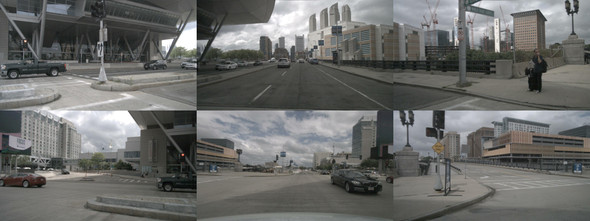}{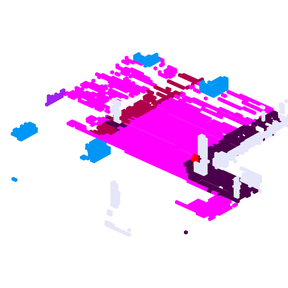}{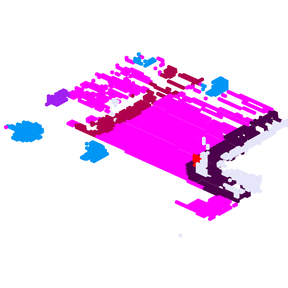}{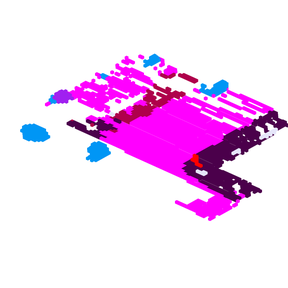}
\occRowCompact{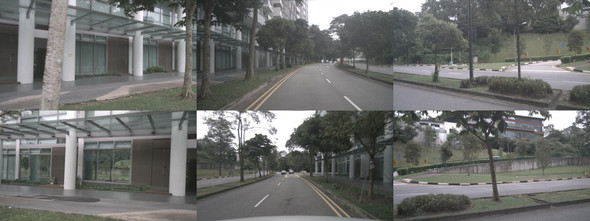}{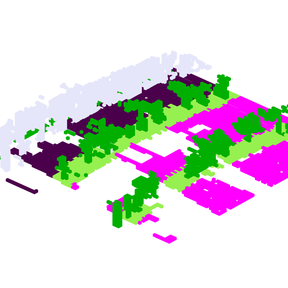}{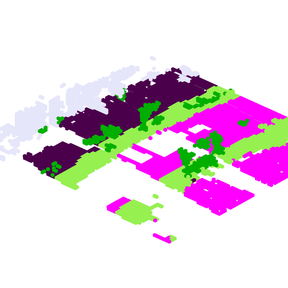}{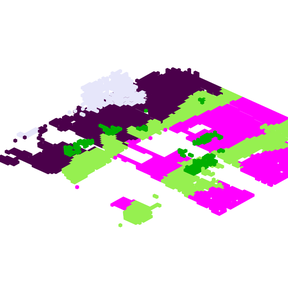}
\occRowCompact{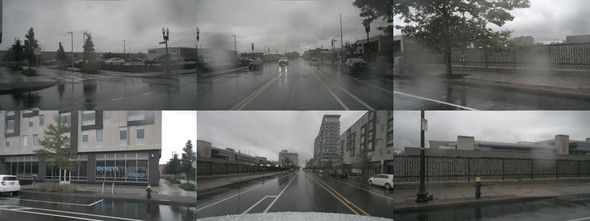}{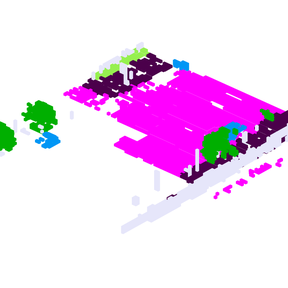}{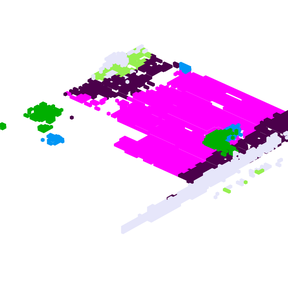}{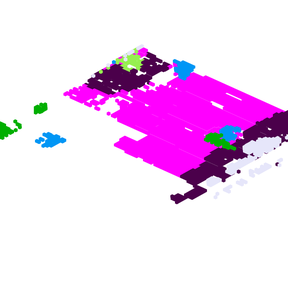}
\occRowCompact{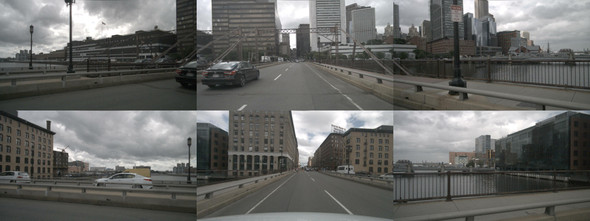}{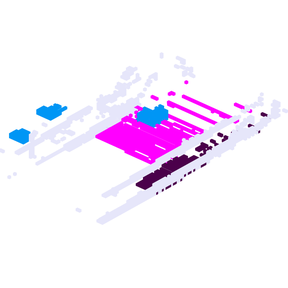}{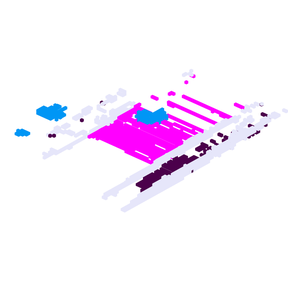}{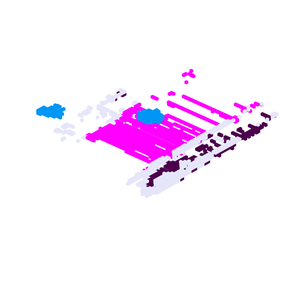}
\occRowCompact{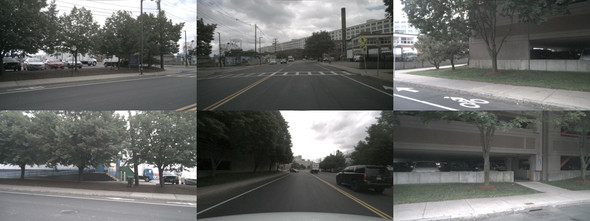}{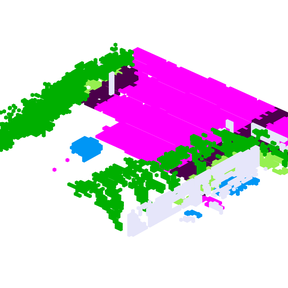}{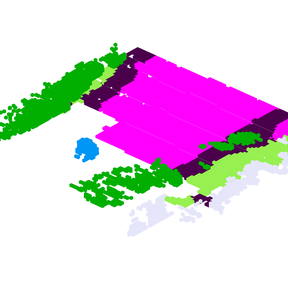}{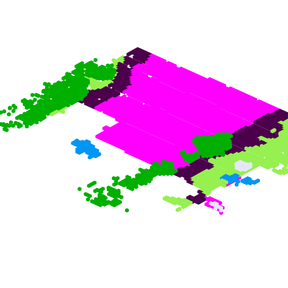}
\occRowCompact{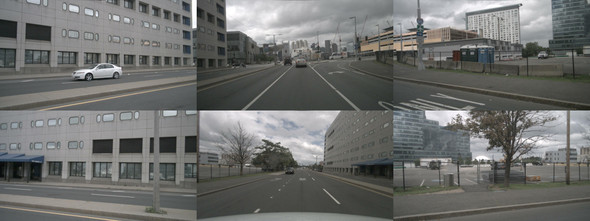}{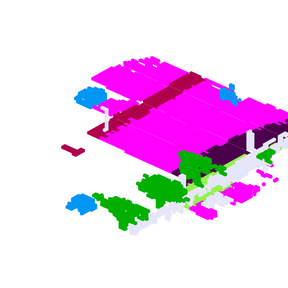}{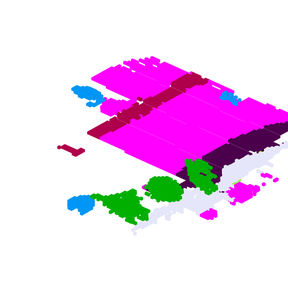}{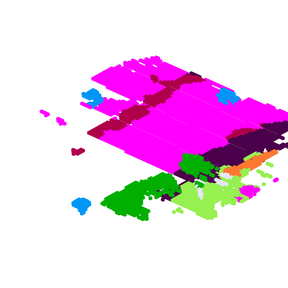}
\occRowCompact{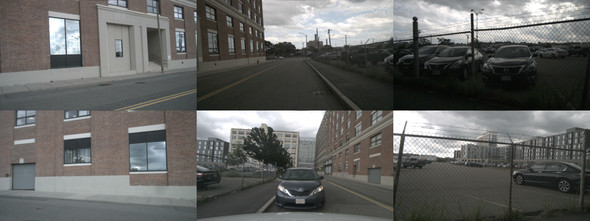}{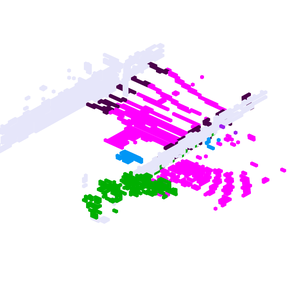}{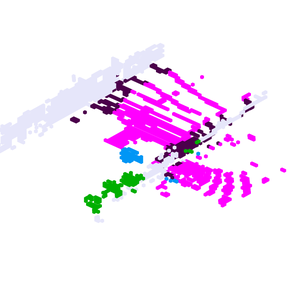}{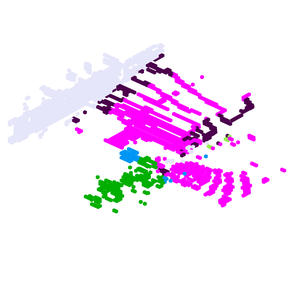}
\caption{Occupancy pretraining quality after the shared 20-epoch budget. Columns compare occupancy-only and joint detection--map--occupancy pretraining against ground truth.}
\label{fig:pretraining_occupancy_gallery_a}
\label{fig:pretraining_occupancy_gallery_b}
\end{fitSupplementFigure}
\end{figure*}
\fi

\ifdefined\arxivpreprint\else
\begin{figure*}[t]
\centering
\qualitativeHeaderFour{Input Images}{Ground Truth}{Occ. Only}{Det. + Map + Occ.}
\qualitativeRowFour{figures/qualitative/pretraining/matched/occupancy/rank_01_ded088a5ef664ca687fd0e8c1e4bda1d__input_cameras.png}{figures/qualitative/pretraining/matched/occupancy/rank_01_ded088a5ef664ca687fd0e8c1e4bda1d__gt.png}{figures/qualitative/pretraining/matched/occupancy/rank_01_ded088a5ef664ca687fd0e8c1e4bda1d__occupancy_only.png}{figures/qualitative/pretraining/matched/occupancy/rank_01_ded088a5ef664ca687fd0e8c1e4bda1d__detection_map_occupancy.png}
\qualitativeRowFour{figures/qualitative/pretraining/matched/occupancy/rank_02_f7a001c474d844618d4c6fb15dc16b8c__input_cameras.png}{figures/qualitative/pretraining/matched/occupancy/rank_02_f7a001c474d844618d4c6fb15dc16b8c__gt.png}{figures/qualitative/pretraining/matched/occupancy/rank_02_f7a001c474d844618d4c6fb15dc16b8c__occupancy_only.png}{figures/qualitative/pretraining/matched/occupancy/rank_02_f7a001c474d844618d4c6fb15dc16b8c__detection_map_occupancy.png}
\qualitativeRowFour{figures/qualitative/pretraining/matched/occupancy/rank_03_9c6aa2cb631841abb2cd4bebea86d03f__input_cameras.jpg}{figures/qualitative/pretraining/matched/occupancy/rank_03_9c6aa2cb631841abb2cd4bebea86d03f__gt.png}{figures/qualitative/pretraining/matched/occupancy/rank_03_9c6aa2cb631841abb2cd4bebea86d03f__occupancy_only.png}{figures/qualitative/pretraining/matched/occupancy/rank_03_9c6aa2cb631841abb2cd4bebea86d03f__detection_map_occupancy.png}
\qualitativeRowFour{figures/qualitative/pretraining/matched/occupancy/rank_04_6e60a07fb4234c1d8b669f62bd49da5b__input_cameras.jpg}{figures/qualitative/pretraining/matched/occupancy/rank_04_6e60a07fb4234c1d8b669f62bd49da5b__gt.png}{figures/qualitative/pretraining/matched/occupancy/rank_04_6e60a07fb4234c1d8b669f62bd49da5b__occupancy_only.png}{figures/qualitative/pretraining/matched/occupancy/rank_04_6e60a07fb4234c1d8b669f62bd49da5b__detection_map_occupancy.png}
\qualitativeRowFour{figures/qualitative/pretraining/matched/occupancy/rank_05_74288be6042148a985ab936b69e90eec__input_cameras.jpg}{figures/qualitative/pretraining/matched/occupancy/rank_05_74288be6042148a985ab936b69e90eec__gt.png}{figures/qualitative/pretraining/matched/occupancy/rank_05_74288be6042148a985ab936b69e90eec__occupancy_only.png}{figures/qualitative/pretraining/matched/occupancy/rank_05_74288be6042148a985ab936b69e90eec__detection_map_occupancy.png}
\caption{Occupancy pretraining quality, set 1, after the shared 20-epoch budget. Columns compare occupancy-only and joint detection--map--occupancy pretraining against ground truth.}
\label{fig:pretraining_occupancy_gallery_a}
\end{figure*}
\fi

\ifdefined\arxivpreprint\else
\begin{figure*}[t]
\centering
\qualitativeHeaderFour{Input Images}{Ground Truth}{Occ. Only}{Det. + Map + Occ.}
\qualitativeRowFour{figures/qualitative/pretraining/matched/occupancy/rank_06_21971c1ec4644d608f8c08bbfd4059f7__input_cameras.jpg}{figures/qualitative/pretraining/matched/occupancy/rank_06_21971c1ec4644d608f8c08bbfd4059f7__gt.png}{figures/qualitative/pretraining/matched/occupancy/rank_06_21971c1ec4644d608f8c08bbfd4059f7__occupancy_only.png}{figures/qualitative/pretraining/matched/occupancy/rank_06_21971c1ec4644d608f8c08bbfd4059f7__detection_map_occupancy.png}
\qualitativeRowFour{figures/qualitative/pretraining/matched/occupancy/rank_07_3511eec3e52940aaaa964b145ed5dbc7__input_cameras.jpg}{figures/qualitative/pretraining/matched/occupancy/rank_07_3511eec3e52940aaaa964b145ed5dbc7__gt.png}{figures/qualitative/pretraining/matched/occupancy/rank_07_3511eec3e52940aaaa964b145ed5dbc7__occupancy_only.png}{figures/qualitative/pretraining/matched/occupancy/rank_07_3511eec3e52940aaaa964b145ed5dbc7__detection_map_occupancy.png}
\qualitativeRowFour{figures/qualitative/pretraining/matched/occupancy/rank_08_f41862469f33424eb660258e9392c4f4__input_cameras.png}{figures/qualitative/pretraining/matched/occupancy/rank_08_f41862469f33424eb660258e9392c4f4__gt.png}{figures/qualitative/pretraining/matched/occupancy/rank_08_f41862469f33424eb660258e9392c4f4__occupancy_only.png}{figures/qualitative/pretraining/matched/occupancy/rank_08_f41862469f33424eb660258e9392c4f4__detection_map_occupancy.png}
\qualitativeRowFour{figures/qualitative/pretraining/matched/occupancy/rank_09_c8b6a861ef0445cd8bef18eb2b7f9e89__input_cameras.jpg}{figures/qualitative/pretraining/matched/occupancy/rank_09_c8b6a861ef0445cd8bef18eb2b7f9e89__gt.png}{figures/qualitative/pretraining/matched/occupancy/rank_09_c8b6a861ef0445cd8bef18eb2b7f9e89__occupancy_only.png}{figures/qualitative/pretraining/matched/occupancy/rank_09_c8b6a861ef0445cd8bef18eb2b7f9e89__detection_map_occupancy.png}
\qualitativeRowFour{figures/qualitative/pretraining/matched/occupancy/rank_10_69b3d34740f049ada3f88e9866aa46d4__input_cameras.jpg}{figures/qualitative/pretraining/matched/occupancy/rank_10_69b3d34740f049ada3f88e9866aa46d4__gt.png}{figures/qualitative/pretraining/matched/occupancy/rank_10_69b3d34740f049ada3f88e9866aa46d4__occupancy_only.png}{figures/qualitative/pretraining/matched/occupancy/rank_10_69b3d34740f049ada3f88e9866aa46d4__detection_map_occupancy.png}
\caption{Occupancy pretraining quality, set 2. Columns follow Figure~\ref{fig:pretraining_occupancy_gallery_a}.}
\label{fig:pretraining_occupancy_gallery_b}
\end{figure*}
\fi

\clearpage

\clearpage

\clearpage

\clearpage

\clearpage

\clearpage

\clearpage

\clearpage

\clearpage

\clearpage

\clearpage

\clearpage

\begin{figure*}[p]
\centering
\gradCamCameraHeader
\gradCamInputRowThree{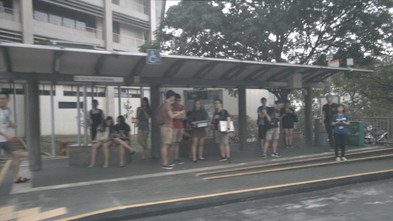}{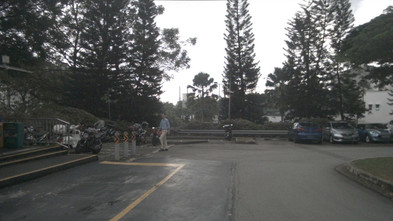}{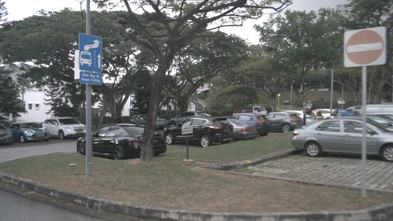}
\gradCamAttributionRowThree{VAD-Tiny}{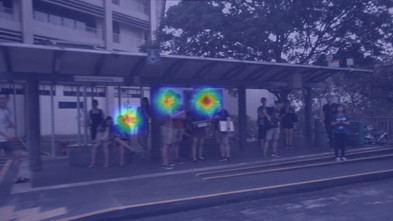}{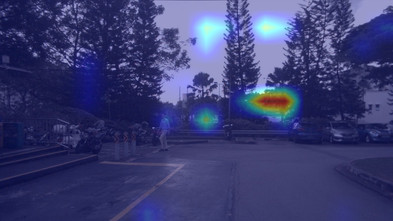}{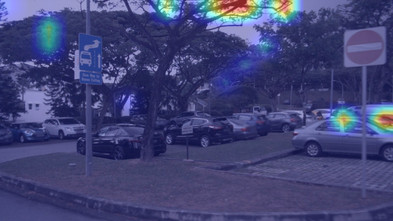}
\gradCamAttributionRowThree{+ PAVER}{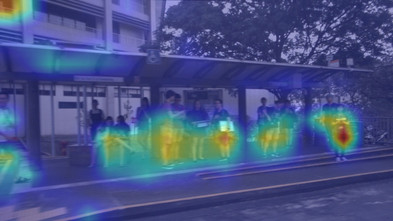}{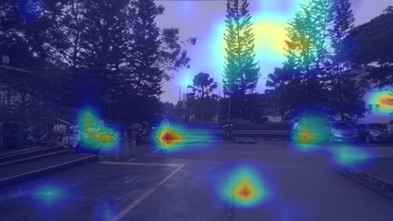}{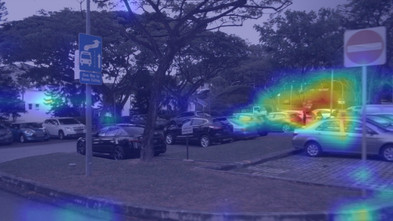}
\caption{Grad-CAM results on the nuScenes validation set.}
\label{fig:supp_vad_tiny_gradcam_three_camera_003}
\vspace{8pt}
\gradCamCameraHeader
\gradCamInputRowThree{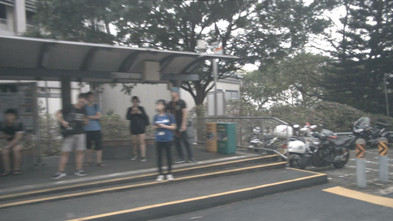}{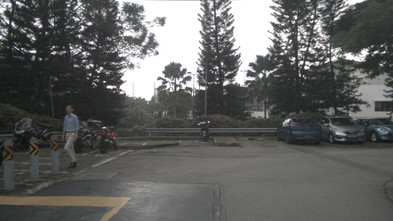}{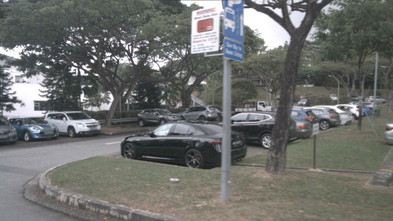}
\gradCamAttributionRowThree{VAD-Tiny}{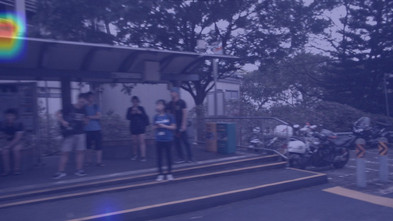}{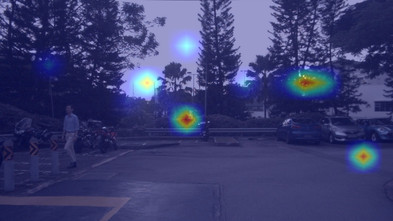}{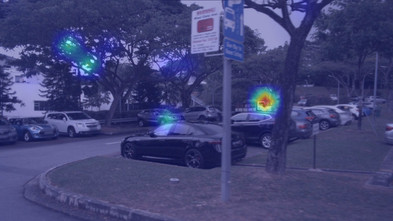}
\gradCamAttributionRowThree{+ PAVER}{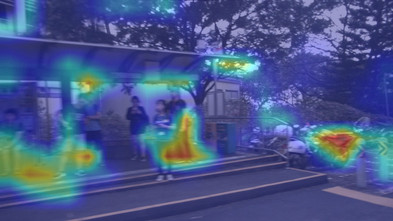}{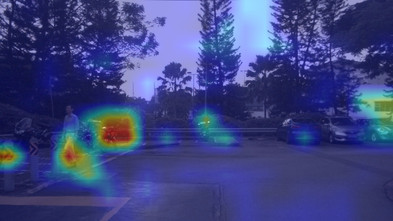}{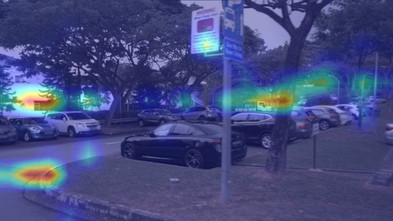}
\caption{Grad-CAM results on the nuScenes validation set.}
\label{fig:supp_vad_tiny_gradcam_three_camera_006}
\end{figure*}
\clearpage

\ifdefined\arxivpreprint\else
\begin{figure*}[p]
\centering
\gradCamCameraHeader
\gradCamInputRowThree{figures/analysis/gradcam_vad_tiny/panels/rank_014__cecb41d3a0b243c3b0d37a269ce855ad__front_left__input.png}{figures/analysis/gradcam_vad_tiny/panels/rank_014__cecb41d3a0b243c3b0d37a269ce855ad__front__input.png}{figures/analysis/gradcam_vad_tiny/panels/rank_014__cecb41d3a0b243c3b0d37a269ce855ad__front_right__input.png}
\gradCamAttributionRowThree{VAD-Tiny}{figures/analysis/gradcam_vad_tiny/panels/rank_014__cecb41d3a0b243c3b0d37a269ce855ad__front_left__vad_tiny_bev.png}{figures/analysis/gradcam_vad_tiny/panels/rank_014__cecb41d3a0b243c3b0d37a269ce855ad__front__vad_tiny_bev.png}{figures/analysis/gradcam_vad_tiny/panels/rank_014__cecb41d3a0b243c3b0d37a269ce855ad__front_right__vad_tiny_bev.png}
\gradCamAttributionRowThree{+ PAVER}{figures/analysis/gradcam_vad_tiny/panels/rank_014__cecb41d3a0b243c3b0d37a269ce855ad__front_left__paver_bev.png}{figures/analysis/gradcam_vad_tiny/panels/rank_014__cecb41d3a0b243c3b0d37a269ce855ad__front__paver_bev.png}{figures/analysis/gradcam_vad_tiny/panels/rank_014__cecb41d3a0b243c3b0d37a269ce855ad__front_right__paver_bev.png}
\caption{Grad-CAM results on the nuScenes validation set.}
\label{fig:supp_vad_tiny_gradcam_three_camera_014}
\vspace{8pt}
\gradCamCameraHeader
\gradCamInputRowThree{figures/analysis/gradcam_vad_tiny/panels/rank_039__609d5177362340458a3bfd4949cd1e64__front_left__input.png}{figures/analysis/gradcam_vad_tiny/panels/rank_039__609d5177362340458a3bfd4949cd1e64__front__input.png}{figures/analysis/gradcam_vad_tiny/panels/rank_039__609d5177362340458a3bfd4949cd1e64__front_right__input.png}
\gradCamAttributionRowThree{VAD-Tiny}{figures/analysis/gradcam_vad_tiny/panels/rank_039__609d5177362340458a3bfd4949cd1e64__front_left__vad_tiny_bev.png}{figures/analysis/gradcam_vad_tiny/panels/rank_039__609d5177362340458a3bfd4949cd1e64__front__vad_tiny_bev.png}{figures/analysis/gradcam_vad_tiny/panels/rank_039__609d5177362340458a3bfd4949cd1e64__front_right__vad_tiny_bev.png}
\gradCamAttributionRowThree{+ PAVER}{figures/analysis/gradcam_vad_tiny/panels/rank_039__609d5177362340458a3bfd4949cd1e64__front_left__paver_bev.png}{figures/analysis/gradcam_vad_tiny/panels/rank_039__609d5177362340458a3bfd4949cd1e64__front__paver_bev.png}{figures/analysis/gradcam_vad_tiny/panels/rank_039__609d5177362340458a3bfd4949cd1e64__front_right__paver_bev.png}
\caption{Grad-CAM results on the nuScenes validation set.}
\label{fig:supp_vad_tiny_gradcam_three_camera_039}
\end{figure*}
\fi
\clearpage

\begin{figure*}[p]
\centering
\gradCamCameraHeader
\gradCamInputRowThree{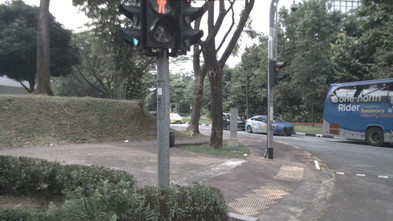}{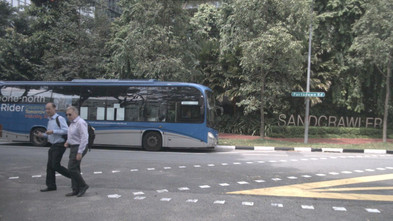}{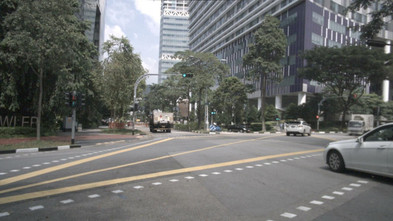}
\gradCamAttributionRowThree{VAD-Tiny}{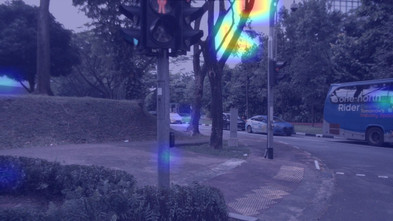}{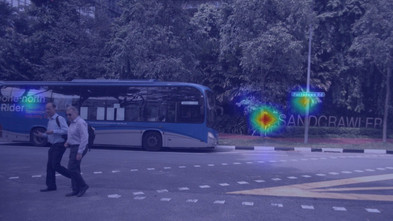}{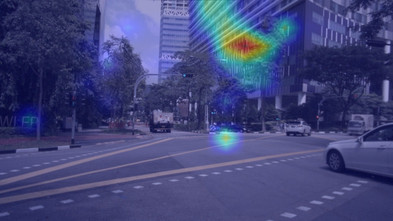}
\gradCamAttributionRowThree{+ PAVER}{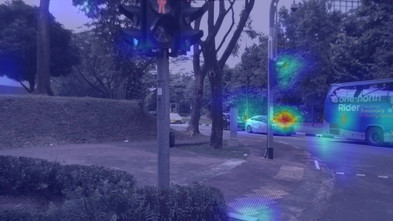}{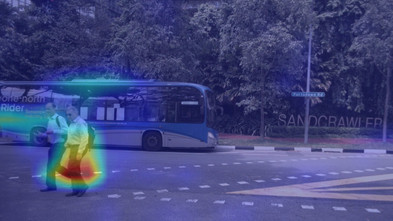}{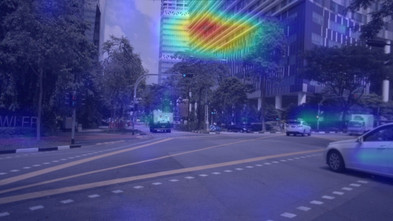}
\caption{Grad-CAM results on the nuScenes validation set.}
\label{fig:supp_vad_tiny_gradcam_three_camera_043}
\vspace{8pt}
\gradCamCameraHeader
\gradCamInputRowThree{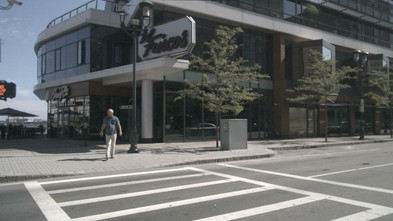}{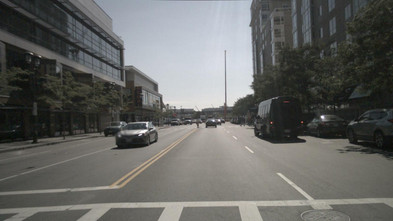}{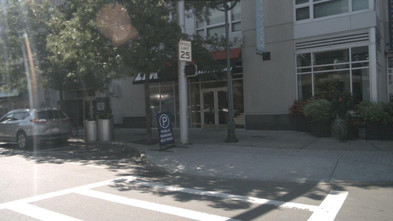}
\gradCamAttributionRowThree{VAD-Tiny}{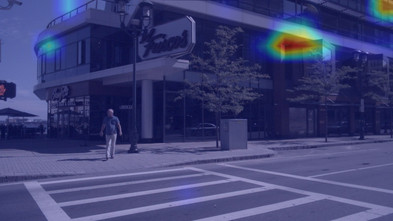}{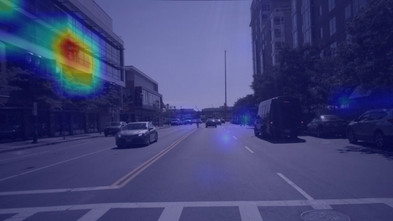}{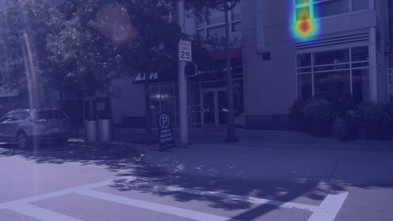}
\gradCamAttributionRowThree{+ PAVER}{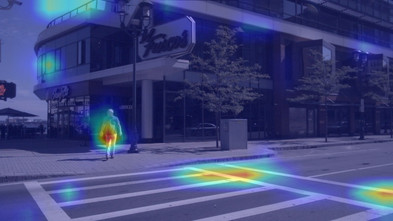}{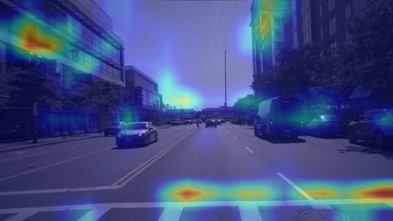}{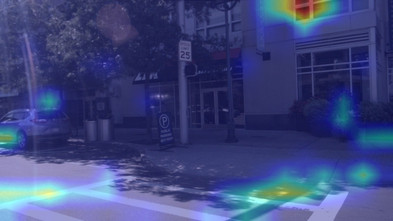}
\caption{Grad-CAM results on the nuScenes validation set.}
\label{fig:supp_vad_tiny_gradcam_three_camera_046}
\end{figure*}
\clearpage

\ifdefined\arxivpreprint\else
\begin{figure*}[p]
\centering
\gradCamCameraHeader
\gradCamInputRowThree{figures/analysis/gradcam_vad_tiny/panels/rank_053__16b5e2757a4647ed97cd9f54f0c42e1c__front_left__input.png}{figures/analysis/gradcam_vad_tiny/panels/rank_053__16b5e2757a4647ed97cd9f54f0c42e1c__front__input.png}{figures/analysis/gradcam_vad_tiny/panels/rank_053__16b5e2757a4647ed97cd9f54f0c42e1c__front_right__input.png}
\gradCamAttributionRowThree{VAD-Tiny}{figures/analysis/gradcam_vad_tiny/panels/rank_053__16b5e2757a4647ed97cd9f54f0c42e1c__front_left__vad_tiny_bev.png}{figures/analysis/gradcam_vad_tiny/panels/rank_053__16b5e2757a4647ed97cd9f54f0c42e1c__front__vad_tiny_bev.png}{figures/analysis/gradcam_vad_tiny/panels/rank_053__16b5e2757a4647ed97cd9f54f0c42e1c__front_right__vad_tiny_bev.png}
\gradCamAttributionRowThree{+ PAVER}{figures/analysis/gradcam_vad_tiny/panels/rank_053__16b5e2757a4647ed97cd9f54f0c42e1c__front_left__paver_bev.png}{figures/analysis/gradcam_vad_tiny/panels/rank_053__16b5e2757a4647ed97cd9f54f0c42e1c__front__paver_bev.png}{figures/analysis/gradcam_vad_tiny/panels/rank_053__16b5e2757a4647ed97cd9f54f0c42e1c__front_right__paver_bev.png}
\caption{Grad-CAM results on the nuScenes validation set.}
\label{fig:supp_vad_tiny_gradcam_three_camera_053}
\vspace{8pt}
\gradCamCameraHeader
\gradCamInputRowThree{figures/analysis/gradcam_vad_tiny/panels/rank_066__371ea4a12b5d4785a1b0c95f288ffed4__front_left__input.png}{figures/analysis/gradcam_vad_tiny/panels/rank_066__371ea4a12b5d4785a1b0c95f288ffed4__front__input.png}{figures/analysis/gradcam_vad_tiny/panels/rank_066__371ea4a12b5d4785a1b0c95f288ffed4__front_right__input.png}
\gradCamAttributionRowThree{VAD-Tiny}{figures/analysis/gradcam_vad_tiny/panels/rank_066__371ea4a12b5d4785a1b0c95f288ffed4__front_left__vad_tiny_bev.png}{figures/analysis/gradcam_vad_tiny/panels/rank_066__371ea4a12b5d4785a1b0c95f288ffed4__front__vad_tiny_bev.png}{figures/analysis/gradcam_vad_tiny/panels/rank_066__371ea4a12b5d4785a1b0c95f288ffed4__front_right__vad_tiny_bev.png}
\gradCamAttributionRowThree{+ PAVER}{figures/analysis/gradcam_vad_tiny/panels/rank_066__371ea4a12b5d4785a1b0c95f288ffed4__front_left__paver_bev.png}{figures/analysis/gradcam_vad_tiny/panels/rank_066__371ea4a12b5d4785a1b0c95f288ffed4__front__paver_bev.png}{figures/analysis/gradcam_vad_tiny/panels/rank_066__371ea4a12b5d4785a1b0c95f288ffed4__front_right__paver_bev.png}
\caption{Grad-CAM results on the nuScenes validation set.}
\label{fig:supp_vad_tiny_gradcam_three_camera_066}
\end{figure*}
\fi
\clearpage

\begin{figure*}[p]
\centering
\gradCamCameraHeader
\gradCamInputRowThree{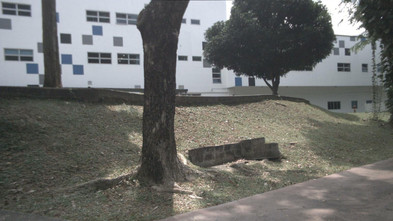}{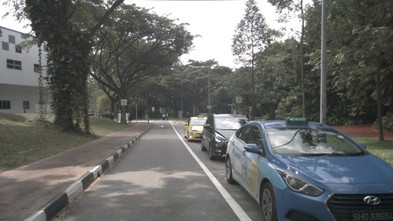}{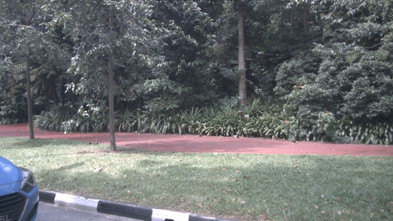}
\gradCamAttributionRowThree{VAD-Tiny}{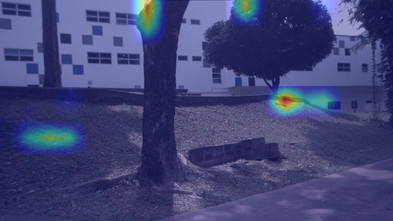}{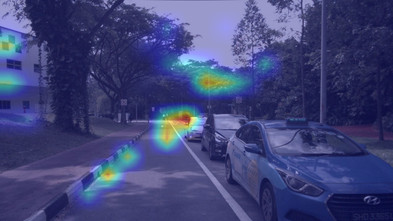}{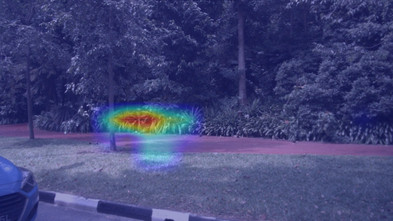}
\gradCamAttributionRowThree{+ PAVER}{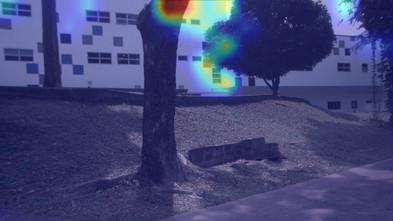}{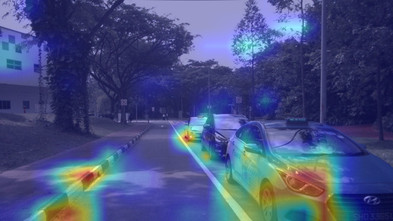}{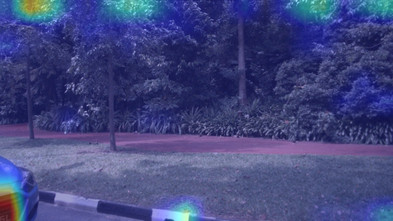}
\caption{Grad-CAM results on the nuScenes validation set.}
\label{fig:supp_vad_tiny_gradcam_three_camera_090}
\vspace{8pt}
\gradCamCameraHeader
\gradCamInputRowThree{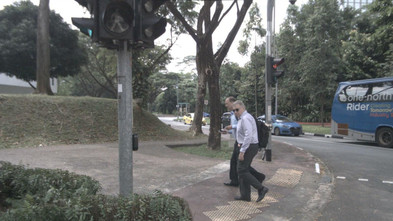}{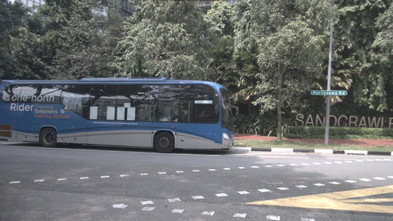}{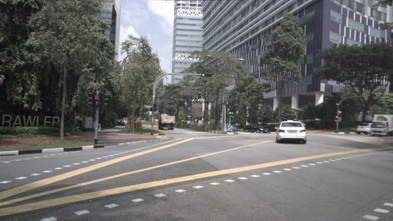}
\gradCamAttributionRowThree{VAD-Tiny}{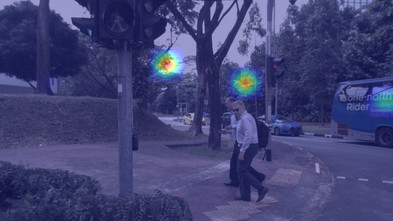}{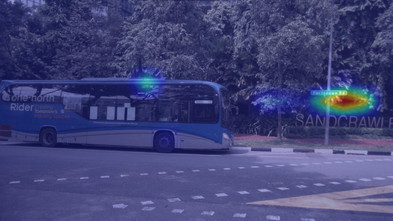}{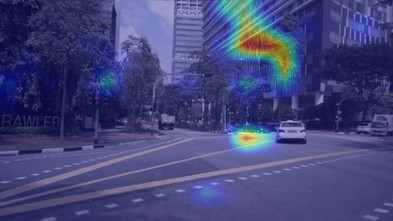}
\gradCamAttributionRowThree{+ PAVER}{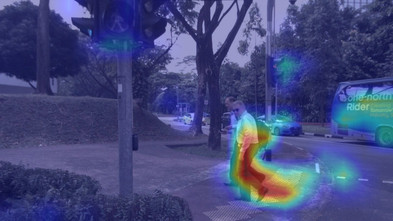}{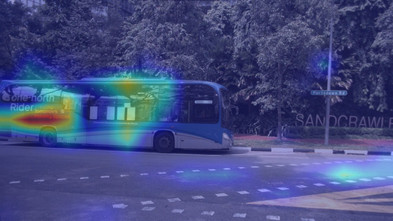}{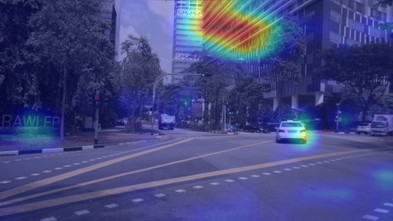}
\caption{Grad-CAM results on the nuScenes validation set.}
\label{fig:supp_vad_tiny_gradcam_three_camera_109}
\end{figure*}
\clearpage

\ifdefined\arxivpreprint\else
\begin{figure*}[p]
\centering
\gradCamCameraHeader
\gradCamInputRowThree{figures/analysis/gradcam_vad_tiny/panels/rank_125__d0501e0542884ab1a4790236b16bc1d4__front_left__input.png}{figures/analysis/gradcam_vad_tiny/panels/rank_125__d0501e0542884ab1a4790236b16bc1d4__front__input.png}{figures/analysis/gradcam_vad_tiny/panels/rank_125__d0501e0542884ab1a4790236b16bc1d4__front_right__input.png}
\gradCamAttributionRowThree{VAD-Tiny}{figures/analysis/gradcam_vad_tiny/panels/rank_125__d0501e0542884ab1a4790236b16bc1d4__front_left__vad_tiny_bev.png}{figures/analysis/gradcam_vad_tiny/panels/rank_125__d0501e0542884ab1a4790236b16bc1d4__front__vad_tiny_bev.png}{figures/analysis/gradcam_vad_tiny/panels/rank_125__d0501e0542884ab1a4790236b16bc1d4__front_right__vad_tiny_bev.png}
\gradCamAttributionRowThree{+ PAVER}{figures/analysis/gradcam_vad_tiny/panels/rank_125__d0501e0542884ab1a4790236b16bc1d4__front_left__paver_bev.png}{figures/analysis/gradcam_vad_tiny/panels/rank_125__d0501e0542884ab1a4790236b16bc1d4__front__paver_bev.png}{figures/analysis/gradcam_vad_tiny/panels/rank_125__d0501e0542884ab1a4790236b16bc1d4__front_right__paver_bev.png}
\caption{Grad-CAM results on the nuScenes validation set.}
\label{fig:supp_vad_tiny_gradcam_three_camera_125}
\vspace{8pt}
\gradCamCameraHeader
\gradCamInputRowThree{figures/analysis/gradcam_vad_tiny/panels/rank_156__b6c420c3a5bd4a219b1cb82ee5ea0aa7__front_left__input.png}{figures/analysis/gradcam_vad_tiny/panels/rank_156__b6c420c3a5bd4a219b1cb82ee5ea0aa7__front__input.png}{figures/analysis/gradcam_vad_tiny/panels/rank_156__b6c420c3a5bd4a219b1cb82ee5ea0aa7__front_right__input.png}
\gradCamAttributionRowThree{VAD-Tiny}{figures/analysis/gradcam_vad_tiny/panels/rank_156__b6c420c3a5bd4a219b1cb82ee5ea0aa7__front_left__vad_tiny_bev.png}{figures/analysis/gradcam_vad_tiny/panels/rank_156__b6c420c3a5bd4a219b1cb82ee5ea0aa7__front__vad_tiny_bev.png}{figures/analysis/gradcam_vad_tiny/panels/rank_156__b6c420c3a5bd4a219b1cb82ee5ea0aa7__front_right__vad_tiny_bev.png}
\gradCamAttributionRowThree{+ PAVER}{figures/analysis/gradcam_vad_tiny/panels/rank_156__b6c420c3a5bd4a219b1cb82ee5ea0aa7__front_left__paver_bev.png}{figures/analysis/gradcam_vad_tiny/panels/rank_156__b6c420c3a5bd4a219b1cb82ee5ea0aa7__front__paver_bev.png}{figures/analysis/gradcam_vad_tiny/panels/rank_156__b6c420c3a5bd4a219b1cb82ee5ea0aa7__front_right__paver_bev.png}
\caption{Grad-CAM results on the nuScenes validation set.}
\label{fig:supp_vad_tiny_gradcam_three_camera_156}
\end{figure*}
\fi
\clearpage

\ifdefined\arxivpreprint\else
\begin{figure*}[p]
\centering
\gradCamCameraHeader
\gradCamInputRowThree{figures/analysis/gradcam_vad_tiny/panels/rank_199__9f00ff39034d4e11ac09542c909742c9__front_left__input.png}{figures/analysis/gradcam_vad_tiny/panels/rank_199__9f00ff39034d4e11ac09542c909742c9__front__input.png}{figures/analysis/gradcam_vad_tiny/panels/rank_199__9f00ff39034d4e11ac09542c909742c9__front_right__input.png}
\gradCamAttributionRowThree{VAD-Tiny}{figures/analysis/gradcam_vad_tiny/panels/rank_199__9f00ff39034d4e11ac09542c909742c9__front_left__vad_tiny_bev.png}{figures/analysis/gradcam_vad_tiny/panels/rank_199__9f00ff39034d4e11ac09542c909742c9__front__vad_tiny_bev.png}{figures/analysis/gradcam_vad_tiny/panels/rank_199__9f00ff39034d4e11ac09542c909742c9__front_right__vad_tiny_bev.png}
\gradCamAttributionRowThree{+ PAVER}{figures/analysis/gradcam_vad_tiny/panels/rank_199__9f00ff39034d4e11ac09542c909742c9__front_left__paver_bev.png}{figures/analysis/gradcam_vad_tiny/panels/rank_199__9f00ff39034d4e11ac09542c909742c9__front__paver_bev.png}{figures/analysis/gradcam_vad_tiny/panels/rank_199__9f00ff39034d4e11ac09542c909742c9__front_right__paver_bev.png}
\caption{Grad-CAM results on the nuScenes validation set.}
\label{fig:supp_vad_tiny_gradcam_three_camera_199}
\vspace{8pt}
\gradCamCameraHeader
\gradCamInputRowThree{figures/analysis/gradcam_vad_tiny/panels/rank_202__4518c3b25dc5423cb72619bcad0d7091__front_left__input.png}{figures/analysis/gradcam_vad_tiny/panels/rank_202__4518c3b25dc5423cb72619bcad0d7091__front__input.png}{figures/analysis/gradcam_vad_tiny/panels/rank_202__4518c3b25dc5423cb72619bcad0d7091__front_right__input.png}
\gradCamAttributionRowThree{VAD-Tiny}{figures/analysis/gradcam_vad_tiny/panels/rank_202__4518c3b25dc5423cb72619bcad0d7091__front_left__vad_tiny_bev.png}{figures/analysis/gradcam_vad_tiny/panels/rank_202__4518c3b25dc5423cb72619bcad0d7091__front__vad_tiny_bev.png}{figures/analysis/gradcam_vad_tiny/panels/rank_202__4518c3b25dc5423cb72619bcad0d7091__front_right__vad_tiny_bev.png}
\gradCamAttributionRowThree{+ PAVER}{figures/analysis/gradcam_vad_tiny/panels/rank_202__4518c3b25dc5423cb72619bcad0d7091__front_left__paver_bev.png}{figures/analysis/gradcam_vad_tiny/panels/rank_202__4518c3b25dc5423cb72619bcad0d7091__front__paver_bev.png}{figures/analysis/gradcam_vad_tiny/panels/rank_202__4518c3b25dc5423cb72619bcad0d7091__front_right__paver_bev.png}
\caption{Grad-CAM results on the nuScenes validation set.}
\label{fig:supp_vad_tiny_gradcam_three_camera_202}
\end{figure*}
\fi
\clearpage

\ifdefined\arxivpreprint\else
\begin{figure*}[p]
\centering
\gradCamCameraHeader
\gradCamInputRowThree{figures/analysis/gradcam_vad_tiny/panels/rank_203__be83af41951d42739ce790d878ea02f9__front_left__input.png}{figures/analysis/gradcam_vad_tiny/panels/rank_203__be83af41951d42739ce790d878ea02f9__front__input.png}{figures/analysis/gradcam_vad_tiny/panels/rank_203__be83af41951d42739ce790d878ea02f9__front_right__input.png}
\gradCamAttributionRowThree{VAD-Tiny}{figures/analysis/gradcam_vad_tiny/panels/rank_203__be83af41951d42739ce790d878ea02f9__front_left__vad_tiny_bev.png}{figures/analysis/gradcam_vad_tiny/panels/rank_203__be83af41951d42739ce790d878ea02f9__front__vad_tiny_bev.png}{figures/analysis/gradcam_vad_tiny/panels/rank_203__be83af41951d42739ce790d878ea02f9__front_right__vad_tiny_bev.png}
\gradCamAttributionRowThree{+ PAVER}{figures/analysis/gradcam_vad_tiny/panels/rank_203__be83af41951d42739ce790d878ea02f9__front_left__paver_bev.png}{figures/analysis/gradcam_vad_tiny/panels/rank_203__be83af41951d42739ce790d878ea02f9__front__paver_bev.png}{figures/analysis/gradcam_vad_tiny/panels/rank_203__be83af41951d42739ce790d878ea02f9__front_right__paver_bev.png}
\caption{Grad-CAM results on the nuScenes validation set.}
\label{fig:supp_vad_tiny_gradcam_three_camera_203}
\vspace{8pt}
\gradCamCameraHeader
\gradCamInputRowThree{figures/analysis/gradcam_vad_tiny/panels/rank_211__621ecaf28ab34822b8747831b42f8524__front_left__input.png}{figures/analysis/gradcam_vad_tiny/panels/rank_211__621ecaf28ab34822b8747831b42f8524__front__input.png}{figures/analysis/gradcam_vad_tiny/panels/rank_211__621ecaf28ab34822b8747831b42f8524__front_right__input.png}
\gradCamAttributionRowThree{VAD-Tiny}{figures/analysis/gradcam_vad_tiny/panels/rank_211__621ecaf28ab34822b8747831b42f8524__front_left__vad_tiny_bev.png}{figures/analysis/gradcam_vad_tiny/panels/rank_211__621ecaf28ab34822b8747831b42f8524__front__vad_tiny_bev.png}{figures/analysis/gradcam_vad_tiny/panels/rank_211__621ecaf28ab34822b8747831b42f8524__front_right__vad_tiny_bev.png}
\gradCamAttributionRowThree{+ PAVER}{figures/analysis/gradcam_vad_tiny/panels/rank_211__621ecaf28ab34822b8747831b42f8524__front_left__paver_bev.png}{figures/analysis/gradcam_vad_tiny/panels/rank_211__621ecaf28ab34822b8747831b42f8524__front__paver_bev.png}{figures/analysis/gradcam_vad_tiny/panels/rank_211__621ecaf28ab34822b8747831b42f8524__front_right__paver_bev.png}
\caption{Grad-CAM results on the nuScenes validation set.}
\label{fig:supp_vad_tiny_gradcam_three_camera_211}
\end{figure*}
\fi
\clearpage

\ifdefined\arxivpreprint\else
\begin{figure*}[p]
\centering
\gradCamCameraHeader
\gradCamInputRowThree{figures/analysis/gradcam_vad_tiny/panels/rank_225__1a1dc7ef4955481ba3755fbaa39a53d8__front_left__input.png}{figures/analysis/gradcam_vad_tiny/panels/rank_225__1a1dc7ef4955481ba3755fbaa39a53d8__front__input.png}{figures/analysis/gradcam_vad_tiny/panels/rank_225__1a1dc7ef4955481ba3755fbaa39a53d8__front_right__input.png}
\gradCamAttributionRowThree{VAD-Tiny}{figures/analysis/gradcam_vad_tiny/panels/rank_225__1a1dc7ef4955481ba3755fbaa39a53d8__front_left__vad_tiny_bev.png}{figures/analysis/gradcam_vad_tiny/panels/rank_225__1a1dc7ef4955481ba3755fbaa39a53d8__front__vad_tiny_bev.png}{figures/analysis/gradcam_vad_tiny/panels/rank_225__1a1dc7ef4955481ba3755fbaa39a53d8__front_right__vad_tiny_bev.png}
\gradCamAttributionRowThree{+ PAVER}{figures/analysis/gradcam_vad_tiny/panels/rank_225__1a1dc7ef4955481ba3755fbaa39a53d8__front_left__paver_bev.png}{figures/analysis/gradcam_vad_tiny/panels/rank_225__1a1dc7ef4955481ba3755fbaa39a53d8__front__paver_bev.png}{figures/analysis/gradcam_vad_tiny/panels/rank_225__1a1dc7ef4955481ba3755fbaa39a53d8__front_right__paver_bev.png}
\caption{Grad-CAM results on the nuScenes validation set.}
\label{fig:supp_vad_tiny_gradcam_three_camera_225}
\vspace{8pt}
\gradCamCameraHeader
\gradCamInputRowThree{figures/analysis/gradcam_vad_tiny/panels/rank_226__ed00e1196ea74e0795313afc75d39dcf__front_left__input.png}{figures/analysis/gradcam_vad_tiny/panels/rank_226__ed00e1196ea74e0795313afc75d39dcf__front__input.png}{figures/analysis/gradcam_vad_tiny/panels/rank_226__ed00e1196ea74e0795313afc75d39dcf__front_right__input.png}
\gradCamAttributionRowThree{VAD-Tiny}{figures/analysis/gradcam_vad_tiny/panels/rank_226__ed00e1196ea74e0795313afc75d39dcf__front_left__vad_tiny_bev.png}{figures/analysis/gradcam_vad_tiny/panels/rank_226__ed00e1196ea74e0795313afc75d39dcf__front__vad_tiny_bev.png}{figures/analysis/gradcam_vad_tiny/panels/rank_226__ed00e1196ea74e0795313afc75d39dcf__front_right__vad_tiny_bev.png}
\gradCamAttributionRowThree{+ PAVER}{figures/analysis/gradcam_vad_tiny/panels/rank_226__ed00e1196ea74e0795313afc75d39dcf__front_left__paver_bev.png}{figures/analysis/gradcam_vad_tiny/panels/rank_226__ed00e1196ea74e0795313afc75d39dcf__front__paver_bev.png}{figures/analysis/gradcam_vad_tiny/panels/rank_226__ed00e1196ea74e0795313afc75d39dcf__front_right__paver_bev.png}
\caption{Grad-CAM results on the nuScenes validation set.}
\label{fig:supp_vad_tiny_gradcam_three_camera_226}
\end{figure*}
\fi
\clearpage

\ifdefined\arxivpreprint\else
\begin{figure*}[p]
\centering
\gradCamCameraHeader
\gradCamInputRowThree{figures/analysis/gradcam_vad_tiny/panels/rank_265__67aad7ad948f44f8af668ea8389bdd52__front_left__input.png}{figures/analysis/gradcam_vad_tiny/panels/rank_265__67aad7ad948f44f8af668ea8389bdd52__front__input.png}{figures/analysis/gradcam_vad_tiny/panels/rank_265__67aad7ad948f44f8af668ea8389bdd52__front_right__input.png}
\gradCamAttributionRowThree{VAD-Tiny}{figures/analysis/gradcam_vad_tiny/panels/rank_265__67aad7ad948f44f8af668ea8389bdd52__front_left__vad_tiny_bev.png}{figures/analysis/gradcam_vad_tiny/panels/rank_265__67aad7ad948f44f8af668ea8389bdd52__front__vad_tiny_bev.png}{figures/analysis/gradcam_vad_tiny/panels/rank_265__67aad7ad948f44f8af668ea8389bdd52__front_right__vad_tiny_bev.png}
\gradCamAttributionRowThree{+ PAVER}{figures/analysis/gradcam_vad_tiny/panels/rank_265__67aad7ad948f44f8af668ea8389bdd52__front_left__paver_bev.png}{figures/analysis/gradcam_vad_tiny/panels/rank_265__67aad7ad948f44f8af668ea8389bdd52__front__paver_bev.png}{figures/analysis/gradcam_vad_tiny/panels/rank_265__67aad7ad948f44f8af668ea8389bdd52__front_right__paver_bev.png}
\caption{Grad-CAM response to a traffic signal on the nuScenes validation set. PAVER concentrates attribution on the signal.}
\label{fig:supp_vad_tiny_gradcam_three_camera_265}
\vspace{8pt}
\gradCamCameraHeader
\gradCamInputRowThree{figures/analysis/gradcam_vad_tiny/panels/rank_304__f1a2b04d085842dfa616d66eb41f1dac__front_left__input.png}{figures/analysis/gradcam_vad_tiny/panels/rank_304__f1a2b04d085842dfa616d66eb41f1dac__front__input.png}{figures/analysis/gradcam_vad_tiny/panels/rank_304__f1a2b04d085842dfa616d66eb41f1dac__front_right__input.png}
\gradCamAttributionRowThree{VAD-Tiny}{figures/analysis/gradcam_vad_tiny/panels/rank_304__f1a2b04d085842dfa616d66eb41f1dac__front_left__vad_tiny_bev.png}{figures/analysis/gradcam_vad_tiny/panels/rank_304__f1a2b04d085842dfa616d66eb41f1dac__front__vad_tiny_bev.png}{figures/analysis/gradcam_vad_tiny/panels/rank_304__f1a2b04d085842dfa616d66eb41f1dac__front_right__vad_tiny_bev.png}
\gradCamAttributionRowThree{+ PAVER}{figures/analysis/gradcam_vad_tiny/panels/rank_304__f1a2b04d085842dfa616d66eb41f1dac__front_left__paver_bev.png}{figures/analysis/gradcam_vad_tiny/panels/rank_304__f1a2b04d085842dfa616d66eb41f1dac__front__paver_bev.png}{figures/analysis/gradcam_vad_tiny/panels/rank_304__f1a2b04d085842dfa616d66eb41f1dac__front_right__paver_bev.png}
\caption{Grad-CAM results on the nuScenes validation set.}
\label{fig:supp_vad_tiny_gradcam_three_camera_304}
\end{figure*}
\fi
\clearpage

\begin{figure*}[p]
\centering
\gradCamCameraHeader
\gradCamInputRowThree{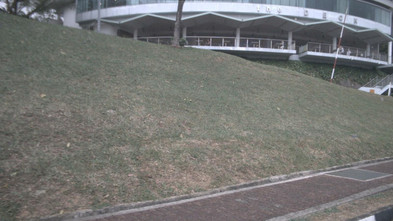}{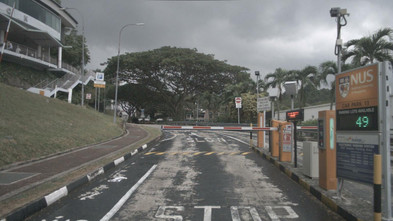}{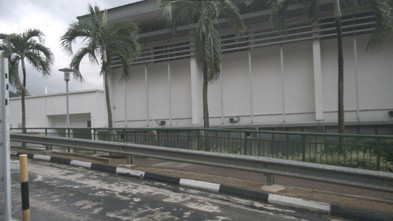}
\gradCamAttributionRowThree{VAD-Tiny}{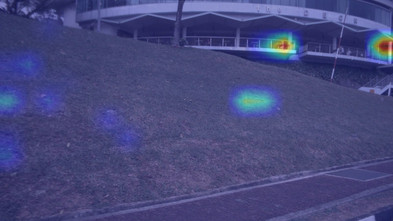}{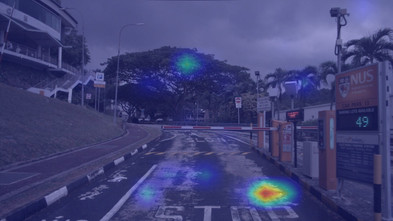}{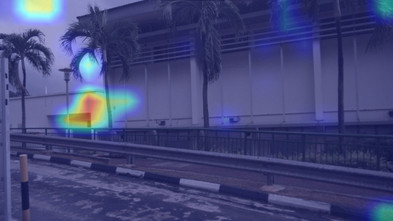}
\gradCamAttributionRowThree{+ PAVER}{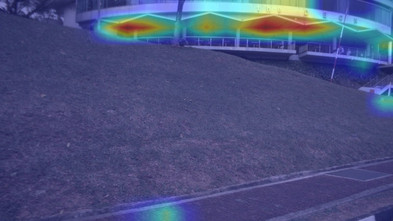}{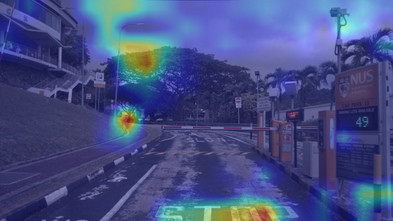}{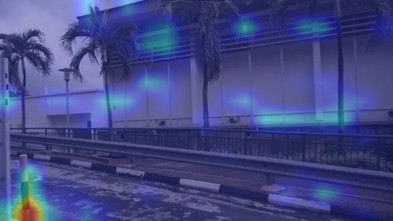}
\caption{Grad-CAM response to a road-surface stop marking on the nuScenes validation set. PAVER concentrates attribution on the marking.}
\label{fig:supp_vad_tiny_gradcam_three_camera_359}
\vspace{8pt}
\gradCamCameraHeader
\gradCamInputRowThree{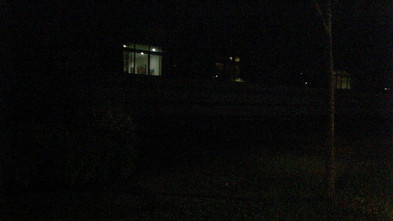}{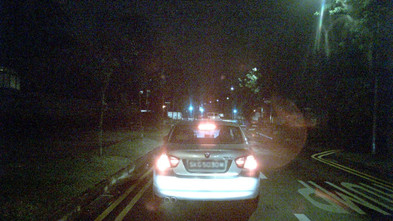}{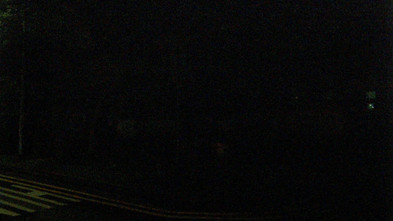}
\gradCamAttributionRowThree{VAD-Tiny}{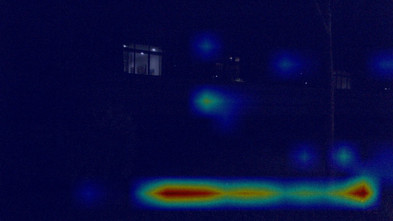}{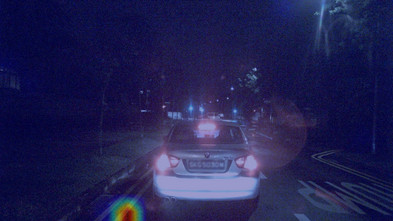}{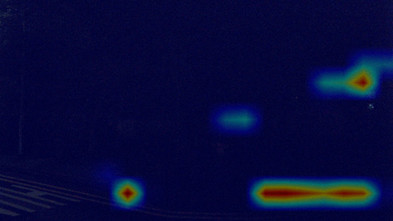}
\gradCamAttributionRowThree{+ PAVER}{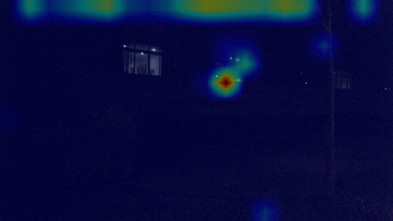}{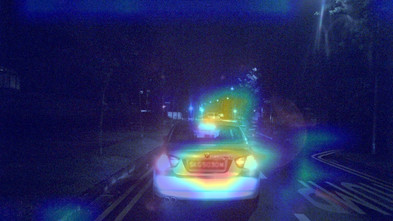}{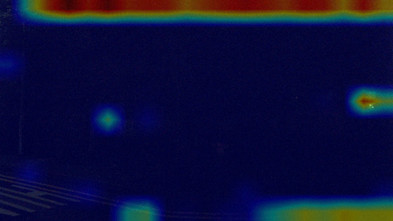}
\caption{Grad-CAM results on the nuScenes validation set.}
\label{fig:supp_vad_tiny_gradcam_three_camera_415}
\end{figure*}
\clearpage

\begin{figure*}[p]
\centering
\gradCamCameraHeader
\gradCamInputRowThree{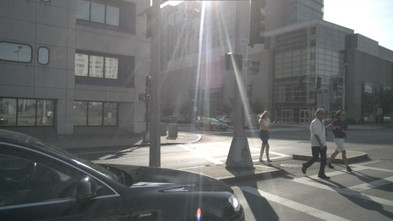}{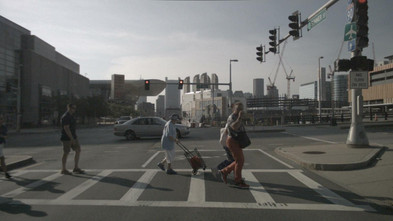}{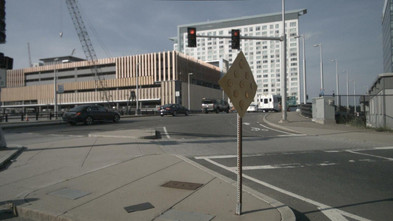}
\gradCamAttributionRowThree{VAD-Tiny}{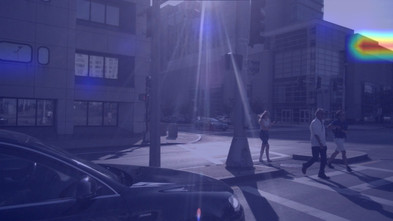}{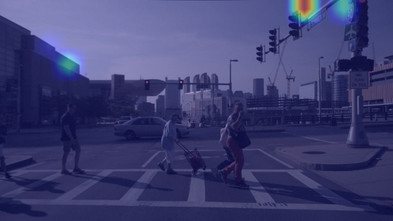}{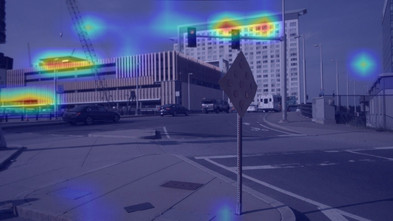}
\gradCamAttributionRowThree{+ PAVER}{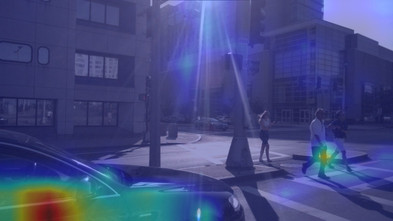}{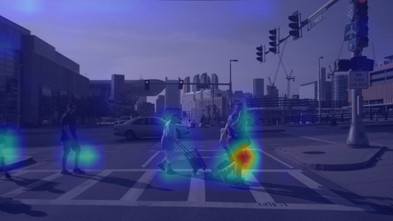}{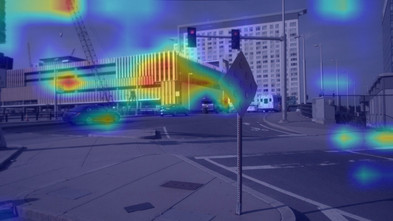}
\caption{Grad-CAM results on the nuScenes validation set.}
\label{fig:supp_vad_tiny_gradcam_three_camera_495}
\vspace{8pt}
\gradCamCameraHeader
\gradCamInputRowThree{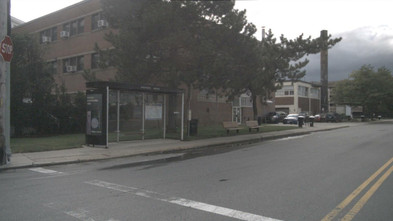}{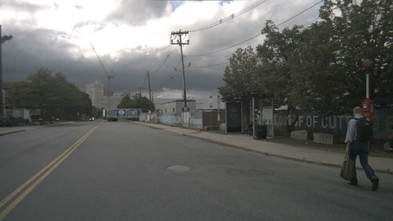}{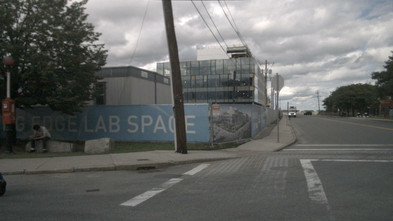}
\gradCamAttributionRowThree{VAD-Tiny}{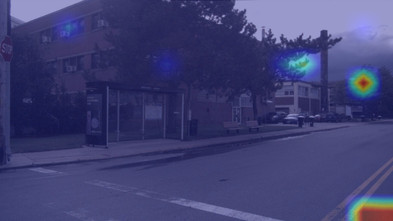}{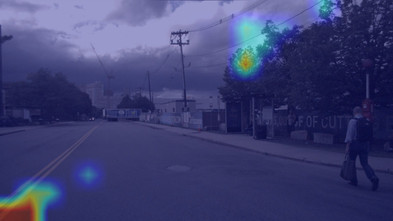}{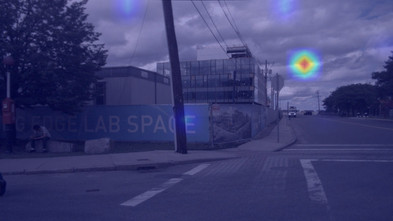}
\gradCamAttributionRowThree{+ PAVER}{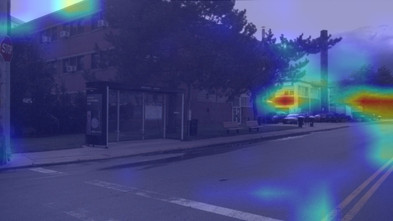}{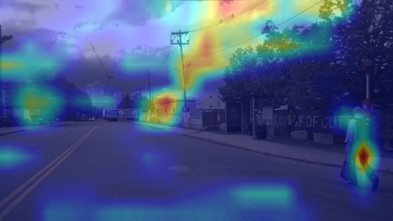}{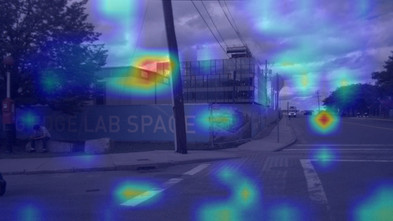}
\caption{Grad-CAM results on the nuScenes validation set.}
\label{fig:supp_vad_tiny_gradcam_three_camera_521}
\end{figure*}
\clearpage

\ifdefined\arxivpreprint\else
\begin{figure*}[p]
\centering
\gradCamCameraHeader
\gradCamInputRowThree{figures/analysis/gradcam_vad_tiny/panels/rank_532__4de7963785f34de7afb9e5aae3ce6f9f__front_left__input.png}{figures/analysis/gradcam_vad_tiny/panels/rank_532__4de7963785f34de7afb9e5aae3ce6f9f__front__input.png}{figures/analysis/gradcam_vad_tiny/panels/rank_532__4de7963785f34de7afb9e5aae3ce6f9f__front_right__input.png}
\gradCamAttributionRowThree{VAD-Tiny}{figures/analysis/gradcam_vad_tiny/panels/rank_532__4de7963785f34de7afb9e5aae3ce6f9f__front_left__vad_tiny_bev.png}{figures/analysis/gradcam_vad_tiny/panels/rank_532__4de7963785f34de7afb9e5aae3ce6f9f__front__vad_tiny_bev.png}{figures/analysis/gradcam_vad_tiny/panels/rank_532__4de7963785f34de7afb9e5aae3ce6f9f__front_right__vad_tiny_bev.png}
\gradCamAttributionRowThree{+ PAVER}{figures/analysis/gradcam_vad_tiny/panels/rank_532__4de7963785f34de7afb9e5aae3ce6f9f__front_left__paver_bev.png}{figures/analysis/gradcam_vad_tiny/panels/rank_532__4de7963785f34de7afb9e5aae3ce6f9f__front__paver_bev.png}{figures/analysis/gradcam_vad_tiny/panels/rank_532__4de7963785f34de7afb9e5aae3ce6f9f__front_right__paver_bev.png}
\caption{Grad-CAM results on the nuScenes validation set.}
\label{fig:supp_vad_tiny_gradcam_three_camera_532}
\vspace{8pt}
\gradCamCameraHeader
\gradCamInputRowThree{figures/analysis/gradcam_vad_tiny/panels/rank_542__e0a7e1df190e48feb416dad36d140c35__front_left__input.png}{figures/analysis/gradcam_vad_tiny/panels/rank_542__e0a7e1df190e48feb416dad36d140c35__front__input.png}{figures/analysis/gradcam_vad_tiny/panels/rank_542__e0a7e1df190e48feb416dad36d140c35__front_right__input.png}
\gradCamAttributionRowThree{VAD-Tiny}{figures/analysis/gradcam_vad_tiny/panels/rank_542__e0a7e1df190e48feb416dad36d140c35__front_left__vad_tiny_bev.png}{figures/analysis/gradcam_vad_tiny/panels/rank_542__e0a7e1df190e48feb416dad36d140c35__front__vad_tiny_bev.png}{figures/analysis/gradcam_vad_tiny/panels/rank_542__e0a7e1df190e48feb416dad36d140c35__front_right__vad_tiny_bev.png}
\gradCamAttributionRowThree{+ PAVER}{figures/analysis/gradcam_vad_tiny/panels/rank_542__e0a7e1df190e48feb416dad36d140c35__front_left__paver_bev.png}{figures/analysis/gradcam_vad_tiny/panels/rank_542__e0a7e1df190e48feb416dad36d140c35__front__paver_bev.png}{figures/analysis/gradcam_vad_tiny/panels/rank_542__e0a7e1df190e48feb416dad36d140c35__front_right__paver_bev.png}
\caption{Grad-CAM results on the nuScenes validation set.}
\label{fig:supp_vad_tiny_gradcam_three_camera_542}
\end{figure*}
\fi
\clearpage

\begin{figure*}[p]
\centering
\gradCamCameraHeader
\gradCamInputRowThree{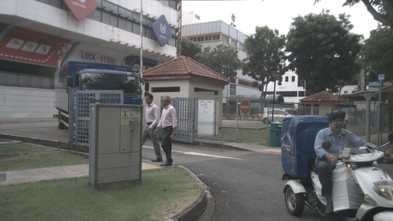}{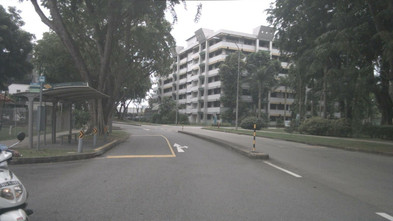}{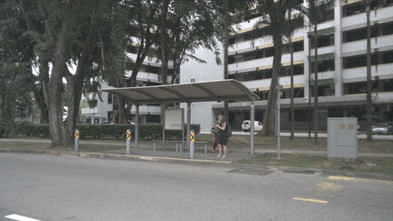}
\gradCamAttributionRowThree{VAD-Tiny}{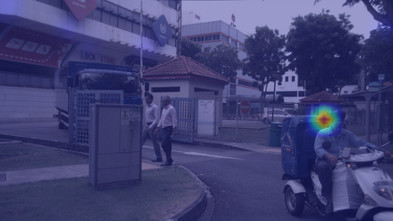}{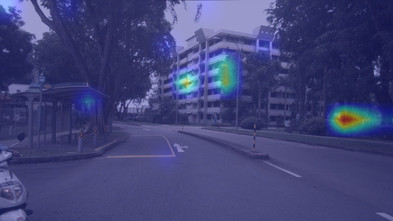}{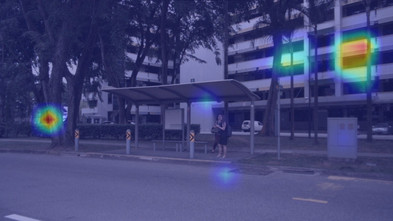}
\gradCamAttributionRowThree{+ PAVER}{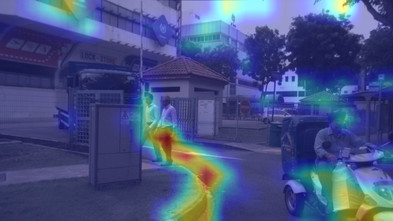}{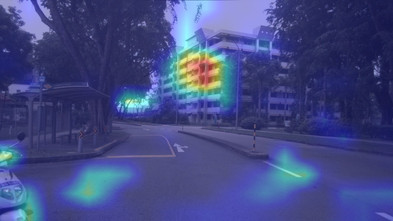}{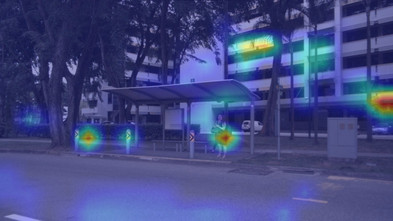}
\caption{Grad-CAM results on the nuScenes validation set.}
\label{fig:supp_vad_tiny_gradcam_three_camera_558}
\vspace{8pt}
\gradCamCameraHeader
\gradCamInputRowThree{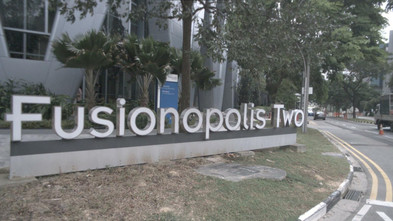}{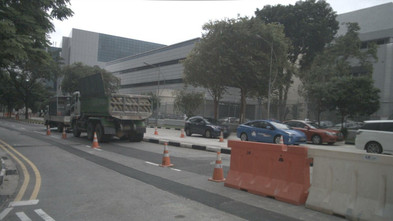}{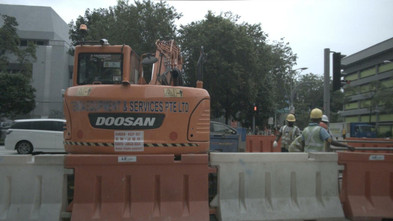}
\gradCamAttributionRowThree{VAD-Tiny}{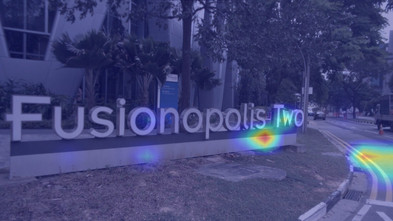}{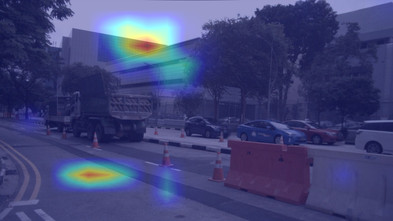}{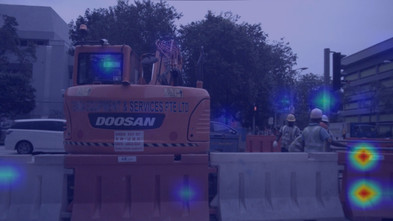}
\gradCamAttributionRowThree{+ PAVER}{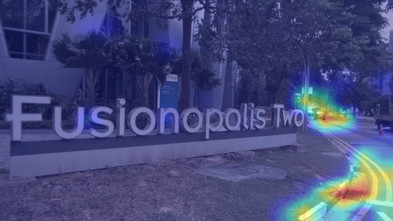}{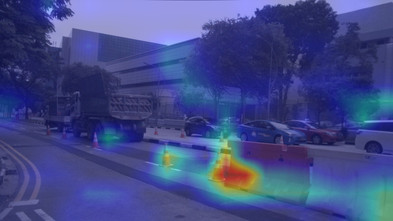}{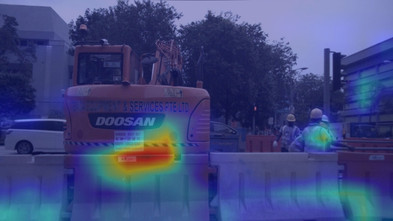}
\caption{Grad-CAM results on the nuScenes validation set.}
\label{fig:supp_vad_tiny_gradcam_three_camera_585}
\end{figure*}
\clearpage

\ifdefined\arxivpreprint\else
\begin{figure*}[p]
\centering
\gradCamCameraHeader
\gradCamInputRowThree{figures/analysis/gradcam_vad_tiny/panels/rank_586__c143943fd6e246448eab88ce1a0aeda9__front_left__input.png}{figures/analysis/gradcam_vad_tiny/panels/rank_586__c143943fd6e246448eab88ce1a0aeda9__front__input.png}{figures/analysis/gradcam_vad_tiny/panels/rank_586__c143943fd6e246448eab88ce1a0aeda9__front_right__input.png}
\gradCamAttributionRowThree{VAD-Tiny}{figures/analysis/gradcam_vad_tiny/panels/rank_586__c143943fd6e246448eab88ce1a0aeda9__front_left__vad_tiny_bev.png}{figures/analysis/gradcam_vad_tiny/panels/rank_586__c143943fd6e246448eab88ce1a0aeda9__front__vad_tiny_bev.png}{figures/analysis/gradcam_vad_tiny/panels/rank_586__c143943fd6e246448eab88ce1a0aeda9__front_right__vad_tiny_bev.png}
\gradCamAttributionRowThree{+ PAVER}{figures/analysis/gradcam_vad_tiny/panels/rank_586__c143943fd6e246448eab88ce1a0aeda9__front_left__paver_bev.png}{figures/analysis/gradcam_vad_tiny/panels/rank_586__c143943fd6e246448eab88ce1a0aeda9__front__paver_bev.png}{figures/analysis/gradcam_vad_tiny/panels/rank_586__c143943fd6e246448eab88ce1a0aeda9__front_right__paver_bev.png}
\caption{Grad-CAM results on the nuScenes validation set.}
\label{fig:supp_vad_tiny_gradcam_three_camera_586}
\vspace{8pt}
\gradCamCameraHeader
\gradCamInputRowThree{figures/analysis/gradcam_vad_tiny/panels/rank_590__cb7c92011a4c48a2a201cd2dc9fb7488__front_left__input.png}{figures/analysis/gradcam_vad_tiny/panels/rank_590__cb7c92011a4c48a2a201cd2dc9fb7488__front__input.png}{figures/analysis/gradcam_vad_tiny/panels/rank_590__cb7c92011a4c48a2a201cd2dc9fb7488__front_right__input.png}
\gradCamAttributionRowThree{VAD-Tiny}{figures/analysis/gradcam_vad_tiny/panels/rank_590__cb7c92011a4c48a2a201cd2dc9fb7488__front_left__vad_tiny_bev.png}{figures/analysis/gradcam_vad_tiny/panels/rank_590__cb7c92011a4c48a2a201cd2dc9fb7488__front__vad_tiny_bev.png}{figures/analysis/gradcam_vad_tiny/panels/rank_590__cb7c92011a4c48a2a201cd2dc9fb7488__front_right__vad_tiny_bev.png}
\gradCamAttributionRowThree{+ PAVER}{figures/analysis/gradcam_vad_tiny/panels/rank_590__cb7c92011a4c48a2a201cd2dc9fb7488__front_left__paver_bev.png}{figures/analysis/gradcam_vad_tiny/panels/rank_590__cb7c92011a4c48a2a201cd2dc9fb7488__front__paver_bev.png}{figures/analysis/gradcam_vad_tiny/panels/rank_590__cb7c92011a4c48a2a201cd2dc9fb7488__front_right__paver_bev.png}
\caption{Grad-CAM results on the nuScenes validation set.}
\label{fig:supp_vad_tiny_gradcam_three_camera_590}
\end{figure*}
\fi
\clearpage

\begin{figure*}[p]
\centering
\gradCamCameraHeader
\gradCamInputRowThree{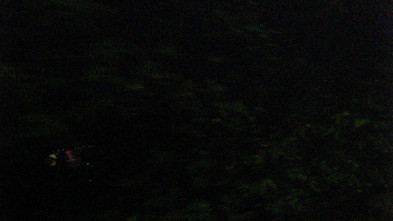}{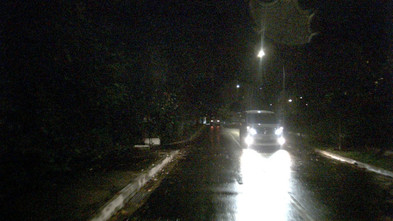}{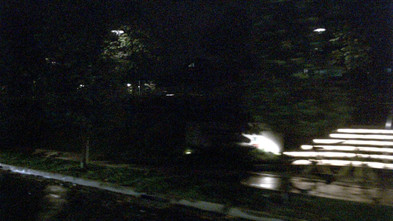}
\gradCamAttributionRowThree{VAD-Tiny}{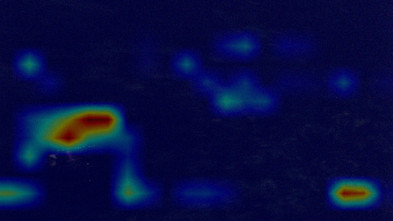}{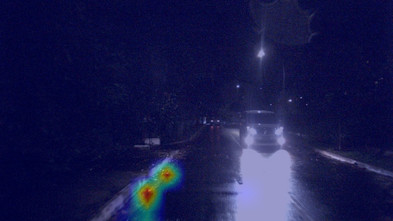}{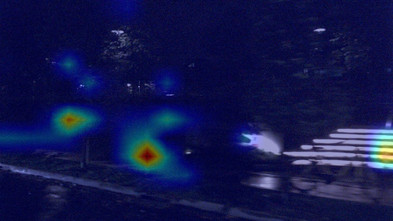}
\gradCamAttributionRowThree{+ PAVER}{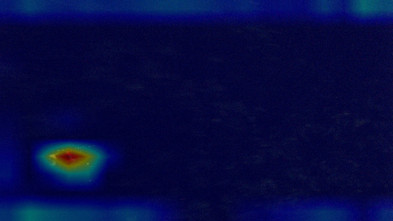}{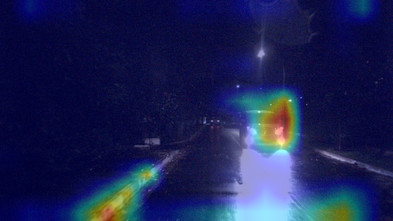}{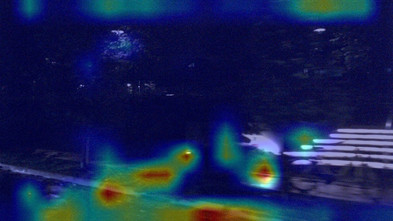}
\caption{Grad-CAM results on the nuScenes validation set.}
\label{fig:supp_vad_tiny_gradcam_three_camera_637}
\vspace{8pt}
\gradCamCameraHeader
\gradCamInputRowThree{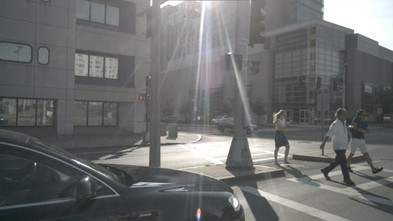}{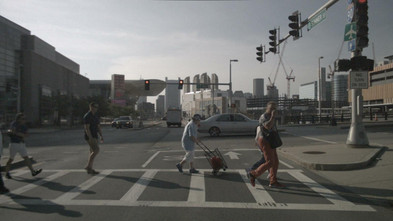}{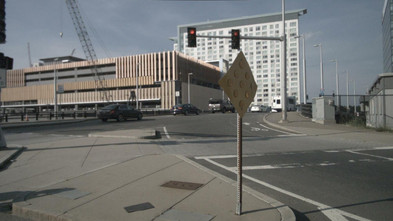}
\gradCamAttributionRowThree{VAD-Tiny}{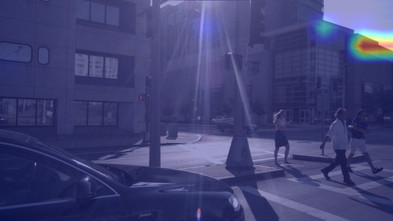}{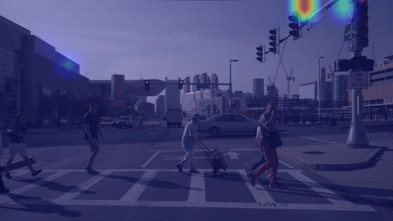}{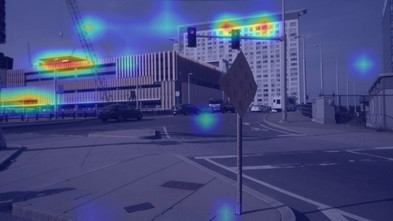}
\gradCamAttributionRowThree{+ PAVER}{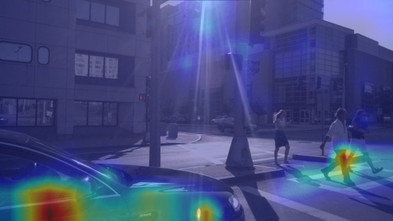}{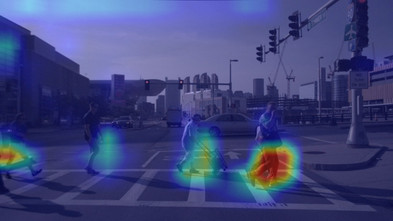}{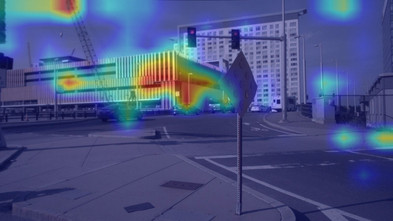}
\caption{Grad-CAM results on the nuScenes validation set.}
\label{fig:supp_vad_tiny_gradcam_three_camera_669}
\end{figure*}
\clearpage

\begin{figure*}[p]
\centering
\gradCamCameraHeader
\gradCamInputRowThree{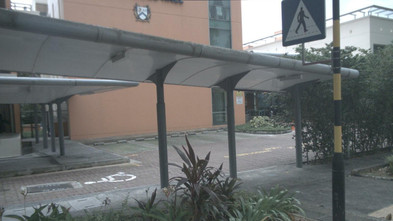}{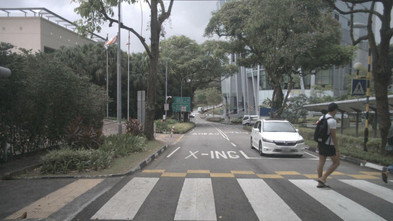}{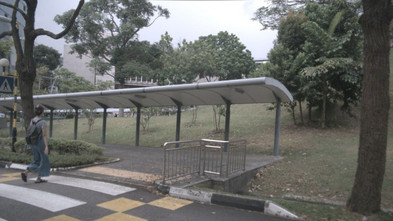}
\gradCamAttributionRowThree{VAD-Tiny}{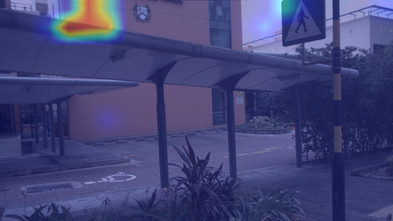}{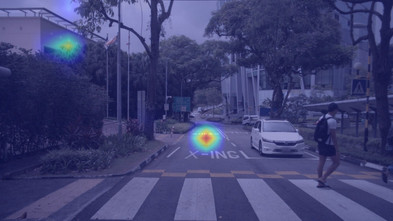}{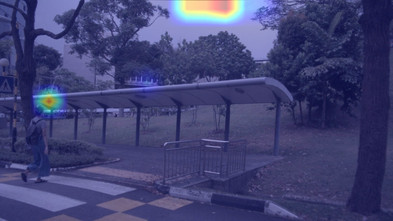}
\gradCamAttributionRowThree{+ PAVER}{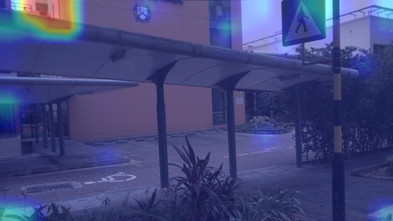}{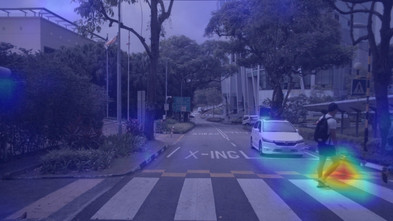}{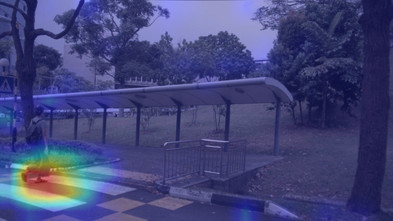}
\caption{Grad-CAM results on the nuScenes validation set.}
\label{fig:supp_vad_tiny_gradcam_three_camera_721}
\vspace{8pt}
\gradCamCameraHeader
\gradCamInputRowThree{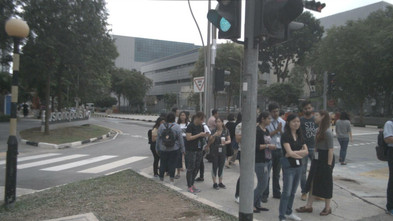}{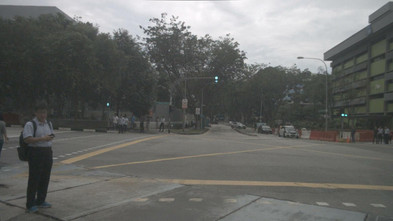}{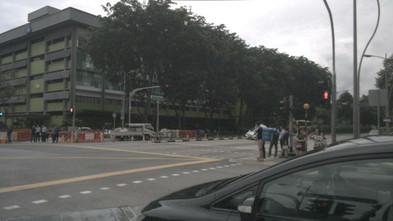}
\gradCamAttributionRowThree{VAD-Tiny}{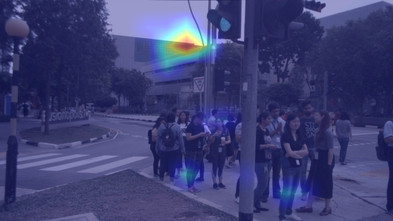}{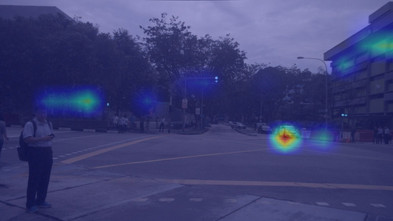}{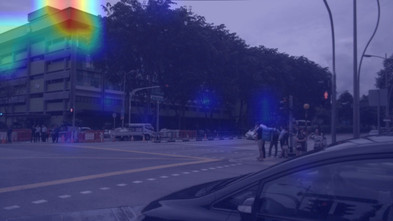}
\gradCamAttributionRowThree{+ PAVER}{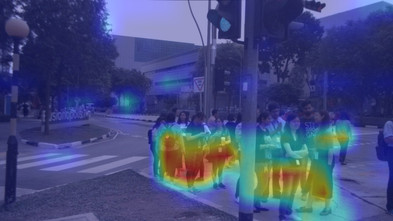}{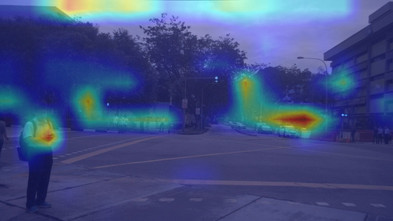}{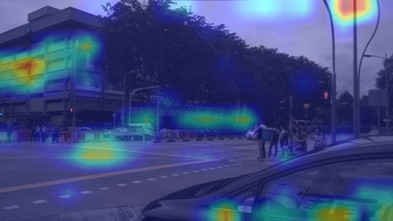}
\caption{Grad-CAM results on the nuScenes validation set.}
\label{fig:supp_vad_tiny_gradcam_three_camera_971}
\end{figure*}
\clearpage

\begin{figure*}[p]
\centering
\gradCamCameraHeader
\gradCamInputRowThree{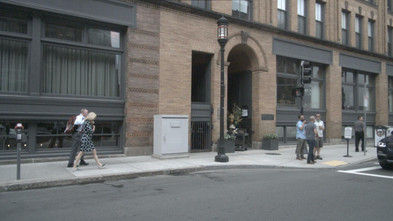}{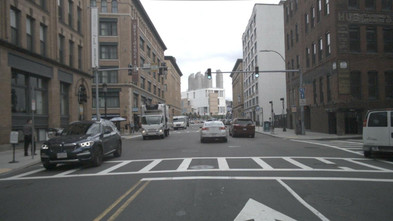}{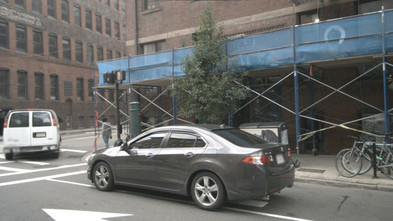}
\gradCamAttributionRowThree{VAD-Tiny}{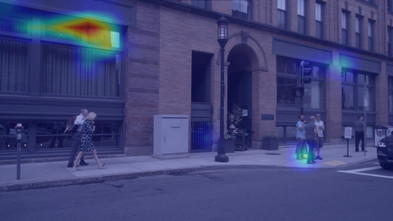}{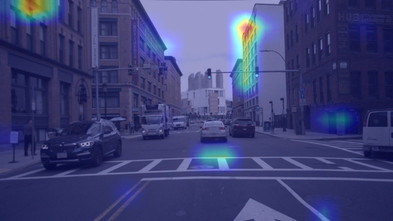}{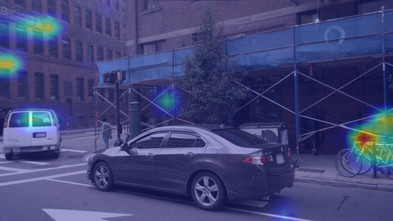}
\gradCamAttributionRowThree{+ PAVER}{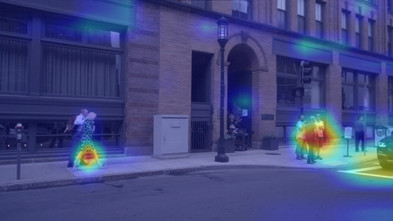}{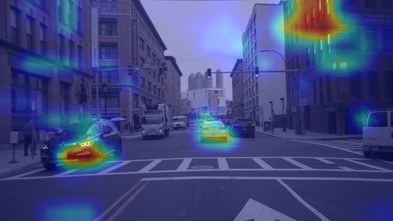}{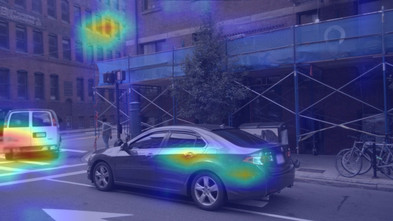}
\caption{Grad-CAM results on the nuScenes validation set.}
\label{fig:supp_vad_tiny_gradcam_three_camera_999}
\vspace{8pt}
\gradCamCameraHeader
\gradCamInputRowThree{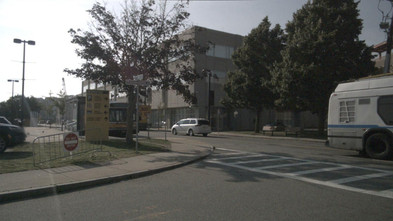}{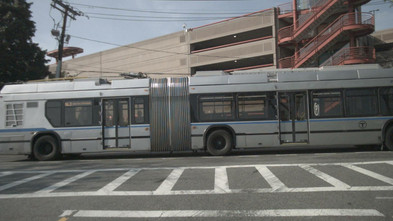}{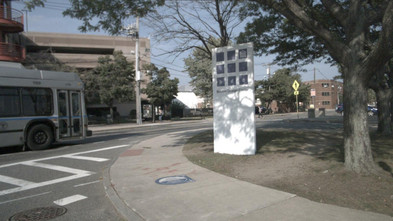}
\gradCamAttributionRowThree{VAD-Tiny}{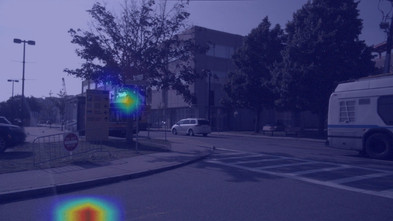}{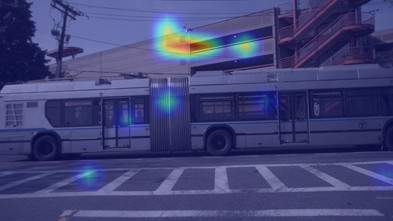}{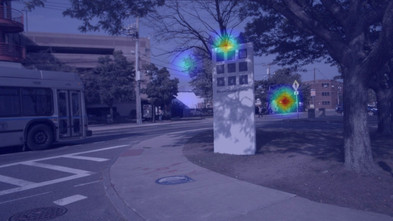}
\gradCamAttributionRowThree{+ PAVER}{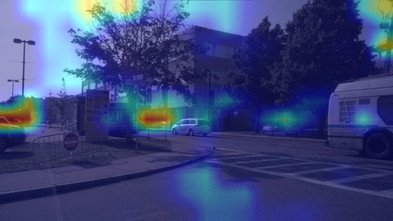}{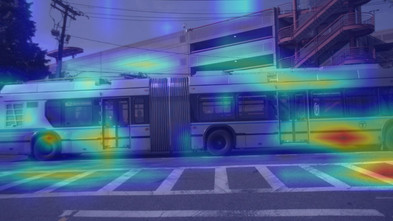}{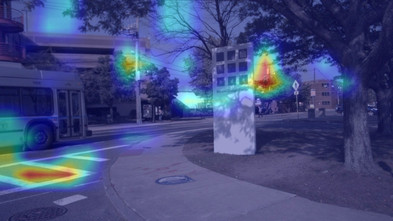}
\caption{Grad-CAM results on the nuScenes validation set.}
\label{fig:supp_vad_tiny_gradcam_three_camera_1022}
\end{figure*}
\clearpage

\begin{figure*}[p]
\centering
\gradCamCameraHeader
\gradCamInputRowThree{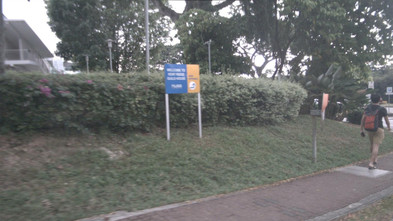}{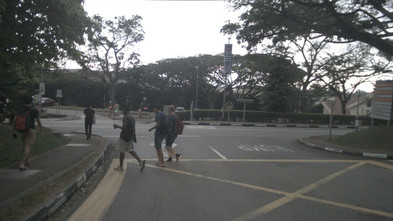}{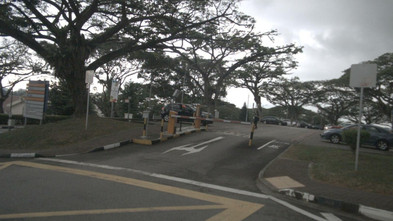}
\gradCamAttributionRowThree{VAD-Tiny}{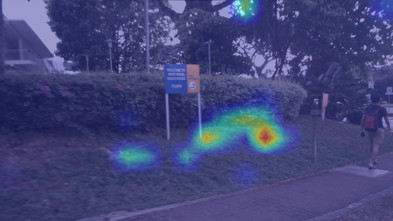}{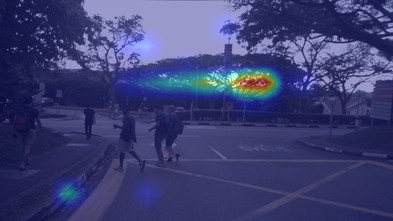}{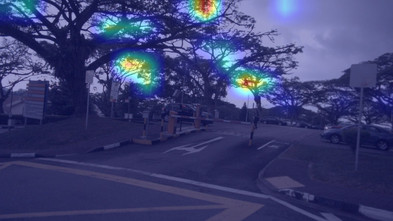}
\gradCamAttributionRowThree{+ PAVER}{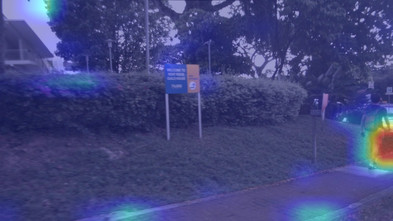}{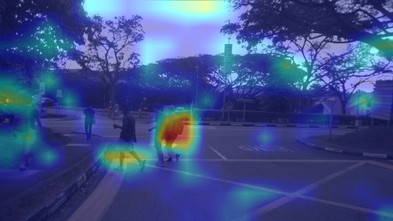}{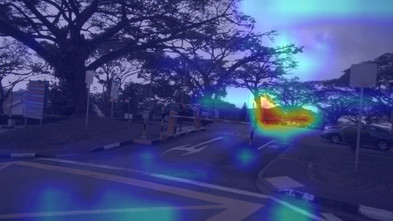}
\caption{Grad-CAM results on the nuScenes validation set.}
\label{fig:supp_vad_tiny_gradcam_three_camera_1298}
\vspace{8pt}
\gradCamCameraHeader
\gradCamInputRowThree{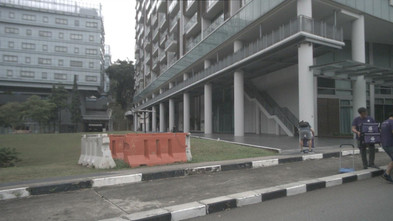}{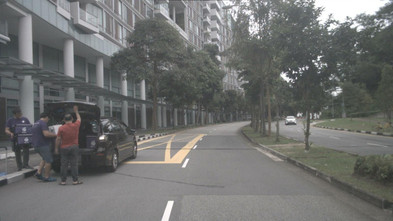}{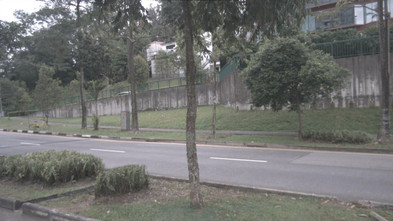}
\gradCamAttributionRowThree{VAD-Tiny}{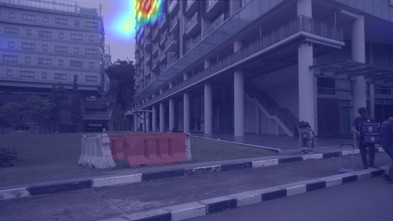}{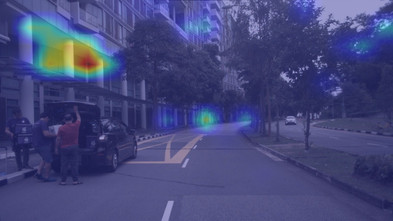}{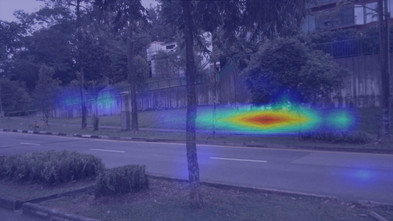}
\gradCamAttributionRowThree{+ PAVER}{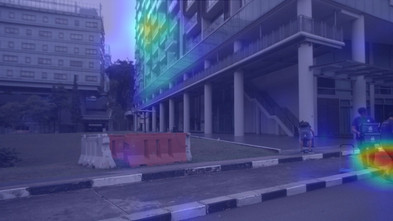}{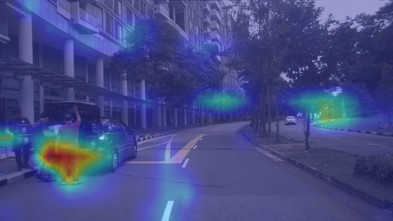}{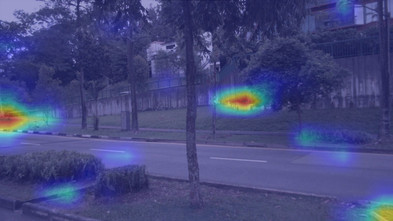}
\caption{Grad-CAM results on the nuScenes validation set.}
\label{fig:supp_vad_tiny_gradcam_three_camera_1401}
\end{figure*}
\clearpage

\begin{figure*}[p]
\centering
\gradCamCameraHeader
\gradCamInputRowThree{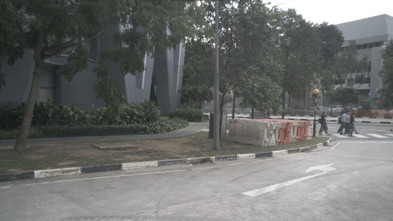}{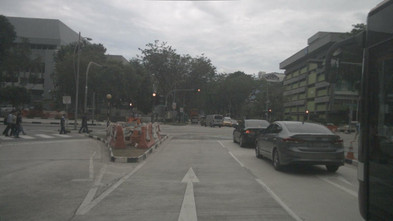}{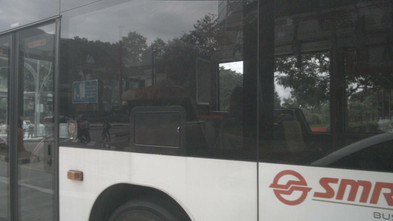}
\gradCamAttributionRowThree{VAD-Tiny}{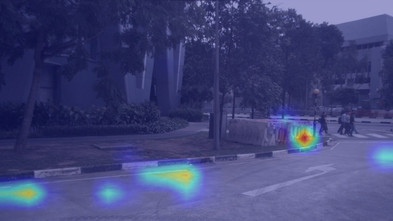}{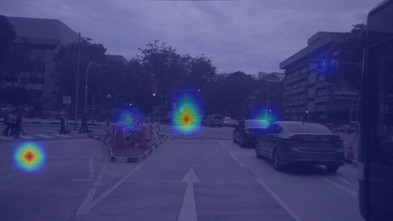}{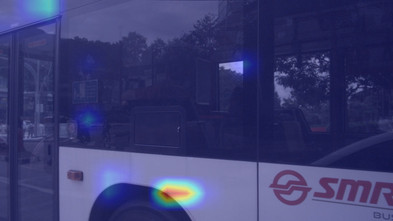}
\gradCamAttributionRowThree{+ PAVER}{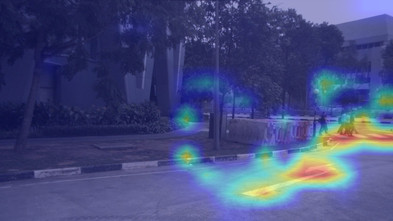}{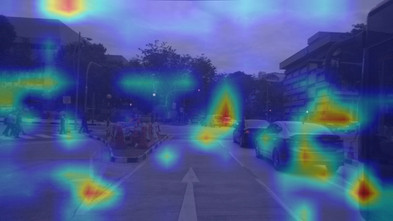}{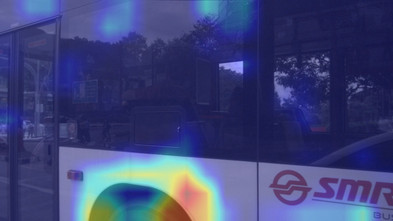}
\caption{Grad-CAM results on the nuScenes validation set.}
\label{fig:supp_vad_tiny_gradcam_three_camera_1430}
\vspace{8pt}
\gradCamCameraHeader
\gradCamInputRowThree{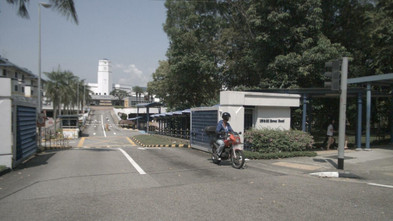}{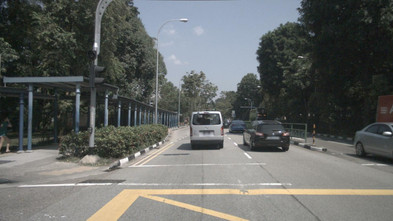}{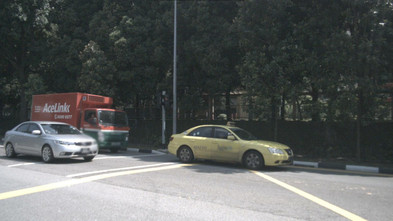}
\gradCamAttributionRowThree{VAD-Tiny}{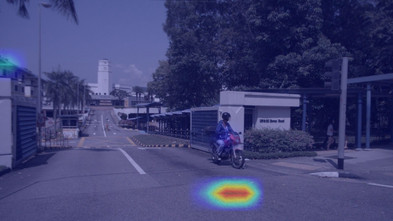}{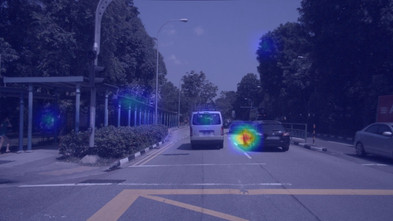}{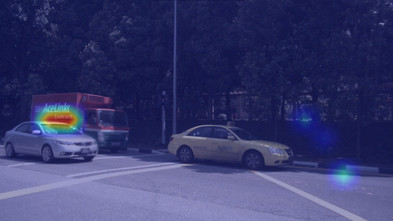}
\gradCamAttributionRowThree{+ PAVER}{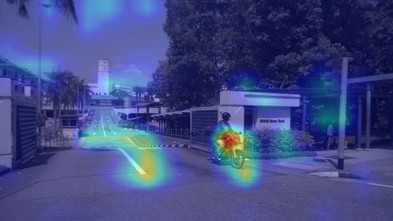}{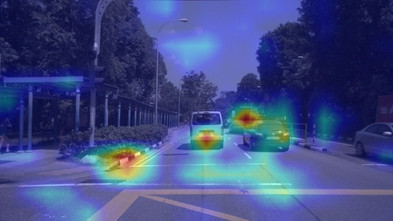}{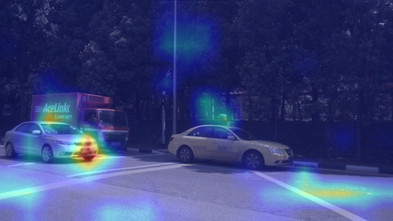}
\caption{Grad-CAM results on the nuScenes validation set.}
\label{fig:supp_vad_tiny_gradcam_three_camera_1440}
\end{figure*}
\clearpage

\ifdefined\arxivpreprint\else
\begin{figure*}[p]
\centering
\gradCamCameraHeader
\gradCamInputRowThree{figures/analysis/gradcam_vad_tiny/panels/rank_1647__59a27b7fc76541ea92d79f7d4c62a2ec__front_left__input.png}{figures/analysis/gradcam_vad_tiny/panels/rank_1647__59a27b7fc76541ea92d79f7d4c62a2ec__front__input.png}{figures/analysis/gradcam_vad_tiny/panels/rank_1647__59a27b7fc76541ea92d79f7d4c62a2ec__front_right__input.png}
\gradCamAttributionRowThree{VAD-Tiny}{figures/analysis/gradcam_vad_tiny/panels/rank_1647__59a27b7fc76541ea92d79f7d4c62a2ec__front_left__vad_tiny_bev.png}{figures/analysis/gradcam_vad_tiny/panels/rank_1647__59a27b7fc76541ea92d79f7d4c62a2ec__front__vad_tiny_bev.png}{figures/analysis/gradcam_vad_tiny/panels/rank_1647__59a27b7fc76541ea92d79f7d4c62a2ec__front_right__vad_tiny_bev.png}
\gradCamAttributionRowThree{+ PAVER}{figures/analysis/gradcam_vad_tiny/panels/rank_1647__59a27b7fc76541ea92d79f7d4c62a2ec__front_left__paver_bev.png}{figures/analysis/gradcam_vad_tiny/panels/rank_1647__59a27b7fc76541ea92d79f7d4c62a2ec__front__paver_bev.png}{figures/analysis/gradcam_vad_tiny/panels/rank_1647__59a27b7fc76541ea92d79f7d4c62a2ec__front_right__paver_bev.png}
\caption{Grad-CAM response to a traffic signal on the nuScenes validation set. PAVER concentrates attribution on the signal.}
\label{fig:supp_vad_tiny_gradcam_three_camera_1647}
\vspace{8pt}
\gradCamCameraHeader
\gradCamInputRowThree{figures/analysis/gradcam_vad_tiny/panels/rank_1670__c62b6ac878934607be0c524a02f1692b__front_left__input.png}{figures/analysis/gradcam_vad_tiny/panels/rank_1670__c62b6ac878934607be0c524a02f1692b__front__input.png}{figures/analysis/gradcam_vad_tiny/panels/rank_1670__c62b6ac878934607be0c524a02f1692b__front_right__input.png}
\gradCamAttributionRowThree{VAD-Tiny}{figures/analysis/gradcam_vad_tiny/panels/rank_1670__c62b6ac878934607be0c524a02f1692b__front_left__vad_tiny_bev.png}{figures/analysis/gradcam_vad_tiny/panels/rank_1670__c62b6ac878934607be0c524a02f1692b__front__vad_tiny_bev.png}{figures/analysis/gradcam_vad_tiny/panels/rank_1670__c62b6ac878934607be0c524a02f1692b__front_right__vad_tiny_bev.png}
\gradCamAttributionRowThree{+ PAVER}{figures/analysis/gradcam_vad_tiny/panels/rank_1670__c62b6ac878934607be0c524a02f1692b__front_left__paver_bev.png}{figures/analysis/gradcam_vad_tiny/panels/rank_1670__c62b6ac878934607be0c524a02f1692b__front__paver_bev.png}{figures/analysis/gradcam_vad_tiny/panels/rank_1670__c62b6ac878934607be0c524a02f1692b__front_right__paver_bev.png}
\caption{Grad-CAM results on the nuScenes validation set.}
\label{fig:supp_vad_tiny_gradcam_three_camera_1670}
\end{figure*}
\fi
\clearpage

\ifdefined\arxivpreprint\else
\begin{figure*}[p]
\centering
\gradCamCameraHeader
\gradCamInputRowThree{figures/analysis/gradcam_vad_tiny/panels/rank_1741__699e94e9f1cc4dffa7c58c00fd3a740f__front_left__input.png}{figures/analysis/gradcam_vad_tiny/panels/rank_1741__699e94e9f1cc4dffa7c58c00fd3a740f__front__input.png}{figures/analysis/gradcam_vad_tiny/panels/rank_1741__699e94e9f1cc4dffa7c58c00fd3a740f__front_right__input.png}
\gradCamAttributionRowThree{VAD-Tiny}{figures/analysis/gradcam_vad_tiny/panels/rank_1741__699e94e9f1cc4dffa7c58c00fd3a740f__front_left__vad_tiny_bev.png}{figures/analysis/gradcam_vad_tiny/panels/rank_1741__699e94e9f1cc4dffa7c58c00fd3a740f__front__vad_tiny_bev.png}{figures/analysis/gradcam_vad_tiny/panels/rank_1741__699e94e9f1cc4dffa7c58c00fd3a740f__front_right__vad_tiny_bev.png}
\gradCamAttributionRowThree{+ PAVER}{figures/analysis/gradcam_vad_tiny/panels/rank_1741__699e94e9f1cc4dffa7c58c00fd3a740f__front_left__paver_bev.png}{figures/analysis/gradcam_vad_tiny/panels/rank_1741__699e94e9f1cc4dffa7c58c00fd3a740f__front__paver_bev.png}{figures/analysis/gradcam_vad_tiny/panels/rank_1741__699e94e9f1cc4dffa7c58c00fd3a740f__front_right__paver_bev.png}
\caption{Grad-CAM results on the nuScenes validation set.}
\label{fig:supp_vad_tiny_gradcam_three_camera_1741}
\vspace{8pt}
\gradCamCameraHeader
\gradCamInputRowThree{figures/analysis/gradcam_vad_tiny/panels/rank_1747__45326326fd9c44e18d44e9e698454ef6__front_left__input.png}{figures/analysis/gradcam_vad_tiny/panels/rank_1747__45326326fd9c44e18d44e9e698454ef6__front__input.png}{figures/analysis/gradcam_vad_tiny/panels/rank_1747__45326326fd9c44e18d44e9e698454ef6__front_right__input.png}
\gradCamAttributionRowThree{VAD-Tiny}{figures/analysis/gradcam_vad_tiny/panels/rank_1747__45326326fd9c44e18d44e9e698454ef6__front_left__vad_tiny_bev.png}{figures/analysis/gradcam_vad_tiny/panels/rank_1747__45326326fd9c44e18d44e9e698454ef6__front__vad_tiny_bev.png}{figures/analysis/gradcam_vad_tiny/panels/rank_1747__45326326fd9c44e18d44e9e698454ef6__front_right__vad_tiny_bev.png}
\gradCamAttributionRowThree{+ PAVER}{figures/analysis/gradcam_vad_tiny/panels/rank_1747__45326326fd9c44e18d44e9e698454ef6__front_left__paver_bev.png}{figures/analysis/gradcam_vad_tiny/panels/rank_1747__45326326fd9c44e18d44e9e698454ef6__front__paver_bev.png}{figures/analysis/gradcam_vad_tiny/panels/rank_1747__45326326fd9c44e18d44e9e698454ef6__front_right__paver_bev.png}
\caption{Grad-CAM results on the nuScenes validation set.}
\label{fig:supp_vad_tiny_gradcam_three_camera_1747}
\end{figure*}
\fi
\clearpage

\ifdefined\arxivpreprint\else
\begin{figure*}[p]
\centering
\gradCamCameraHeader
\gradCamInputRowThree{figures/analysis/gradcam_vad_tiny/panels/rank_1781__67fa7b67228546379e22d0481e3d0331__front_left__input.png}{figures/analysis/gradcam_vad_tiny/panels/rank_1781__67fa7b67228546379e22d0481e3d0331__front__input.png}{figures/analysis/gradcam_vad_tiny/panels/rank_1781__67fa7b67228546379e22d0481e3d0331__front_right__input.png}
\gradCamAttributionRowThree{VAD-Tiny}{figures/analysis/gradcam_vad_tiny/panels/rank_1781__67fa7b67228546379e22d0481e3d0331__front_left__vad_tiny_bev.png}{figures/analysis/gradcam_vad_tiny/panels/rank_1781__67fa7b67228546379e22d0481e3d0331__front__vad_tiny_bev.png}{figures/analysis/gradcam_vad_tiny/panels/rank_1781__67fa7b67228546379e22d0481e3d0331__front_right__vad_tiny_bev.png}
\gradCamAttributionRowThree{+ PAVER}{figures/analysis/gradcam_vad_tiny/panels/rank_1781__67fa7b67228546379e22d0481e3d0331__front_left__paver_bev.png}{figures/analysis/gradcam_vad_tiny/panels/rank_1781__67fa7b67228546379e22d0481e3d0331__front__paver_bev.png}{figures/analysis/gradcam_vad_tiny/panels/rank_1781__67fa7b67228546379e22d0481e3d0331__front_right__paver_bev.png}
\caption{Grad-CAM results on the nuScenes validation set.}
\label{fig:supp_vad_tiny_gradcam_three_camera_1781}
\vspace{8pt}
\gradCamCameraHeader
\gradCamInputRowThree{figures/analysis/gradcam_vad_tiny/panels/rank_1809__77d0e39819774014a8344a6d482edd2d__front_left__input.png}{figures/analysis/gradcam_vad_tiny/panels/rank_1809__77d0e39819774014a8344a6d482edd2d__front__input.png}{figures/analysis/gradcam_vad_tiny/panels/rank_1809__77d0e39819774014a8344a6d482edd2d__front_right__input.png}
\gradCamAttributionRowThree{VAD-Tiny}{figures/analysis/gradcam_vad_tiny/panels/rank_1809__77d0e39819774014a8344a6d482edd2d__front_left__vad_tiny_bev.png}{figures/analysis/gradcam_vad_tiny/panels/rank_1809__77d0e39819774014a8344a6d482edd2d__front__vad_tiny_bev.png}{figures/analysis/gradcam_vad_tiny/panels/rank_1809__77d0e39819774014a8344a6d482edd2d__front_right__vad_tiny_bev.png}
\gradCamAttributionRowThree{+ PAVER}{figures/analysis/gradcam_vad_tiny/panels/rank_1809__77d0e39819774014a8344a6d482edd2d__front_left__paver_bev.png}{figures/analysis/gradcam_vad_tiny/panels/rank_1809__77d0e39819774014a8344a6d482edd2d__front__paver_bev.png}{figures/analysis/gradcam_vad_tiny/panels/rank_1809__77d0e39819774014a8344a6d482edd2d__front_right__paver_bev.png}
\caption{Grad-CAM results on the nuScenes validation set.}
\label{fig:supp_vad_tiny_gradcam_three_camera_1809}
\end{figure*}
\fi
\clearpage

\clearpage
\newcommand{\vadBaseGradCamPath}[3]{figures/analysis/gradcam_vad_base/panels/rank_#1__#2__#3}

\newcommand{\vadBaseGradCamScene}[4]{%
\begin{figure*}[p]
\centering
\gradCamCameraHeader
\gradCamGridRow{\footnotesize Input Image}
  {\vadBaseGradCamPath{#1}{#2}{front_left__input}}
  {\vadBaseGradCamPath{#1}{#2}{front__input}}
  {\vadBaseGradCamPath{#1}{#2}{front_right__input}}
\gradCamAttributionRowThree{\footnotesize Scratch (BEV)}
  {\vadBaseGradCamPath{#1}{#2}{front_left__scratch_bev}}
  {\vadBaseGradCamPath{#1}{#2}{front__scratch_bev}}
  {\vadBaseGradCamPath{#1}{#2}{front_right__scratch_bev}}
\gradCamAttributionRowThree{\footnotesize PAVER (BEV)}
  {\vadBaseGradCamPath{#1}{#2}{front_left__paver_bev}}
  {\vadBaseGradCamPath{#1}{#2}{front__paver_bev}}
  {\vadBaseGradCamPath{#1}{#2}{front_right__paver_bev}}
\gradCamAttributionRowThree{\footnotesize Scratch (Plan.)}
  {\vadBaseGradCamPath{#1}{#2}{front_left__scratch_planning}}
  {\vadBaseGradCamPath{#1}{#2}{front__scratch_planning}}
  {\vadBaseGradCamPath{#1}{#2}{front_right__scratch_planning}}
\gradCamAttributionRowThree{\footnotesize PAVER (Plan.)}
  {\vadBaseGradCamPath{#1}{#2}{front_left__paver_planning}}
  {\vadBaseGradCamPath{#1}{#2}{front__paver_planning}}
  {\vadBaseGradCamPath{#1}{#2}{front_right__paver_planning}}
\caption{Grad-CAM spatial localization for VAD-Base, scene #4. Grad-CAM maps for the BEV and planning targets with and without PAVER pretraining, shown for the three front cameras.}
\label{fig:supp_vad_base_gradcam_#4a}
\end{figure*}
\clearpage

\begin{figure*}[p]
\centering
\gradCamCameraHeader
\gradCamGridRow{\footnotesize Input Image}
  {\vadBaseGradCamPath{#1}{#2}{front_left__input}}
  {\vadBaseGradCamPath{#1}{#2}{front__input}}
  {\vadBaseGradCamPath{#1}{#2}{front_right__input}}
\gradCamAttributionRowThree{\footnotesize Scratch (Det.)}
  {\vadBaseGradCamPath{#1}{#2}{front_left__scratch_detection}}
  {\vadBaseGradCamPath{#1}{#2}{front__scratch_detection}}
  {\vadBaseGradCamPath{#1}{#2}{front_right__scratch_detection}}
\gradCamAttributionRowThree{\footnotesize PAVER (Det.)}
  {\vadBaseGradCamPath{#1}{#2}{front_left__paver_detection}}
  {\vadBaseGradCamPath{#1}{#2}{front__paver_detection}}
  {\vadBaseGradCamPath{#1}{#2}{front_right__paver_detection}}
\gradCamAttributionRowThree{\footnotesize Scratch (Mot.)}
  {\vadBaseGradCamPath{#1}{#2}{front_left__scratch_motion}}
  {\vadBaseGradCamPath{#1}{#2}{front__scratch_motion}}
  {\vadBaseGradCamPath{#1}{#2}{front_right__scratch_motion}}
\gradCamAttributionRowThree{\footnotesize PAVER (Mot.)}
  {\vadBaseGradCamPath{#1}{#2}{front_left__paver_motion}}
  {\vadBaseGradCamPath{#1}{#2}{front__paver_motion}}
  {\vadBaseGradCamPath{#1}{#2}{front_right__paver_motion}}
\gradCamAttributionRowThree{\footnotesize Scratch (Map)}
  {\vadBaseGradCamPath{#1}{#2}{front_left__scratch_map}}
  {\vadBaseGradCamPath{#1}{#2}{front__scratch_map}}
  {\vadBaseGradCamPath{#1}{#2}{front_right__scratch_map}}
\gradCamAttributionRowThree{\footnotesize PAVER (Map)}
  {\vadBaseGradCamPath{#1}{#2}{front_left__paver_map}}
  {\vadBaseGradCamPath{#1}{#2}{front__paver_map}}
  {\vadBaseGradCamPath{#1}{#2}{front_right__paver_map}}
\caption{Grad-CAM spatial localization for VAD-Base, scene #4. Grad-CAM maps for the detection, motion and map targets with and without PAVER pretraining, shown for the three front cameras.}
\label{fig:supp_vad_base_gradcam_#4b}
\end{figure*}
\clearpage
}

\vadBaseGradCamScene{028}{7fb67ec23c2b4086b4e97b76b82c3f07}{7fb67ec2}{1}
\vadBaseGradCamScene{113}{b9d7fcc30e964b7ba11099fa06d7f8bf}{b9d7fcc3}{3}
\vadBaseGradCamScene{128}{c62b6ac878934607be0c524a02f1692b}{c62b6ac8}{4}
\vadBaseGradCamScene{131}{e8dd2a90a84142b2aa1c42126d3f135b}{e8dd2a90}{5}

\end{document}